\documentclass{article}
\usepackage{amsmath,amssymb,amsthm}
\usepackage{iclr2027_conference,times}
\usepackage{hyperref}
\usepackage{url}
\usepackage{graphicx}
\newtheorem{theorem}{Theorem}[section]
\newtheorem{lemma}[theorem]{Lemma}

\newtheorem{proposition}[theorem]{Proposition}

\newtheorem{definition}{Definition}
\newtheorem{assumption}{Assumption}

\usepackage{booktabs}
\usepackage{mathtools}
\usepackage{xcolor}
\usepackage{enumitem}

\DeclareMathOperator{\Cov}{Cov}
\DeclareMathOperator{\Tr}{Tr}
\DeclareMathOperator{\Var}{Var}

\title{Model Collapse Is Avoidable Even as Real Data Vanishes: A Sharp Criterion for Iterative Gaussian Estimation}

\author{Roy Turgeman \\
Faculty of Engineering\\
Bar-Ilan University\\
Ramat Gan, Israel \\
\And
Tom Tirer \\
Faculty of Engineering \\
Bar-Ilan University \\
Ramat Gan, Israel \\
}

\iclrfinalcopy
\begin{document}
\maketitle

\begin{abstract}
Large generative models are commonly trained on data collected from the open web, which may contain data produced by previous generations of the same model. Previous work has shown that recursively training on model-generated data can cause a progressive degradation in model performance, a phenomenon known as \emph{model collapse}. Yet, currently, generative models continue to improve dramatically over time. Existing theoretical research suggests that this is enabled by the availability of new real data at each iteration, kept separate from model-generated data. However, exact separation is not realistic, and uncurated data is still commonly used. We bridge this gap by considering a less restrictive and more realistic setup in which real and generated data are not handled differently during training. Specifically, we present a comprehensive theoretical study of iterative estimation of the mean and covariance of a multivariate Gaussian data distribution. At each iteration, samples are drawn from a mixture of the real data distribution and the estimated distribution from the previous iteration, but are used indistinguishably. We provide a single equivalent condition for consistency of the mean and covariance estimators in Euclidean norms, 
and convergence to zero of the expected squared $2$-Wasserstein distance between the estimated and true distributions. Using this result, we then show that consistency holds, i.e., model collapse is avoided, for any fixed real data fraction above zero, and even for a fraction that vanishes across iterations, provided it does not vanish too quickly. 
However, more real data (or data curation) remains important at practical finite times:
we show that the convergence rate of the mean estimator deteriorates as the real data fraction approaches zero.
We support our results with an empirical investigation of the theoretical setup, 
and complement our theoretical work with experiments on iterative training of a diffusion model on CIFAR-10 
\textcolor{black}{that correlate with our theory}.

\end{abstract}

\section{Introduction}
\label{sec:introduction}

A growing body of empirical and theoretical work has raised the concern that machine
learning generative models trained iteratively on data produced by earlier generations of
themselves — rather than on real, independently sourced data — can drift away from the true underlying distribution over successive generations, a phenomenon often called \emph{model collapse} \citep{shumailov2024ai, alemohammad2024selfconsuming}. As more of the data on which models are trained is itself generated by other models, understanding the conditions under which this recursive process is destructive to the model and how we can deal with that has become an important theoretical question \citep{bertrand2023stability,gerstgrasser2024model, kazdan2024collapse, seddik2024bad,dohmatob2025strong,dohmatob2024model,barzilai2025models,he2025golden,garg2026preventing}.

There are several approaches to iterative retraining, two of which are commonly referred to as \emph{Accumulate} and \emph{Replace}. In the \emph{Accumulate} method, we start with a set of real data, and at each iteration, we augment the training set with newly generated data while retaining all data from previous iterations. In contrast, in the \emph{Replace} method, the training set at each iteration consists only of the newly generated data from the previous iteration; all data from earlier iterations are discarded. 
The Accumulate method is generally considered a more realistic setting for iterative retraining and has been shown to exhibit degradation that remains bounded across iterations \citep{gerstgrasser2024model, kazdan2024collapse, seddik2024bad}. In contrast, the Replace method can lead to unbounded degradation, with the learned distribution progressively diverging from the original data distribution \citep{shumailov2024ai, kazdan2024collapse}. %

Existing literature tries to theoretically explain the dramatic improvement of current top-tier generative models. This is done by considering the existence of independent real data in each iteration. However, it has only been investigated under the assumption that the real data can be perfectly separated from the generated data \citep{alemohammad2024selfconsuming, he2025golden, garg2026preventing}. In this case, an optimal weighting scheme for the separate real and generated data used for training has been studied \citep{he2025golden}.
However, exact separation is not realistic \citep{sadasivan2025can,li2025bridging}, and uncurated data is still commonly used.
In this paper, to the best of our knowledge, we are the first to bridge this gap by considering a less restrictive and more realistic setup in which real and generated data are not
handled differently during training. 

We study \emph{Accumulate}-based iterative estimation 
of the mean $\boldsymbol{\mu} \in \mathbb{R}^d$ and covariance matrix $\boldsymbol{\Sigma} \in \mathbb{R}^{d\times d}$ of a Gaussian data distribution $\mathcal{N}(\boldsymbol{\mu},\boldsymbol{\Sigma})$, a standard setting in theoretical works on model collapse \citep{shumailov2024ai, alemohammad2024selfconsuming, kazdan2024collapse, he2025golden}. At each round $t$, a batch of new samples is drawn from a \emph{mixture} of (i) the current fitted Gaussian model and (ii) the true distribution, and all data accumulated so far is used to refit the mean and covariance by standard empirical estimators, $\boldsymbol{\mu}^{(t)}$ and $\boldsymbol{\Sigma}^{(t)}$, but \emph{without distinguishing} between real and generated samples. This setting strips away the complications of neural networks while retaining the essential recursive feedback loop and admits %
rigorous analysis.

Our guiding question is: under what conditions does this process \emph{not} collapse. In other words, when do the iterated estimators $(\boldsymbol{\mu}^{(t)},\boldsymbol{\Sigma}^{(t)})$ converge to the true $(\boldsymbol{\mu},\boldsymbol{\Sigma})$, and when do they instead settle on a biased,
degenerate limit? We answer this question completely %
for both estimators.
For the mean estimator, we also analyze the rate of convergence. Here, the \emph{mixing fraction} $\alpha_t \in [0,1]$, 
the expected fraction of data in the new batch drawn from the current model, %
plays a decisive role, and we provide an explicit, checkable condition on how it may vary across rounds while still guaranteeing consistency. That separates collapse from non-collapse. Our main contributions include:
\begin{itemize}[leftmargin=*]
    \item We prove that the mean estimator $\boldsymbol{\mu}^{(t)}$ is unbiased, while the covariance estimator $\boldsymbol{\Sigma}^{(t)}$ is biased, with the exact bias given by $-\Cov(\boldsymbol{\mu}^{(t)})$.

    \item We provide a closed-form equivalent condition for the consistency of the estimated mean in $L^2$, the consistency of the covariance estimator in {\color{black} expected} Frobenius norm, and the convergence to zero of the {\color{black} expected} squared $2$-Wasserstein distance between the estimated and true distributions.  %
    
    \item We use our result to show that several scenarios that naturally arise when scraping the web for data are sufficient to guarantee convergence, thereby preventing model collapse. First, convergence holds for an arbitrarily small but fixed positive fraction of real data ($1-\alpha$) at every iteration. Second, convergence still holds when the fraction of real data vanishes across iterations, provided that it does not vanish too quickly.

    \item For a fixed synthetic fraction $0 \le \alpha < 1$ (i.e., when not all data is generated), we derive the exact rate of convergence (up to constant factors) of the mean estimator to the true mean. Specifically, we prove that if less than half of each new batch is generated, then the rate of convergence matches the optimal rate. However, when more than half of each new batch is generated, the rate of convergence becomes arbitrarily slow as the fraction of synthetic data increases. This result motivates the ongoing research on the detection and curation of generated samples: it shows that for practical finite $t$ in the high-synthetic regime, synthetic data detection can substantially improve estimation accuracy.

    \item We present an empirical investigation of the theoretical model that supports our theoretical results and provides further insight into the rate of convergence.

    \item We complement our theoretical work with an empirical study of how different fixed real-data ratios affect the FID of a diffusion model iteratively trained on CIFAR-10, showing results that correlate with our theory. %

\end{itemize}

\section{Related work}
\label{sec:related_work}

{\color{black}
Existing works have shown that under the \emph{Replace} workflow, iterative retraining degrades without bound \citep{shumailov2024ai, dohmatob2024model, kazdan2024collapse}, whereas under the \emph{Accumulate} workflow the degradation remains bounded \citep{gerstgrasser2024model, kazdan2024collapse, seddik2024bad}.

Several works show that collapse can be avoided, but under restrictive conditions. \citet{bertrand2023stability} prove the stability of iterative retraining on a mixture of real and synthetic data, but only locally, when the initial model is sufficiently close to the data distribution and the proportion of real data is sufficiently large. \citet{barzilai2025models} provide non-asymptotic guarantees under data accumulation; however, for a fixed sample size $n$, their bound holds only up to a finite number of iterations $t$ that depends on $n$, and thus does not cover the regime of fixed $n$ with $t\to\infty$. Moreover, beyond an initial (or fixed) real dataset, these works do not incorporate new real data at subsequent iterations.

In practice, data scraped from the web at each iteration contains new real data alongside generated data. Several works study this setting \citep{alemohammad2024selfconsuming, he2025golden, garg2026preventing}, but either discard earlier data or assume that real and generated samples can be told apart, e.g., in order to weight them separately. The latter is unrealistic, since reliably detecting generated content remains an open problem \citep{sadasivan2025can, li2025bridging}, although detection, when possible, can substantially mitigate collapse \citep{drayson2025machine}. To the best of our knowledge, no prior work analyzes iterative retraining with accumulated data in which fresh real samples are not distinguished from generated ones.}

\section{Theory}
\label{sec:theory}

In this section, we formalize the iterative retraining process, define the corresponding mean and covariance estimators, and state our main theoretical results, which characterize when the estimators converge to the true parameters and the rate at which this convergence occurs. Finally, we validate our results through experiments and provide additional insights into the rate of decay.

\subsection{Problem setup: data generating process and estimators} %
\label{sec:setup}

In the theoretical model, we consider the accumulate method, where the process is initialized with a dataset of real samples. Similar to a vast body of theoretical work on model collapse \citep{shumailov2024ai, alemohammad2024selfconsuming, kazdan2024collapse, dohmatob2024model, he2025golden, dohmatob2025strong}, we assume that the real data distribution is a multivariate Gaussian distribution. At each iteration, the training dataset is augmented with both generated and real data. Since the model does not know which samples are real and which are generated, it treats all samples identically and estimates the underlying distribution using the entire accumulated dataset. Formally, at each iteration, each newly added sample is drawn from the real distribution with some probability and from the current estimated distribution with the complementary probability.

\textbf{Data-generating process.} Let $\boldsymbol{\mu}\in\mathbb{R}^d$, and let $\boldsymbol{\Sigma} \in \mathbb{R}^{d\times d}$ be symmetric positive-definite. The process unfolds in discrete iterations $t=1,2,3,\ldots$

\medskip
\noindent\textbf{\emph{Iteration 1 (real data).}}
Observe $n_r \ge 1$ i.i.d.\ samples
$\{\boldsymbol{x}^{(0)}_j\}_{j=1}^{n_r}\sim\mathcal{N}(\boldsymbol{\mu},\boldsymbol{\Sigma})$ and form the empirical
mean and covariance
\begin{equation}\label{eq:init}
  \boldsymbol{\mu}^{(1)}=\frac{1}{n_r}\sum_{j=1}^{n_r}\boldsymbol{x}^{(0)}_j,\qquad
  \boldsymbol{\Sigma}^{(1)}=\frac{1}{n_r}\sum_{j=1}^{n_r}
    \bigl(\boldsymbol{x}^{(0)}_j-\boldsymbol{\mu}^{(1)}\bigr)\bigl(\boldsymbol{x}^{(0)}_j-\boldsymbol{\mu}^{(1)}\bigr)^\top.
\end{equation}

\noindent\textbf{\emph{Iteration $t \ge 2$ (mixed data).}}
Given $\boldsymbol{\mu}^{(t-1)}$ and $\boldsymbol{\Sigma}^{(t-1)}$, draw
$n_b(t-1)\ge1$ i.i.d. samples from the mixture
\begin{equation}\label{eq:mixture}
  \boldsymbol{x}^{(t-1)}_i\ \sim
    \alpha_{t-1}\,\mathcal{N}\!\bigl(\boldsymbol{\mu}^{(t-1)},\boldsymbol{\Sigma}^{(t-1)}\bigr)
    +(1-\alpha_{t-1})\,\mathcal{N}(\boldsymbol{\mu},\boldsymbol{\Sigma}),
  \quad\alpha_{t-1}\in[0,1],
\end{equation}
where $1 \le i \le n_b(t - 1)$. The parameter $\alpha_{t-1} \in [0,1]$ controls the fraction of the new batch that
is drawn from the current model. Specifically, $\alpha_{t-1}=1$ means the new batch is entirely
synthetic, while $\alpha_{t-1}=0$ means it is entirely real.

Denote the total number of samples at iteration $t$ by $N_t$ and the inverse ratio of the growth by $\rho_t$:
\begin{equation}\label{eq:counts}
    N_t := n_r+\sum_{s=1}^{t-1}n_b(s), \qquad
    \rho_t := \frac{N_t}{N_{t+1}} \qquad (t \ge 1).
\end{equation}
Note that $\rho_t \in (0,1)$ for every $t \ge 1$. %

\textbf{Estimators.} At every round, the model is refit using the pooled empirical mean and covariance of all
samples accumulated so far. For every $t \ge 1$, we write $\boldsymbol{y}_1,\ldots,\boldsymbol{y}_{N_t}$ for the $N_t$
accumulated points, consisting of the original real batch $\{\boldsymbol{x}^{(0)}_j\}_{j=1}^{n_r}$ and the
accumulated batches  $\{\boldsymbol{x}^{(s)}_i\}_{i=1}^{n_b(s)}, \  1 \le s \le t-1$. Define the pooled empirical estimators:
\begin{equation}
\label{eq:pooled-def}
  \boldsymbol{\mu}^{(t)} \coloneqq \frac{1}{N_t}\sum_{k=1}^{N_t}\boldsymbol{y}_k,\qquad
  \boldsymbol{\Sigma}^{(t)} \coloneqq \frac{1}{N_t}\sum_{k=1}^{N_t}
    \bigl(\boldsymbol{y}_k-\boldsymbol{\mu}^{(t)}\bigr)\bigl(\boldsymbol{y}_k-\boldsymbol{\mu}^{(t)}\bigr)^\top,
\end{equation}
which for $t=1$ coincides with \eqref{eq:init}. These iterative estimators are standard and widely used in the literature \citep{shumailov2024ai,kazdan2024collapse}. 
Conveniently, they admit online recursions in terms of the previous estimates and the newly drawn batch. 
Finally, we write
\begin{equation}
\label{eq:model-def}
  \mathcal{D}_t \coloneqq \mathcal{N}\bigl(\boldsymbol{\mu}^{(t)},\boldsymbol{\Sigma}^{(t)}\bigr),
  \qquad
  \mathcal{D}_{\mathrm{real}} \coloneqq \mathcal{N}(\boldsymbol{\mu},\boldsymbol{\Sigma}),
\end{equation}
for the estimated distribution at iteration $t$ and the true data distribution, respectively. 
{\color{black} We measure the discrepancy between them by the $2$-Wasserstein distance, 
\textcolor{black}{$W_2(\cdot, \cdot)$, defined in Appendix~\ref{sec:W2->0 cond}, which has also been used by \citet{shumailov2024ai}. %
For $\boldsymbol{\mu}^{(t)}$ and $\boldsymbol{\Sigma}^{(t)}$ themselves, we measure the discrepancy from the true values by the Euclidean norm, $\|\cdot\|_2$, and the Frobenius norm, $\| \cdot \|_F$.}

{\color{black}
Finally, we state our standing assumption on how new samples are generated.

\begin{assumption}[Conditional independence]\label{ass:cond-ind}
For every $t\ge2$, \textbf{conditionally} on $(\boldsymbol{\mu}^{(t-1)},\boldsymbol{\Sigma}^{(t-1)})$, the new samples $\{\boldsymbol{x}^{(t-1)}_i\}_{i=1}^{n_b(t-1)}$ are drawn i.i.d.\ from the mixture~\eqref{eq:mixture} and are independent of all previously drawn samples $\{\boldsymbol{x}^{(s)}_i\}$, $0 \le s\le t-2$. 
\end{assumption}

In words, the past influences a new batch only through the current fitted model. This is not a restriction on the setting, but a formalization of the data-generating process described above.}

\subsection{Theoretical results}
\label{sec:theoretical_results}

In this section, we present our key theoretical results. %
In Section~\ref{sec:bias_of_estimators}, we establish that the mean estimator is unbiased and derive an explicit formula for the bias of the covariance estimator. In Section~\ref{sec:convergence_true_dist}, we provide an equivalent (if and only if) condition for the convergence of the estimated parameters to the truth, and hence for the convergence of the estimated distribution to the true distribution. We then show that the model converges when the fraction of real data is fixed and arbitrarily small, as well as when the fraction of real data tends to zero sufficiently slowly. In Section~\ref{sec:convergence_rate_sec}, we analyze the convergence rate of the mean estimator.

The proofs for all the claims are deferred to Appendix~\ref{sec:app}.

\subsubsection{Bias of estimators and covariance of the mean estimator}
\label{sec:bias_of_estimators}
Here, we derive
the bias of each estimator and the covariance of the mean estimator. First, let us denote by $\boldsymbol{C}_t$ the covariance of the mean estimator
\begin{equation}
\label{eq:mu_cov_first_def}
    \boldsymbol{C}_t \coloneqq \operatorname{Cov}(\boldsymbol{\mu}^{(t)}) = \mathbb{E}\left [ \big ( \boldsymbol{\mu}^{(t)} - \mathbb{E}[\boldsymbol{\mu}^{(t)}] \big ) \big ( \boldsymbol{\mu}^{(t)} - \mathbb{E}[\boldsymbol{\mu}^{(t)}] \big )^\top \right ],
\end{equation}
and by $\boldsymbol{\delta}^{(t)}$ the deviation of the mean estimator from the truth:
\begin{equation}
\label{eq:delta-def}
    \boldsymbol{\delta}^{(t)} \coloneqq \boldsymbol{\mu}^{(t)}-\boldsymbol{\mu}.
\end{equation}

We now establish the unbiasedness of the mean estimator and the bias of the covariance estimator.

\begin{theorem}[Bias of estimators] \label{thm:estimators_bias}
Under the setup of Section~\ref{sec:setup}, %
for all $t \ge 1$
\begin{enumerate}
    \item[(a)] The mean estimator is unbiased:
    \begin{equation}
    \label{eq:mu_is_unbiased}           \mathbb{E}\bigl[\boldsymbol{\mu}^{(t)}\bigr]=\boldsymbol{\mu}.
    \end{equation}
    
    \item[(b)] The bias of the covariance estimator is:
    \begin{equation}
    \label{eq:Sigma-bias}
      \mathbb{E}\bigl[\boldsymbol{\Sigma}^{(t)}\bigr]=\boldsymbol{\Sigma}-\boldsymbol{C}_t.
    \end{equation}
\end{enumerate}
\end{theorem}

{\color{black} \textbf{Proof idea.} Part (a) follows by induction using the recursion formula obtained for $\boldsymbol{\mu}^{(t)}$ (Eq.~\eqref{eq:mu-update} in the appendix). For part (b), we write $\boldsymbol{\Sigma}^{(t)}$ as the empirical second moment around the \emph{true} mean $\boldsymbol{\mu}$ minus $\boldsymbol{\delta}^{(t)}\boldsymbol{\delta}^{(t)\top}$ (Lemma~\ref{lem:pooled_scatter_identity}), and then use induction.}

\textbf{Remark.} The covariance estimator is \emph{negatively} biased by exactly the covariance of the mean estimator $\boldsymbol{C}_t\succeq0$ (`$\succeq 0$' should be understood in the positive semi-definite sense). In particular, since $\boldsymbol{C}_1=\boldsymbol{\Sigma}/n_r$, we have $\mathbb{E}[\boldsymbol{\Sigma}^{(1)}]
=\frac{n_r-1}{n_r}\boldsymbol{\Sigma}$, recovering the familiar downward bias of the maximum likelihood covariance estimator under Gaussian distribution.
{\color{black} We now show that this bias never exceeds twice its value at $t = 1$, regardless of how much generated data is used.
\begin{lemma}[Bound on $\boldsymbol{C}_t$]\label{lem:bound}
Under the setup of Section~\ref{sec:setup}, for all $t \ge 1$:
\begin{equation*}
    \boldsymbol{0} \preceq \boldsymbol{C}_t \preceq \frac{2}{n_r}\,\boldsymbol{\Sigma}.
\end{equation*}
\end{lemma}
\vspace{-3mm}

\textbf{Remark.} By Theorem~\ref{thm:estimators_bias}(b), the upper bound in Lemma~\ref{lem:bound} is equivalent to $\mathbb{E}\bigl[\boldsymbol{\Sigma}^{(t)}\bigr]\succeq\bigl(1-\tfrac{2}{n_r}\bigr)\boldsymbol{\Sigma}$: the downward bias of the covariance estimator is at most twice its value at $t=1$. Note that for $n_r = 1$ the bound is weaker than $\boldsymbol{C}_t\preceq\boldsymbol{\Sigma}$, which also holds for every $t \ge 1$: it follows from Theorem~\ref{thm:estimators_bias}(b) and $\boldsymbol{\Sigma}^{(t)}\succeq\boldsymbol{0}$} 
\textcolor{black}{(by construction)}.

Since $\boldsymbol{\mu}^{(t)}$ is unbiased
(Theorem~\ref{thm:estimators_bias}(a)), from \eqref{eq:mu_cov_first_def} we have that
\begin{equation}
\label{eq:CT-def}
  \boldsymbol{C}_t 
  = \mathbb{E}\bigl[(\boldsymbol{\mu}^{(t)}-\boldsymbol{\mu})
    (\boldsymbol{\mu}^{(t)}-\boldsymbol{\mu})^\top\bigr]
  = \mathbb{E}\bigl[\boldsymbol{\delta}^{(t)}\boldsymbol{\delta}^{(t)\top}\bigr].
\end{equation}
Using $\|\boldsymbol{v}\|_2^2 = \Tr(\boldsymbol{v}\boldsymbol{v}^\top)$, the linearity of
the trace and expectation, and \eqref{eq:CT-def},
\begin{equation}
\label{eq:MSE-trace}
\begin{aligned}
    \mathbb{E}\bigl[\|\boldsymbol{\mu}^{(t)} - \boldsymbol{\mu}\|_2^2\bigr] &= \mathbb{E}\bigl[\|\boldsymbol{\delta}^{(t)}\|_2^2\bigr] = \mathbb{E}\bigl[\Tr\bigl(\boldsymbol{\delta}^{(t)}\boldsymbol{\delta}^{(t)\top}\bigr)\bigr] = \Tr\!\left(\mathbb{E}\bigl[\boldsymbol{\delta}^{(t)}\boldsymbol{\delta}^{(t)\top}\bigr]\right) = \Tr(\boldsymbol{C}_t).
\end{aligned}
\end{equation}
Thus, $\Tr(\boldsymbol{C}_t)$ is the mean-squared
error of $\boldsymbol{\mu}^{(t)}$ around $\boldsymbol{\mu}$. 

\subsubsection{Convergence to the true distribution}
\label{sec:convergence_true_dist}
Crucially, we analyze in closed form when this iterative estimation process converges to the true distribution. %
Below, we establish
that the condition for convergence of the estimated parameters to the truth collapses to a single elementary series in $\{\alpha_t\}_{t \ge 1}$, $\{n_b(t)\}_{t \ge 1}$, and $n_r$. Recall from \eqref{eq:counts} that $\rho_t \coloneqq N_t/N_{t+1}$.

\begin{theorem}[Sharp elementary criterion for consistency] 
\label{thm:unified_sharp_criterion} Under the setup of Section~\ref{sec:setup}, the following are equivalent:
\begin{enumerate}
    \item[(a)] The following series diverges:
    \begin{equation}
    \label{eq:sharp critertion series sharp thm}
        \sum_{t = 1}^\infty (1 - \alpha_t)(1 - \rho_t) = \sum_{t = 1}^\infty (1 - \alpha_t)\frac{n_b(t)}{N_{t + 1}} = \infty.
    \end{equation}
    \item[(b)] The mean estimator $\boldsymbol{\mu}^{(t)}$ converges to $\boldsymbol{\mu}$ in $L^2$, {\color{black} i.e., in mean square:}
    \begin{equation}
    \label{eq:mu(t)->mu in L2 sharp thm}
    \mathbb{E}\bigl[\|\boldsymbol{\mu}^{(t)}-\boldsymbol{\mu}\|_2^2\bigr] \xrightarrow{t\rightarrow\infty} 0.
    \end{equation}
    \item[(c)] The covariance estimator converges to $\boldsymbol{\Sigma}$ in {\color{black} expected} Frobenius norm:
    \begin{equation}
    \label{eq:sigma(t)->sigma in F norm sharp thm}
    \mathbb{E}\bigl[\|\boldsymbol{\Sigma}^{(t)}-\boldsymbol{\Sigma}\|_F\bigr] \xrightarrow{t\rightarrow\infty} 0.
    \end{equation}
    \item[(d)] The expected squared $2$-Wasserstein distance between the estimated distribution and the true distribution tends to zero:
    \begin{equation}
    \label{eq:W_2->0 sharp thm}
    \mathbb{E}\bigl[W_2^2(\mathcal{D}_t,\mathcal{D}_{\mathrm{real}})\bigr] \xrightarrow{t\rightarrow\infty} 0.
    \end{equation}
\end{enumerate}
\end{theorem}

{\color{black} \textbf{Proof idea.} Since $\mathbb{E}\bigl[\|\boldsymbol{\mu}^{(t)}-\boldsymbol{\mu}\|_2^2\bigr]=\Tr(\boldsymbol{C}_t)$, we derive a recursion for $\boldsymbol{C}_t$ (Theorem~\ref{thm:CT}) and use it to show that $\boldsymbol{C}_t\to\boldsymbol{0}$ if and only if~\eqref{eq:sharp critertion series sharp thm} holds. For the covariance, a similar recursion for the empirical second moment around the true mean gives sufficiency, while necessity follows from the bias $-\boldsymbol{C}_t$ (Theorem~\ref{thm:estimators_bias}). For $W_2$, the Gaussian closed form splits $W_2^2(\mathcal{D}_t,\mathcal{D}_{\mathrm{real}})$ into $\|\boldsymbol{\mu}^{(t)}-\boldsymbol{\mu}\|_2^2$ plus a nonnegative term bounded by $\sqrt{d}\,\|\boldsymbol{\Sigma}^{(t)}-\boldsymbol{\Sigma}\|_F$. Taking expectations yields sufficiency from the mean and covariance results, and necessity from the mean result alone.}
\textbf{Remark (interpretation).} The factor $1-\alpha_t$ is the fraction of the batch added at round $t$ that is drawn from the true distribution, while $1-\rho_t=n_b(t)/N_{t+1}$ is the weight this batch carries in the pool. Their product is therefore the share of \emph{genuinely fresh real information} injected at round $t$, and the model escapes collapse precisely when these shares are non-summable. The convergences of the mean, the covariance and the fitted distribution hold or fail together, and, by Markov's inequality, each implies convergence in probability.

\textbf{Discussion (implications).} 
In the remainder of this section, we characterize when the series $\sum_{t=1}^\infty (1-\alpha_t)(1-\rho_t)$ diverges, which, using Theorem~\ref{thm:unified_sharp_criterion}, allows us to determine when the iterative process converges to the true parameters. 

We start by providing a sufficient condition for the mixing fraction for convergence.

\begin{lemma}[Sufficient condition on the mixing fraction]
\label{lem: sup_t alpha_t < 1 => cond holds} If $\sup_{t \ge 1} \alpha_t < 1$, then 
\begin{equation}
\label{eq: sup_T alpha_T < 1 => cond holds}
    \sum_{t = 1}^\infty (1 - \alpha_t)(1 - \rho_t) = \infty
\end{equation}
\end{lemma}
In particular, for a fixed mixing fraction $\alpha_t \equiv \alpha \in [0,1)$, the condition of Lemma~\ref{lem: sup_t alpha_t < 1 => cond holds} holds. Thus, any fixed positive fraction $1-\alpha$ of real data in each batch suffices for consistency, even when the real samples added at each iteration are vastly outnumbered by the generated ones. This is encouraging in practice: scraping the web naturally yields some new real data alongside generated data, and any fixed fraction of it suffices for convergence. %

The next case we consider is an asymptotically vanishing fraction of real data.
We show that the model can still be consistent, provided that the synthetic fraction tends to $1$ slowly enough. Specifically, take $n_b(t) \equiv n_b \ge 1$, so that
\begin{equation*}
    N_{t+1} = n_r + n_b t \, \Longrightarrow \, 1 - \rho_t = \frac{n_b}{N_{t+1}} \asymp \frac{1}{t}.
\end{equation*}
Here, $f(t)\asymp g(t)$ means that for all sufficiently large $t$, the ratio $f(t)/g(t)$ is bounded above and below by positive constants (independent of $t$). The criterion \eqref{eq:sharp critertion series sharp thm} for convergence of the estimated parameters to the true parameters then reads $\sum_{t=1}^\infty \frac{1 - \alpha_t}{t} = \infty$. For mixing fractions $\{\alpha_t\}_{t=1}^\infty$ tending to $1$ from below, this gives
\begin{gather*}
    \alpha_t = 1-\tfrac{1}{t},\,\,\, \alpha_t = 1-\tfrac{1}{\sqrt t}
    \;\Longrightarrow\; \sum_{t = 1}^\infty \frac{1 - \alpha_t}{t} < \infty,
    \\
    \alpha_t=1-\tfrac{1}{\ln (t+2)},\,\,\, \alpha_t=1-\tfrac{1}{\ln\ln (t + 16)}
    \;\Longrightarrow\;\sum_{t=1}^\infty \frac{1 - \alpha_t}{t} = \infty.
\end{gather*}
Thus, consistency holds in the last two cases, despite the fraction of real data tending to zero as $t \to \infty$. Note that all four sequences satisfy $\sup_{t \ge 1} \alpha_t=1$, so Lemma~\ref{lem: sup_t alpha_t < 1 => cond holds} covers none of them.

Finally, we consider the case $\alpha_t \equiv 1$, in which no real samples are added at any iteration. Then, $\sum_{t = 1}^\infty (1 - \alpha_t)(1 - \rho_t) = 0$, so the condition from Theorem~\ref{thm:unified_sharp_criterion} is not satisfied. Thus, the estimated parameters do not converge to the true parameters, consistent with existing results in the literature \citep{dohmatob2025strong}. In a purely replacement-based scheme, the expected squared 2-Wasserstein distance $W_2$ can diverge \citep{shumailov2024ai}.
In contrast, for our accumulation-based scheme, {\color{black} we next show that the estimators neither converge to the true parameters nor diverge. This is consistent with prior work showing that accumulating data keeps the error bounded \citep{gerstgrasser2024model, kazdan2024collapse}. Here, we provide explicit two-sided bounds in terms of $n_r$.} This highlights a fundamental difference between the two schemes.

\begin{proposition}[Bounded error at $\alpha = 1$] \label{prop: alpha = 1 perm bias} Under the setup of Section~\ref{sec:setup}, if $\alpha_t \equiv 1$, then %
\begin{align*}
    \frac{\Tr(\boldsymbol{\Sigma})}{n_r} &\le \mathbb{E}\bigl[\|\boldsymbol{\mu}^{(t)} - \boldsymbol{\mu}\|_2^2\bigr] \le \frac{2\Tr(\boldsymbol{\Sigma})}{n_r}
    \\
    \frac{\|\boldsymbol{\Sigma}\|_F}{n_r} &\le \mathbb{E}\bigl[\|\boldsymbol{\Sigma}^{(t)}-\boldsymbol{\Sigma}\|_F\bigr] \le \frac{2\Tr(\boldsymbol{\Sigma})}{n_r} + \frac{1}{\sqrt{n_r}} \sqrt{A_{\boldsymbol{\Sigma}} + \frac{3}{n_r}\exp\!\left(\frac{3}{n_r}\right)B_{\boldsymbol{\Sigma}}}
\end{align*}
for every $t \ge 1$, 
where $A_{\boldsymbol{\Sigma}} = 5\Tr(\boldsymbol{\Sigma})^2 + 3\|\boldsymbol{\Sigma}\|_F^2$ and $B_{\boldsymbol{\Sigma}} = 5\Tr(\boldsymbol{\Sigma})^2 + 4\|\boldsymbol{\Sigma}\|_F^2$.
\end{proposition}

{\color{black} \textbf{Proof idea.} For the mean, we use the recursion for $\boldsymbol{C}_t$ (Theorem~\ref{thm:CT}) to show that $\Tr(\boldsymbol{C}_t)$ is non-decreasing in $t$, starting from $\Tr(\boldsymbol{C}_1)=\Tr(\boldsymbol{\Sigma})/n_r$, and bounded. For the covariance, the lower bound follows from the bias $-\boldsymbol{C}_t$ (Theorem~\ref{thm:estimators_bias}), and the upper bound combines the bound on $\Tr(\boldsymbol{C}_t)$ with an upper bound on the error of the empirical second moment around the true mean.}

\subsubsection{Rate of convergence}
\label{sec:convergence_rate_sec}

We have shown that, for every fixed $\alpha < 1$, the estimators are consistent. This result concerns the limit of infinitely many iterations. In practical training scenarios, however, only a finite number of iterations is performed. We, therefore, ask: how quickly do the estimated parameters converge to the true parameters? {\color{black} Below, we establish the rate of convergence for the mean estimator. In Section~\ref{sec:empirical_verification}, we show similar behavior for the covariance estimator.}

The next result gives the exact convergence rate and reveals a sharp transition at $\alpha = \tfrac12$. 

\begin{theorem}[Rate of decay]
\label{prop:rate}
Assume 
the setup of Section~\ref{sec:setup}, 
$\alpha_t\equiv\alpha\in[0,1]$ and $n_b(t)\equiv n_b$. Then
\begin{equation}
\label{eq:rate}
    \mathbb{E}\bigl[\|\boldsymbol{\mu}^{(t)} - \boldsymbol{\mu}\|_2^2\bigr] \;\asymp\;
    \begin{cases}
    N_t^{-1}, & \alpha<\tfrac12,\\[3pt]
    N_t^{-1}\ln N_t, & \alpha=\tfrac12,\\[3pt]
    N_t^{-2(1-\alpha)}, & \alpha>\tfrac12,
    \end{cases}
\end{equation}
where $N_t=n_r+(t-1)n_b\asymp t$. The implied constants depend only on $\alpha,n_b,n_r$ and $\Tr(\boldsymbol{\Sigma})$.
\end{theorem}

{\color{black} \textbf{Proof idea.} We derive the asymptotic rate of each term in the closed form of $\boldsymbol{C}_t$ (Eq.~\eqref{eq:CT-closed} in the appendix) by comparing sums with integrals.} %

\textbf{Remark (interpretation).} For $\alpha<\tfrac12$, the estimator attains the parametric rate $N_t^{-1}$ of the mean of $N_t$ i.i.d.\ real samples: {\color{black} synthetic data inflates the constant but not the exponent.} At $\alpha=\tfrac12$ a phase transition occurs, and for $\alpha>\tfrac12$ the exponent $2(1-\alpha)$ decreases linearly to $0$ as $\alpha\to1$. Consistency thus survives for every $\alpha<1$, but becomes arbitrarily slow: halving the error requires multiplying $N_t$ by $2^{1/[2(1-\alpha)]}$, a factor of $32$ at $\alpha=0.9$ and of $10^{15}$ at $\alpha=0.99$. {\color{black} Equivalently, the pooled estimator is only as accurate as the mean of order $N_t^{2(1-\alpha)}$ i.i.d.\ real samples, a vanishing fraction of $N_t$.} At $\alpha=1$, the rate reads $N_t^{-2(1-\alpha)}=1$: the error does not vanish (Theorem~\ref{thm:unified_sharp_criterion}), but it does not diverge either (Proposition~\ref{prop: alpha = 1 perm bias}).

{\color{black}
\textbf{Remark (the value of identifying real samples).} The gain from perfect detection of synthetic samples, i.e., from keeping only the real samples, stays bounded for $\alpha<\tfrac12$, but grows like $N_t^{2\alpha-1}$ for $\alpha>\tfrac12$, i.e., faster as the fraction of generated samples increases. For large but finite $N_t$, it is largest at some $\alpha$ strictly between $0$ and $1$, i.e., when generated data dominates but real data is still being added in meaningful amounts (see Appendix~\ref{sec:detection_value} for more discussion).}

\subsection{Empirical verification}
\label{sec:empirical_verification}

In this section, we simulate the theoretical setup in order to further support our theoretical results and also gain more insights into the model, e.g., the rate of convergence of the covariance estimator.

{\color{black}
\textbf{Setup.}
We fix $d = 4$, $\boldsymbol{\mu} = (-1, -\tfrac13, \tfrac13, 1)^\top$, and
$[\boldsymbol{\Sigma}]_{ij} = s_i s_j 2^{-|i-j|}$ with $s = (1, \tfrac43, \tfrac53, 2)^\top$, so that $\Tr(\boldsymbol{\Sigma}) = 9.56$, $\|\boldsymbol{\Sigma}\|_F = 6.18$, and $\boldsymbol{\Sigma}$ has a condition number $9.58$. Additionally, we take $n_b(t) \equiv n_b = 10$ and $n_r = 20$. Two families of mixing fractions are used: a grid of $15$ constant values $\alpha \in \{0, 0.1, 0.2, 0.3, 0.4, 0.45, 0.5, 0.55, 0.6, 0.7, 0.8, 0.9, 0.95, 0.99, 1\}$, spanning the transition of Theorem~\ref{prop:rate} and both endpoints of
Theorem~\ref{thm:unified_sharp_criterion}, and the four sequences $\alpha_t = 1 - \tfrac{1}{t},\, 1 - \tfrac{1}{\sqrt{t}},\, 1 - \tfrac{1}{\ln(t+2)},\, 1 - \tfrac{1}{\ln\ln(t+16)}$,
all with $\sup_{t \ge 1} \alpha_t = 1$. Each new sample is drawn independently from the
mixture~\eqref{eq:mixture}, and in each iteration we update $(\boldsymbol{\mu}^{(t)}, \boldsymbol{\Sigma}^{(t)})$ according to the iterative estimation procedure. Each regime is run for $T=10^5$ iterations ($N_T \approx 10^6$). Unless stated otherwise, expectations are estimated by averages over $M=20{,}000$ independent runs, each drawing its own
$n_r$ initial real samples. {\color{black} At this $M$, the Monte-Carlo standard error of the averaged mean-square error is about $1\%$ of its value, below the resolution of the figures (see also Appendix~\ref{sec:empirical_verification_details}).}}

\begin{figure}[t]
  \centering
  \includegraphics[width=0.75\textwidth]{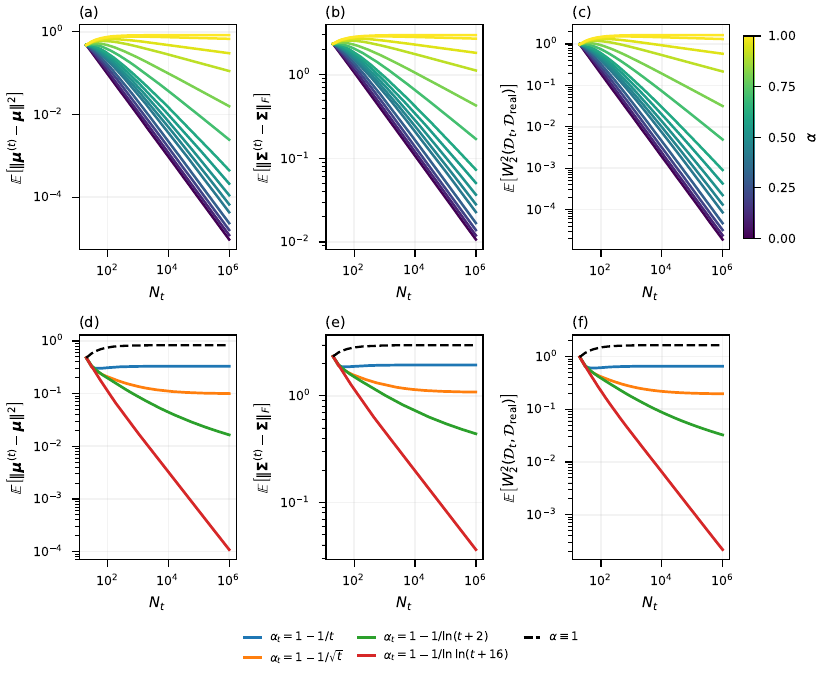}
  \vspace{-4mm}
  \caption{{\color{black} Convergence under constant and time-varying synthetic fractions. Top: constant $\alpha$, from $0$ (dark) to $1$ (light, see color bar): (a) $\mathbb{E}[\|\boldsymbol{\mu}^{(t)}-\boldsymbol{\mu}\|_2^2]$, (b) $\mathbb{E}[\|\boldsymbol{\Sigma}^{(t)}-\boldsymbol{\Sigma}\|_F]$, and (c) $\mathbb{E}[W_2^2(\mathcal{D}_t,\mathcal{D}_{\mathrm{real}})]$. Bottom: (d)--(f) the same quantities for $\alpha_t=1-\tfrac{1}{t}$, $1-\tfrac{1}{\sqrt{t}}$, $1-\tfrac{1}{\ln(t+2)}$ and $1-\tfrac{1}{\ln\ln(t+16)}$, with $\alpha\equiv1$ (dashed) for reference.}}
  \label{fig:alpha_convergence}

\vspace{3mm}

  \centering
  \includegraphics[width=0.75\textwidth]{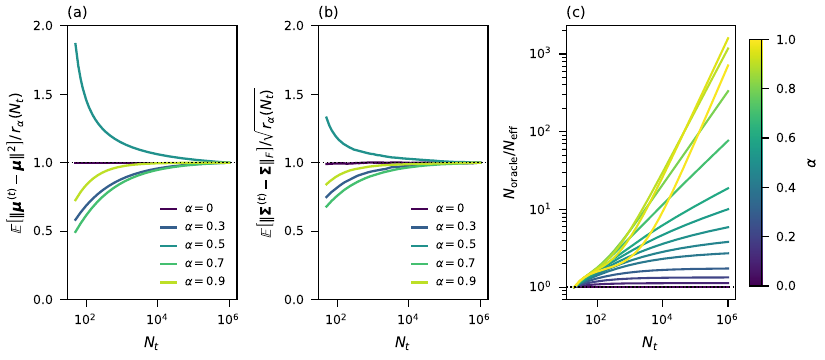}
  \vspace{-4mm}
  \caption{{\color{black} Rates of convergence and the effect of synthetic-sample detection. (a) $\mathbb{E}[\|\boldsymbol{\mu}^{(t)}-\boldsymbol{\mu}\|_2^2]/r_\alpha(N_t)$ and (b) $\mathbb{E}\bigl[\|\boldsymbol{\Sigma}^{(t)}-\boldsymbol{\Sigma}\|_F\bigr]/\sqrt{r_\alpha(N_t)}$, where $r_\alpha$ is the rate of Theorem~\ref{prop:rate}, for $\alpha\in\{0,0.3,0.5,0.7,0.9\}$. Each curve is scaled to equal $1$ at the largest $N_t$. (c) The gain from \emph{perfect} detection of synthetic samples, $N_{\mathrm{oracle}}/N_{\mathrm{eff}}$, for constant $\alpha\in[0,1)$ (see color bar).}} %
  \label{fig:rate_and_gain}
    \vspace{-4mm}
\end{figure}

{\color{black}
\textbf{Convergence at fixed $\alpha$.}
Every curve in Figure~\ref{fig:alpha_convergence} is a Monte-Carlo average over the
$M=20{,}000$ runs. The exception is $W_2^2$, computed from its closed form for Gaussians
(\eqref{eq:bures metric def} and~\eqref{eq:W2-split}): since each evaluation requires a matrix
square root, we average it over $10{,}000$ of the runs and evaluate it at $55$ logarithmically
spaced iterations rather than at every iteration. The top panels (a)--(c) illustrate Theorem~\ref{thm:unified_sharp_criterion} for constant $\alpha$: for every $0\le\alpha<1$, the errors of the mean (a), the covariance (b) and the fitted distribution (c) decrease with $N_t$, whereas at $\alpha\equiv1$, when no real data is added after the first iteration, they level off at a positive value
(Proposition~\ref{prop: alpha = 1 perm bias}). Thus, when the real-data fraction is fixed, consistency depends only on whether real data is present at all. Its proportion governs the rate instead (Theorem~\ref{prop:rate}): the decay slows as $\alpha$ grows, and becomes very slow as $\alpha\to1$. At $\alpha=0.99$ the mean-square error decays only like $N_t^{-0.02}$, too slowly to approach zero within the simulated horizon. %
For large $\alpha$, the mean-square error also first \emph{increases}, as predicted by {\color{black} an additional result (Proposition~\ref{prop:transient}) that we deferred to  Appendix~\ref{app:transient} due to space limitations}: above $\alpha^*(n_r,n_b) \approx 0.65$ it rises before it decreases. By the recursion of Theorem~\ref{thm:CT}, it rises by $25\%$ at $\alpha=0.9$ and by $62\%$ at $\alpha=0.99$, and by
$74\%$ at $\alpha=1$, where it never decreases.}

{\color{black}
\textbf{The sharpness of the criterion.} Panels (d)--(f) in
Figure~\ref{fig:alpha_convergence} show the four sequences mentioned above. 
All four have $\sup_{t \ge 1}\alpha_t=1$ and a vanishing fraction of real data, yet they split as predicted by~\eqref{eq:sharp critertion series sharp thm}: for $\alpha_t=1-1/t$ and $1-1/\sqrt{t}$, whose series converges, the errors approach strictly positive limits, as at $\alpha \equiv 1$, while for
$1-1/\ln(t+2)$ and $1-1/\ln\ln(t+16)$, whose series diverges, they keep decreasing. Thus, what matters is not whether the real-data fraction vanishes but how fast.}

{\color{black}
\textbf{Rate of convergence.}
Figure~\ref{fig:rate_and_gain}(a) divides $\mathbb{E}[\|\boldsymbol{\mu}^{(t)}-\boldsymbol{\mu}\|_2^2]$ by the rate $r_\alpha(N_t)$ stated in~\eqref{eq:rate}.
Here, the error is computed exactly by $\Tr(\boldsymbol{C}_t)$ (Equation~\eqref{eq:MSE-trace}) using the recursion of Theorem~\ref{thm:CT}. {\color{black} Appendix~\ref{sec:empirical_verification_details} (Figure~\ref{fig:mu_theory_validation}) confirms that these exact values agree with the simulation.} Each curve in Figure~\ref{fig:rate_and_gain}(a) is normalized by its own endpoint, so that flattening at $1$ confirms the predicted rate. All five values $\alpha\in\{0,0.3,0.5,0.7,0.9\}$ level off, on
both sides of the transition at $\alpha=\tfrac12$. At $\alpha=0$ the curve is flat for every $N_t$, since then $\boldsymbol{C}_t=\boldsymbol{\Sigma}/N_t$ exactly. Figure~\ref{fig:rate_and_gain}(b) repeats this for the covariance, dividing
$\mathbb{E}[\|\boldsymbol{\Sigma}^{(t)}-\boldsymbol{\Sigma}\|_F]$, estimated from the $M = 20{,}000$ Monte-Carlo runs, by $\sqrt{r_\alpha(N_t)}$, the square root because the Frobenius error is a distance while the mean-square error in (a) is a squared distance. The same five curves level off, so empirically the covariance converges at the rate of the mean, including the transition at $\alpha=\tfrac12$, although Theorem~\ref{prop:rate} covers only the mean.}

{\color{black}
\textbf{The value of identifying real samples.} Figure~\ref{fig:rate_and_gain}(c) shows the gain $G_\alpha(N_t)=N_{\mathrm{oracle}}(N_t)/N_{\mathrm{eff}}(N_t)$ from perfect detection of synthetic samples. 
Here, $N_{\mathrm{eff}}=\Tr(\boldsymbol{\Sigma})/\mathbb{E}\bigl[\|\boldsymbol{\mu}^{(t)}-\boldsymbol{\mu}\|_2^2\bigr]$ is the number of real i.i.d.\ samples that would give the same mean-square error, and $N_{\mathrm{oracle}}=(1-\alpha)N_t+\alpha n_r$ is the expected number of real samples in the pool (discussed more in Appendix~\ref{sec:detection_value}). 
$N_{\mathrm{eff}}$ is computed exactly using the recursion of Theorem~\ref{thm:CT}. The gain saturates for $\alpha<\tfrac12$ and grows like $N_t^{2\alpha-1}$ for $\alpha>\tfrac12$, reaching about $10^3$ at $N_t=10^6$ for $\alpha\in[0.9,0.95]$. At $\alpha=0.99$ it is smaller, since the oracle also runs out of real data as $\alpha\to1$.}

\section{Experiments with Diffusion Models}
\label{sec:empirical}
Our paper focuses on theoretical contributions. 
Yet, to demonstrate that insights generalize beyond the theoretical setup, in Appendix~\ref{app:empirical} we present results of iterative training of a DDPM-based diffusion model \citep{ho2020denoising}, on the CIFAR-10 dataset \citep{krizhevsky2009learning}.
We examine the effect of different fixed values of $\alpha$, with no distinction between real and generated data, on the Fr\'echet Inception Distance (FID) quality metric. %
Aligning with our theory: 1) for every $\alpha<1$, the FID improves upon its
initial value at every iteration, although real and generated images are pooled indistinguishably; %
2) for $\alpha=1$, the FID never improves but remains bounded; %
3) the FID is ordered by $\alpha$ at every
iteration, and its improvement slows as $\alpha$ grows; %
4) for $\alpha=0.25<\tfrac12$, the curve follows
that of $\alpha=0$ up to a roughly constant offset, i.e., synthetic data
inflates the constant but not the rate; %
and 5)
discarding the generated images would improve the final FID far more for
$\alpha=0.75>\tfrac12$ than for $\alpha=0.25<\tfrac12$. %
An apparent gap between the theoretical setup and the diffusion experiments is that, for a large $\alpha<1$, the FID stops improving at some $t$. This may result from the simple training strategy used (which could be improved) and the relatively simple diffusion model (which could be scaled up).

\section{Conclusion}
\label{sec:conclusion}

{\color{black}
Large generative models are commonly trained on data scraped from the web, an increasing share of which is generated by earlier models. Prior work shows that iterative training degrades performance: 1) without bound when earlier data is discarded, and 2) in a bounded manner when data accumulates. The continued improvement of generative models has been explained by the availability of fresh real data, but only under the unrealistic assumption that it can be perfectly separated from generated data. In this paper, we studied the more realistic setting in which fresh real and generated samples are pooled indistinguishably. We showed, for a tractable Gaussian data model, that the estimated distribution converges to the true one if and only if the share of fresh real information (quantified exactly) is not summable. Our results reveal that consistency holds for any fixed positive fraction of real data, and even for a fraction that vanishes slowly enough. Yet, the convergence rate of the mean estimator deteriorates as generated samples become more prevalent: it matches the rate of purely real data as long as generated samples are a minority of the newly collected data, and becomes arbitrarily slow once they are a majority. This motivates ongoing research on detecting and curating generated samples, and indicates when such efforts are worthwhile: as more data is collected, the gain from perfect detection is bounded when generated samples are a minority of the newly collected data, but grows without bound with the amount of collected data otherwise.
\textcolor{black}{Finally, while our work focuses on iterative training of generative models from a distribution estimation perspective, a compelling direction for future work is to explore whether even biased models can still generate augmented data that is useful for downstream tasks, similar to how low-level pre-processing can benefit downstream classification despite idealized information-theoretic principles \citep{turgeman2026does}.}
\subsubsection*{Acknowledgments}
The work is supported by the Israel Science Foundation
grant No.~1940/23 and the MOST grant No.~0007091.

\bibliography{refs}
\bibliographystyle{iclr2027_conference}

\newpage 

\appendix
\section{Proofs}
\label{sec:app}

Section~\ref{sec:scalar_seq} defines three scalar sequences, which are used widely in the proofs.

\subsection{Scalar sequences}
\label{sec:scalar_seq}

For $t \ge 1$, we define three scalar sequences:
\begin{equation}\label{eq:scalars}
  \lambda_t \coloneqq \frac{N_t+n_b(t)\alpha_t}{N_{t+1}},\qquad
  \beta_t \coloneqq \frac{n_b(t)}{N_{t+1}^2},\qquad
  \gamma_t \coloneqq \lambda_t^2-\alpha_t^2\beta_t.
\end{equation}

Additionally, let us define
\begin{equation}
\label{eq: Gamma_(s,t) def}
    \Gamma_{s,t} \coloneqq 
    \begin{cases}
        \prod_{k=s}^t \gamma_k \qquad s\le t
        \\
        1 \qquad s > t
    \end{cases}.
\end{equation}

Section~\ref{sec:technical_lemmas} collects the technical lemmas used throughout.

\subsection{Technical Lemmas}
\label{sec:technical_lemmas}

{\color{black}
We first formalize the conditioning used throughout the proofs.

\begin{definition}[Natural filtration]
\label{def:filtration}
For every $t\ge1$, let
\begin{equation}
\label{eq: Ft definition}
    \mathcal{F}_t \coloneqq \sigma\Bigl(
      \bigl\{\boldsymbol{x}^{(0)}_j\bigr\}_{j=1}^{n_r},\;
      \bigl\{\boldsymbol{x}^{(s)}_i\bigr\}_{1\le i\le n_b(s),\; 1\le s\le t-1}
    \Bigr)
\end{equation}
denote the $\sigma$-algebra generated by all samples drawn before the batch $\{\boldsymbol{x}^{(t)}_i\}_{i=1}^{n_b(t)}$. For $t=1$, the second family is empty, so $\mathcal{F}_1$ is generated by the original real batch alone.
\end{definition}

Since each iteration only adds samples, $\mathcal{F}_1\subseteq\mathcal{F}_2\subseteq\cdots$ is a filtration, and $(\boldsymbol{\mu}^{(t)},\boldsymbol{\Sigma}^{(t)})$ is $\mathcal{F}_t$-measurable by~\eqref{eq:pooled-def}. Hence, Assumption~\ref{ass:cond-ind} is equivalent to the following: for every $t\ge1$, conditionally on $\mathcal{F}_t$, the samples $\{\boldsymbol{x}^{(t)}_i\}_{i=1}^{n_b(t)}$ are i.i.d.\ from the mixture~\eqref{eq:mixture} with parameters $(\boldsymbol{\mu}^{(t)},\boldsymbol{\Sigma}^{(t)})$. This is the form used throughout the proofs.}

Next, we record the conditional first and second moments of a newly drawn sample. These
are used in the proof of Theorems~\ref{thm:estimators_bias} and \ref{thm:CT}.

\begin{lemma}[Conditional moments of the mixture samples]
\label{lem:cond-moments}
Under Assumption~\ref{ass:cond-ind}, conditional on $\mathcal{F}_t$, each newly drawn sample $\boldsymbol{x}^{(t)}_i$ $(t\ge1)$ satisfies, using \eqref{eq:delta-def} $(\boldsymbol{\delta}^{(t)} = \boldsymbol{\mu}^{(t)} - \boldsymbol{\mu})$:
\begin{align}
  \mathbb{E}\bigl[\boldsymbol{x}^{(t)}_i\mid\mathcal{F}_t\bigr] &= \alpha_t\boldsymbol{\mu}^{(t)}+(1-\alpha_t)\boldsymbol{\mu} = \boldsymbol{\mu} +\alpha_t\boldsymbol{\delta}^{(t)},
  \label{eq:cond-mean}\\[2pt]
  \Cov\bigl(\boldsymbol{x}^{(t)}_i\mid\mathcal{F}_t\bigr) &= \alpha_t\boldsymbol{\Sigma}^{(t)}+(1-\alpha_t)\boldsymbol{\Sigma} +\alpha_t(1-\alpha_t)\,\boldsymbol{\delta}^{(t)}(\boldsymbol{\delta}^{(t)})^\top, 
  \label{eq:cond-cov}\\[2pt]
  \mathbb{E}\bigl[(\boldsymbol{x}^{(t)}_i-\boldsymbol{\mu})(\boldsymbol{x}^{(t)}_i-\boldsymbol{\mu})^\top\mid\mathcal{F}_t\bigr] &= \alpha_t\boldsymbol{\Sigma}^{(t)}+(1-\alpha_t)\boldsymbol{\Sigma} +\alpha_t\,\boldsymbol{\delta}^{(t)} (\boldsymbol{\delta}^{(t)})^\top.
  \label{eq:cond-sq}
\end{align}
\end{lemma}

\begin{proof}
Given $\mathcal{F}_t$, the sample $\boldsymbol{x}^{(t)}_i$ is drawn from the two-component
mixture~\eqref{eq:mixture} with weights $\alpha_t$ on
$\mathcal{N}(\boldsymbol{\mu}^{(t)},\boldsymbol{\Sigma}^{(t)})$ and $1-\alpha_t$ on
$\mathcal{N}(\boldsymbol{\mu},\boldsymbol{\Sigma})$.

\smallskip\noindent\textit{Equation~\eqref{eq:cond-mean}.} By linearity, the mixture mean is
$\alpha_t\boldsymbol{\mu}^{(t)}+(1-\alpha_t)\boldsymbol{\mu}=\boldsymbol{\mu}+\alpha_t\boldsymbol{\delta}^{(t)}$.

\smallskip\noindent\textit{Equation~\eqref{eq:cond-cov}.} Let $Z\in\{1,2\}$ be the
component indicator with $\mathbb{P}(Z=1\mid\mathcal{F}_t)=\alpha_t$, so that
$\boldsymbol{x}^{(t)}_i\mid\{Z=1\}\sim\mathcal{N}(\boldsymbol{\mu}^{(t)},\boldsymbol{\Sigma}^{(t)})$ and
$\boldsymbol{x}^{(t)}_i\mid\{Z=2\}\sim\mathcal{N}(\boldsymbol{\mu},\boldsymbol{\Sigma})$. The law of total
covariance gives
\[
  \Cov\bigl(\boldsymbol{x}^{(t)}_i\mid\mathcal{F}_t\bigr)
  =\underbrace{\mathbb{E}\bigl[\Cov(\boldsymbol{x}^{(t)}_i\mid Z,\mathcal{F}_t)\mid\mathcal{F}_t\bigr]}_A
  +\underbrace{\Cov\bigl(\mathbb{E}[\boldsymbol{x}^{(t)}_i\mid Z,\mathcal{F}_t]\mid\mathcal{F}_t\bigr)}_{B}.
\]
The first summand $A$ reads:
\begin{equation}
\label{First summand in cond-cov}
    \Cov(\boldsymbol{x}^{(t)}_i\mid Z,\mathcal{F}_t) = \begin{cases}
        \boldsymbol{\Sigma}^{(t)} & Z = 1 \\ \boldsymbol{\Sigma} & Z = 2
    \end{cases} \Rightarrow A = \alpha_t\boldsymbol{\Sigma}^{(t)}+(1-\alpha_t)\boldsymbol{\Sigma}.
\end{equation}
We now compute the second summand $B$. We have
\begin{equation*}
    \mathcal{P}(Z) \coloneqq \mathbb{E}[\boldsymbol{x}^{(t)}_i\mid Z,\mathcal{F}_t] = \begin{cases}
        \boldsymbol{\mu}^{(t)} & Z=1 \\ \boldsymbol{\mu} & Z=2
    \end{cases} \Rightarrow B = \mathbb{E}\bigl[\left(\mathcal{P}(Z) - \mu_{\mathcal{P}}\right)\left(\mathcal{P}(Z) - \mu_{\mathcal{P}}\right)^\top \mid \mathcal{F}_t\bigr]
\end{equation*}
where
\begin{equation*}
    \mu_{\mathcal{P}} = \mathbb{E}[\mathcal{P}(Z) \mid \mathcal{F}_t] = \alpha_t\boldsymbol{\mu}^{(t)} + (1-\alpha_t)\boldsymbol{\mu}.
\end{equation*}
That is, $B = \mathbb{E}\left[\mathcal{P}_c(Z) \mid \mathcal{F}_t\right]$, where
\begin{align*}
    &\left(\mathcal{P}(Z) - \mu_{\mathcal{P}}\right) = \begin{cases}
    (1-\alpha_t)(\boldsymbol{\mu}^{(t)} - \boldsymbol{\mu}) & Z = 1 \\ -\alpha_t(\boldsymbol{\mu}^{(t)} - \boldsymbol{\mu}) & Z=2
    \end{cases} 
    = \begin{cases}
    (1-\alpha_t)\boldsymbol{\delta}^{(t)} & Z = 1 \\ -\alpha_t\boldsymbol{\delta}^{(t)} & Z=2
    \end{cases}
    \\
    \\
    &\mathcal{P}_c(Z) = \left(\mathcal{P}(Z) - \mu_{\mathcal{P}}\right)
    \left(\mathcal{P}(Z) - \mu_{\mathcal{P}}\right)^\top = \begin{cases}
    (1-\alpha_t)^2\boldsymbol{\delta}^{(t)}(\boldsymbol{\delta}^{(t)})^\top & Z = 1 \\ \alpha_t^2\boldsymbol{\delta}^{(t)}(\boldsymbol{\delta}^{(t)})^\top & Z=2
    \end{cases}
\end{align*}
Finally, the second part reads
\[
  B=\alpha_t\bigl((1-\alpha_t)\boldsymbol{\delta}^{(t)}\bigr)\bigl((1-\alpha_t)\boldsymbol{\delta}^{(t)}\bigr)^\top
   +(1-\alpha_t)\bigl(\alpha_t\boldsymbol{\delta}^{(t)}\bigr)\bigl(\alpha_t\boldsymbol{\delta}^{(t)}\bigr)^\top
   =\alpha_t(1-\alpha_t)\,\boldsymbol{\delta}^{(t)}(\boldsymbol{\delta}^{(t)})^\top,
\]
using $\alpha_t(1-\alpha_t)^2+(1-\alpha_t)\alpha_t^2=\alpha_t(1-\alpha_t)$. Adding the two
terms yields~\eqref{eq:cond-cov}.

\smallskip\noindent\textit{Equation~\eqref{eq:cond-sq}.} For any random vector $\boldsymbol{x}$
and fixed $\boldsymbol{\mu}$,
$\mathbb{E}[(\boldsymbol{x}-\boldsymbol{\mu})(\boldsymbol{x}-\boldsymbol{\mu})^\top]=\Cov(\boldsymbol{x})+(\mathbb{E}[\boldsymbol{x}]-\boldsymbol{\mu})(\mathbb{E}[\boldsymbol{x}]-\boldsymbol{\mu})^\top$.
Applying this conditionally on $\mathcal{F}_t$ with
$\mathbb{E}[\boldsymbol{x}^{(t)}_i\mid\mathcal{F}_t]-\boldsymbol{\mu}=\alpha_t\boldsymbol{\delta}^{(t)}$ and
substituting~\eqref{eq:cond-cov},
\[
  \alpha_t\boldsymbol{\Sigma}^{(t)}+(1-\alpha_t)\boldsymbol{\Sigma}
   +\alpha_t(1-\alpha_t)\boldsymbol{\delta}^{(t)}(\boldsymbol{\delta}^{(t)})^\top
   +\alpha_t^2\boldsymbol{\delta}^{(t)}(\boldsymbol{\delta}^{(t)})^\top
  =\alpha_t\boldsymbol{\Sigma}^{(t)}+(1-\alpha_t)\boldsymbol{\Sigma}+\alpha_t\boldsymbol{\delta}^{(t)}(\boldsymbol{\delta}^{(t)})^\top,
\]
since $\alpha_t(1-\alpha_t)+\alpha_t^2=\alpha_t$.
\end{proof}

Next, we derive a more useful representation of $\boldsymbol{\Sigma}^{(t)}$ that is centered at $\boldsymbol{\mu}$. It
is used in the proof of Theorems~\ref{thm:estimators_bias} and~\ref{thm:Frobenius norm consistency of Sigma}.

\begin{lemma}[Pooled scatter identity] 
\label{lem:pooled_scatter_identity} Fix $t \ge 1$. The estimators $\boldsymbol{\mu}^{(t)}, \boldsymbol{\Sigma}^{(t)}$ defined in \eqref{eq:pooled-def} satisfy, using \eqref{eq:delta-def} $(\boldsymbol{\delta}^{(t)} = \boldsymbol{\mu}^{(t)} - \boldsymbol{\mu})$:
\begin{equation}
\label{eq: pooled scatter identity}
    \boldsymbol{\Sigma}^{(t)} = \frac{1}{N_t} \left( \sum_{j=1}^{n_r} (\boldsymbol{x}^{(0)}_j -\boldsymbol{\mu})(\boldsymbol{x}^{(0)}_j -\boldsymbol{\mu})^\top + \sum_{s=1}^{t-1}\sum_{i=1}^{n_b(s)} (\boldsymbol{x}^{(s)}_i - \boldsymbol{\mu})(\boldsymbol{x}^{(s)}_i - \boldsymbol{\mu})^\top\right) - \boldsymbol{\delta}^{(t)}(\boldsymbol{\delta}^{(t)})^\top.
\end{equation}
\end{lemma}

\begin{proof}

By~\eqref{eq:pooled-def}, we have 
\begin{equation*}
    \sum_{k=1}^{N_t} \boldsymbol{y}_k = N_t\boldsymbol{\mu}^{(t)}.
\end{equation*}
Thus,
\begin{equation*}
    \sum_{k=1}^{N_t} (\boldsymbol{y}_k - \boldsymbol{\mu}) = N_t\boldsymbol{\mu}^{(t)} - N_t\boldsymbol{\mu} = N_t\boldsymbol{\delta}^{(t)}.
\end{equation*}
Expanding the defining sum in~\eqref{eq:pooled-def}, we get
\begin{align*}
    N_t\boldsymbol{\Sigma}^{(t)} &= \sum_{k=1}^{N_t} \left(\boldsymbol{y}_k - \boldsymbol{\mu}^{(t)}\right)\left(\boldsymbol{y}_k - \boldsymbol{\mu}^{(t)}\right)^\top 
    \\
    &= \sum_{k=1}^{N_t} \left(\left(\boldsymbol{y}_k - \boldsymbol{\mu}\right) - \boldsymbol{\delta}^{(t)}\right)\left(\left(\boldsymbol{y}_k - \boldsymbol{\mu}\right) - \boldsymbol{\delta}^{(t)}\right)^\top
    \\
    &= \sum_{k=1}^{N_t} \left(\boldsymbol{y}_k - \boldsymbol{\mu}\right)\left(\boldsymbol{y}_k - \boldsymbol{\mu}\right)^\top - \sum_{k=1}^{N_t} (\boldsymbol{y}_k - \boldsymbol{\mu})(\boldsymbol{\delta}^{(t)})^\top - \boldsymbol{\delta}^{(t)} \sum_{k=1}^{N_t} (\boldsymbol{y}_k - \boldsymbol{\mu})^\top + N_t\boldsymbol{\delta}^{(t)}(\boldsymbol{\delta}^{(t)})^\top
    \\
    &= \sum_{k=1}^{N_t} (\boldsymbol{y}_k - \boldsymbol{\mu})(\boldsymbol{y}_k - \boldsymbol{\mu})^\top - N_t\cdot \boldsymbol{\delta}^{(t)}(\boldsymbol{\delta}^{(t)})^\top - N_t\cdot \boldsymbol{\delta}^{(t)}(\boldsymbol{\delta}^{(t)})^\top + N_t\cdot \boldsymbol{\delta}^{(t)}(\boldsymbol{\delta}^{(t)})^\top
    \\
    &= \sum_{k=1}^{N_t} (\boldsymbol{y}_k - \boldsymbol{\mu})(\boldsymbol{y}_k - \boldsymbol{\mu})^\top - N_t\cdot \boldsymbol{\delta}^{(t)}(\boldsymbol{\delta}^{(t)})^\top.
\end{align*}
Dividing both sides by $N_t$, we get the required identity: 
\begin{equation*}
    \boldsymbol{\Sigma}^{(t)} = \frac{1}{N_t}\left(\sum_{j=1}^{n_r} (\boldsymbol{x}^{(0)}_j -\boldsymbol{\mu})(\boldsymbol{x}^{(0)}_j -\boldsymbol{\mu})^\top + \sum_{s=1}^{t-1}\sum_{i=1}^{n_b(s)} (\boldsymbol{x}^{(s)}_i - \boldsymbol{\mu})(\boldsymbol{x}^{(s)}_i - \boldsymbol{\mu})^\top\right) - \boldsymbol{\delta}^{(t)}(\boldsymbol{\delta}^{(t)})^\top.
\end{equation*}

\end{proof}

{\color{black}
Next, we record an elementary consequence of the Cauchy-Schwarz inequality, used in the proofs of Lemmas~\ref{lem:trace-frobenius}, \ref{lem:bures-frobenius}, and~\ref{lem:uniform-L2-Shat}.

\begin{lemma}[$\ell_1$--$\ell_2$ comparison]
\label{lem:l1-l2}
Let $n\ge1$ and $a_1,\dots,a_n\in\mathbb{R}$. Then
\begin{equation}
\label{eq:l1-l2}
    \sum_{i=1}^n |a_i| \le \sqrt{n} \, \left(\sum_{i=1}^n a_i^2\right)^{1/2},
    \qquad \text{and consequently} \qquad
    \left(\sum_{i=1}^n a_i\right)^2 \le n\sum_{i=1}^n a_i^2.
\end{equation}
\end{lemma}

\begin{proof}
Write $\boldsymbol{a} \coloneqq (|a_1|,\dots,|a_n|)^\top$ and $\boldsymbol{1}_n \coloneqq (1,\dots,1)^\top$. The Cauchy--Schwarz inequality gives
\begin{equation*}
    \sum_{i=1}^n |a_i|
    = \langle\boldsymbol{a},\boldsymbol{1}_n\rangle
    \le \|\boldsymbol{a}\|_2\,\|\boldsymbol{1}_n\|_2
    = \sqrt{n}\,\left(\sum_{i=1}^n a_i^2\right)^{1/2},
\end{equation*}
which is the first inequality. For the second, the triangle inequality gives $\bigl|\sum_{i=1}^n a_i\bigr| \le \sum_{i=1}^n |a_i|$. Both sides are nonnegative, so squaring and applying the first inequality yields
\begin{equation*}
    \left(\sum_{i=1}^n a_i\right)^2 \le \left(\sum_{i=1}^n |a_i|\right)^2 \le n\sum_{i=1}^n a_i^2 .
\end{equation*}
\end{proof}}

We now record an elementary fact about positive-semi-definite matrices. It is used in the proofs of Lemma~\ref{lem: CT->0 <=> mu(t)-> mu in L2}, Theorems~\ref{thm:NS-theorem-for-CT->0} and \ref{thm:Frobenius norm consistency of Sigma}, and Lemma~\ref{lem:fourth-moment}.

\begin{lemma}[Trace--Frobenius comparison]
\label{lem:trace-frobenius}
Let $\boldsymbol{A}\in\mathbb{R}^{d\times d}$ be symmetric with
$\boldsymbol{A}\succeq0$. Then:
\begin{enumerate}
  \item[(a)] $\|\boldsymbol{A}\|_F\le\Tr(\boldsymbol{A})\le\sqrt{d}\,\|\boldsymbol{A}\|_F$. In particular, $\|\boldsymbol{A}\|_F^2\le\Tr(\boldsymbol{A})^2$.
  \item[(b)] If $\{\boldsymbol{A}_t\}_{t\ge1}\subset\mathbb{R}^{d\times d}$ are symmetric
    and positive-semi-definite, then
    \begin{equation}
    \label{eq: frobenius iff trace}
      \|\boldsymbol{A}_t\|_F\xrightarrow{t\to\infty}0
      \quad\Longleftrightarrow\quad
      \Tr(\boldsymbol{A}_t)\xrightarrow{t\to\infty}0 .
    \end{equation}
\end{enumerate}
\end{lemma}

\begin{proof}
\emph{(a)} By the spectral theorem, $\boldsymbol{A}=\boldsymbol{Q}\boldsymbol{D}\boldsymbol{Q}^\top$
with $\boldsymbol{Q}$ orthogonal and
$\boldsymbol{D}=\operatorname{diag}(\nu_1,\dots,\nu_d)$, where $\nu_1,\dots,\nu_d\ge0$ are the
eigenvalues of $\boldsymbol{A}$ (real by symmetry, nonnegative since
$\boldsymbol{A}\succeq0$). As the trace and the Frobenius norm are invariant under
orthogonal conjugation,
\begin{equation}
\label{eq: trace and frobenius via eigenvalues}
  \Tr(\boldsymbol{A})=\sum_{i=1}^{d}\nu_i,
  \qquad
  \|\boldsymbol{A}\|_F^2=\sum_{i=1}^{d}\nu_i^2 .
\end{equation}

\smallskip\noindent\emph{Lower bound.} Expanding the square and discarding the
cross terms, which are nonnegative because every $\nu_i\ge0$,
\begin{equation*}
  \Tr(\boldsymbol{A})^2=\!\left(\sum_{i=1}^{d}\nu_i\right)^2
  =\sum_{i=1}^{d}\nu_i^2+\sum_{i\ne j}\nu_i\nu_j
  \;\ge\;\sum_{i=1}^{d}\nu_i^2=\|\boldsymbol{A}\|_F^2 .
\end{equation*}
Both $\Tr(\boldsymbol{A})$ and $\|\boldsymbol{A}\|_F$ are nonnegative, so taking square
roots gives $\|\boldsymbol{A}\|_F\le\Tr(\boldsymbol{A})$, together with the squared form
$\|\boldsymbol{A}\|_F^2\le\Tr(\boldsymbol{A})^2$.

\smallskip\noindent\emph{Upper bound.} Since every $\nu_i\ge0$, Lemma~\ref{lem:l1-l2} gives
\begin{equation*}
  \Tr(\boldsymbol{A})=\sum_{i=1}^{d}\nu_i \le\sqrt{d}\,\left(\sum_{i=1}^{d}\nu_i^2\right)^{1/2} =\sqrt{d}\,\|\boldsymbol{A}\|_F.
\end{equation*}

\smallskip\noindent\emph{(b)} Apply~(a) to each $\boldsymbol{A}_t$:
\begin{equation*}
  \|\boldsymbol{A}_t\|_F\le\Tr(\boldsymbol{A}_t)\le\sqrt{d}\,\|\boldsymbol{A}_t\|_F ,
  \qquad t\ge1 .
\end{equation*}
The constants $1$ and $\sqrt{d}$ do not depend on $t$, so the two sequences are
squeezed to $0$ together, which is~\eqref{eq: frobenius iff trace}.
\end{proof}

Next, we prove that the sequence $\beta_t$, defined in \eqref{eq:scalars}, is summable. It is used in the proof of Theorems~\ref{thm:sharp-criterion-mu}, \ref{prop:rate}, and \ref{thm:Frobenius norm consistency of Sigma}.

\begin{lemma}[Summability of $\beta_s$]\label{lem:beta-summability}
The weights $\beta_s$ from~\eqref{eq:scalars}
satisfy
\begin{equation*}
    \sum_{s=1}^{\infty}\beta_s \le \frac{1}{n_r}.
\end{equation*}
\end{lemma}

\begin{proof}
Since $N_{s+1}=N_s+n_b(s)\ge N_s$ from \eqref{eq:counts},
\begin{equation*}
  \beta_s=\frac{n_b(s)}{N_{s+1}^2}=\frac{N_{s+1}-N_s}{N_{s+1}^2}
    \le\frac{N_{s+1}-N_s}{N_sN_{s+1}}=\frac{1}{N_s}-\frac{1}{N_{s+1}},
\end{equation*}
where we used the definition of $\beta_s$ from~\eqref{eq:scalars}. Summing telescopes:
\begin{equation*}
    \sum_{s=1}^{t-1}\beta_s\le\frac{1}{N_1}-\frac{1}{N_t}<\frac{1}{N_1}=\frac{1}{n_r}.
\end{equation*}
Taking the limit $t \to \infty$, we get $\sum_{s=1}^\infty\beta_s\le1/n_r$.
\end{proof}

Next, we prove a lemma that is used in the proof of Theorem~\ref{thm:sharp-criterion-mu} and Lemma~\ref{lem:diverging_series}.

\begin{lemma}[Product-decay criterion]\label{lem:product-decay}
Let $(r_s)_{s \ge 1}\subset (0,1)$. If $\prod_{s=1}^{t} r_s \to 0$ as $t \to \infty$, then
$\sum_{s=1}^{\infty}(1-r_s)=\infty$.
\end{lemma}

\begin{proof}
We prove the contrapositive: if $\sum_{s=1}^\infty(1-r_s)<\infty$, then $\prod_{s=1}^t r_s$ has a positive
limit (as $1-r_s>0$, the series either
converges or diverges to $+\infty$, so this is indeed the negation of the
conclusion).

Assume $\sum_{s=1}^\infty (1-r_s) < \infty$. Then, $1 - r_s \to 0$, so there is $s_0 \ge 1$ with $r_s \ge \tfrac 12$ for $s \ge s_0$. 

In addition, for $x \in [\tfrac12,1)$, the function $g(x) \coloneqq 2(1-x)+\ln x$ is decreasing: $g'(x)=1/x-2 \le 0$, and $g(1)=0$. Thus, $g \ge 0$ on $[\tfrac12,1)$, i.e., $-\ln x\le2(1-x)$.

Finally, for $s \ge s_0$, we have $r_s \ge \tfrac 12$, so $\sum_{s \ge s_0}(-\ln r_s)\le2\sum_{s \ge s_0}(1-r_s)<\infty$. That is,
$\sum_{s\ge1}(-\ln r_s)<\infty$ (the finitely many remaining terms are finite as $r_s > 0$) and
$\prod_{s=1}^{t}r_s=\exp\!\bigl(-\sum_{s=1}^{t}(-\ln r_s)\bigr)$ converges to a strictly positive limit, as claimed.
\end{proof}

We now provide more useful expressions concerning the estimators $\boldsymbol{\mu}^{(t)}$ and $\boldsymbol{\Sigma}^{(t)}$.

\begin{theorem}[Estimators formulas] 
\label{thm:iterative-estimators-formula}
Fix $t \ge 2$ and consider the setup in Section~\ref{sec:setup}. Then:
\begin{enumerate}
  \item[(a)] The mean obeys the recursion
  \begin{equation}\label{eq:mu-update}
    \boldsymbol{\mu}^{(t)} = \rho_{t - 1}\boldsymbol{\mu}^{(t-1)}
      +\frac{1}{N_t}\sum_{i=1}^{n_b(t-1)}\boldsymbol{x}^{(t-1)}_i.
  \end{equation}
  \item[(b)] Splitting the pool into the initial real batch and the subsequent batches, the covariance reads
  \begin{equation}\label{eq:Sigma-update}
    \boldsymbol{\Sigma}^{(t)} = \frac{1}{N_t}
      \Biggl(\,\sum_{j=1}^{n_r}\bigl(\boldsymbol{x}^{(0)}_j-\boldsymbol{\mu}^{(t)}\bigr)
      \bigl(\boldsymbol{x}^{(0)}_j-\boldsymbol{\mu}^{(t)}\bigr)^\top
      +\sum_{s=1}^{t-1}\sum_{i=1}^{n_b(s)}
      \bigl(\boldsymbol{x}^{(s)}_i-\boldsymbol{\mu}^{(t)}\bigr)\bigl(\boldsymbol{x}^{(s)}_i-\boldsymbol{\mu}^{(t)}\bigr)^\top
      \Biggr).
  \end{equation}
\end{enumerate}
\end{theorem}

\textbf{Remark.} Equation~\eqref{eq:mu-update} expresses the updated mean as a weighted average of the previous estimate and the new batch mean. Equation~\eqref{eq:Sigma-update} is the pooled sample covariance centered at the estimated mean $\boldsymbol{\mu}^{(t)}$.

\begin{proof}

\noindent\textbf{Step 1 (part (a): mean recursion~\eqref{eq:mu-update}).}
Split the most recent batch $s=t-1$ off from the definition~\eqref{eq:pooled-def} of
$\boldsymbol{\mu}^{(t)}$:
\begin{equation*}
    \boldsymbol{\mu}^{(t)}
    =\frac{1}{N_t}\Biggl(\sum_{j=1}^{n_r}\boldsymbol{x}^{(0)}_j
    +\sum_{s=1}^{t-2}\sum_{i=1}^{n_b(s)}\boldsymbol{x}^{(s)}_i\Biggr)
    +\frac{1}{N_t}\sum_{i=1}^{n_b(t-1)}\boldsymbol{x}^{(t-1)}_i.
\end{equation*}
The parenthesized sum is exactly the pool of the first $N_{t-1}=n_r+\sum_{s=1}^{t-2}n_b(s)$
points, so applying the definition~\eqref{eq:pooled-def} at iteration $t-1$—which
for $t=2$ is precisely~\eqref{eq:init}—gives
$\sum_{j=1}^{n_r}\boldsymbol{x}^{(0)}_j+\sum_{s=1}^{t-2}\sum_{i=1}^{n_b(s)}\boldsymbol{x}^{(s)}_i
=N_{t-1}\,\boldsymbol{\mu}^{(t-1)}$. With $\rho_{t - 1} = N_{t-1}/N_t$ from~\eqref{eq:counts},
\begin{equation*}
    \boldsymbol{\mu}^{(t)}
    =\frac{N_{t-1}}{N_t}\,\boldsymbol{\mu}^{(t-1)}
    +\frac{1}{N_t}\sum_{i=1}^{n_b(t-1)}\boldsymbol{x}^{(t-1)}_i
    =\rho_{t - 1}\,\boldsymbol{\mu}^{(t-1)}
    +\frac{1}{N_t}\sum_{i=1}^{n_b(t-1)}\boldsymbol{x}^{(t-1)}_i,
\end{equation*}
which is~\eqref{eq:mu-update}.

\medskip
\noindent\textbf{Step 2 (part (b)).}
Relabeling the $N_t$ points in~\eqref{eq:pooled-def} by provenance splits the pooled scatter into the initial real batch and the subsequent batches:
\begin{equation*}
    \boldsymbol{\Sigma}^{(t)}=\frac{1}{N_t}
    \Biggl(\sum_{j=1}^{n_r}\bigl(\boldsymbol{x}^{(0)}_j-\boldsymbol{\mu}^{(t)}\bigr)
    \bigl(\boldsymbol{x}^{(0)}_j-\boldsymbol{\mu}^{(t)}\bigr)^\top
    +\sum_{s=1}^{t-1}\sum_{i=1}^{n_b(s)}
    \bigl(\boldsymbol{x}^{(s)}_i-\boldsymbol{\mu}^{(t)}\bigr)
    \bigl(\boldsymbol{x}^{(s)}_i-\boldsymbol{\mu}^{(t)}\bigr)^\top\Biggr),
\end{equation*}
which is exactly~\eqref{eq:Sigma-update}. This completes the proof.
\end{proof}

Finally, we prove that the estimated covariance $\boldsymbol{\Sigma}^{(t)}$ is PSD. %

\begin{lemma}[The covariance estimator is PSD]
\label{lem:sample covariance is PSD}
Fix $t \ge 1$. Consider the setup in Section~\ref{sec:setup}. For every realization of the data, the sample covariance matrix $\boldsymbol{\Sigma}^{(t)}$ is positive-semi-definite.
\end{lemma}

\begin{proof}
Fix $t \ge 1$. By~\eqref{eq:Sigma-update} --- with the convention that the double sum
$\sum_{s=1}^{t-1}$ is empty when $t=1$, so that $\boldsymbol{\Sigma}^{(1)}$ reduces
to~\eqref{eq:init} --- the estimator is a nonnegatively weighted sum of outer
products of the centered points $\boldsymbol{x}^{(s)}_i-\boldsymbol{\mu}^{(t)}$.
Each such outer product is symmetric, hence so is $\boldsymbol{\Sigma}^{(t)}$.
Moreover, for every $\boldsymbol{v}\in\mathbb{R}^d$,
\begin{equation*}
    \boldsymbol{v}^\top\boldsymbol{\Sigma}^{(t)}\boldsymbol{v}
    =\frac{1}{N_t}\Biggl(
    \sum_{j=1}^{n_r}\bigl(\boldsymbol{v}^\top(\boldsymbol{x}^{(0)}_j-\boldsymbol{\mu}^{(t)})\bigr)^2
    +\sum_{s=1}^{t-1}\sum_{i=1}^{n_b(s)}
      \bigl(\boldsymbol{v}^\top(\boldsymbol{x}^{(s)}_i-\boldsymbol{\mu}^{(t)})\bigr)^2
    \Biggr)\ \ge\ 0,
\end{equation*}
since it is a sum of squares scaled by $1/N_t>0$. Hence $\boldsymbol{\Sigma}^{(t)} \succeq 0$.
\end{proof}

\textbf{Remark (singular covariance).} If $\boldsymbol{\Sigma}^{(t)}$ is singular, $\mathcal{N}(\boldsymbol{\mu}^{(t)},\boldsymbol{\Sigma}^{(t)})$ in~\eqref{eq:mixture} denotes the corresponding degenerate Gaussian, i.e., the law of $\boldsymbol{\mu}^{(t)}+(\boldsymbol{\Sigma}^{(t)})^{1/2}\boldsymbol{z}$ with $\boldsymbol{z}\sim\mathcal{N}(\boldsymbol{0},\boldsymbol{I}_d)$, where $(\boldsymbol{\Sigma}^{(t)})^{1/2}$ is the unique symmetric PSD square root, which exists by Lemma~\ref{lem:sample covariance is PSD}.

In the following subsections, we present the proofs of Theorem~\ref{thm:estimators_bias}, Lemma~\ref{lem:bound}, Theorem~\ref{thm:unified_sharp_criterion}, Lemma~\ref{lem: sup_t alpha_t < 1 => cond holds}, Proposition~\ref{prop: alpha = 1 perm bias}, Theorem~\ref{prop:rate} and Proposition~\ref{prop:transient}.

\subsection{Proof of Theorem~\ref{thm:estimators_bias}}
\label{app:estimators_bias_thm} 

\begin{proof}[Proof of part~(a): $\boldsymbol{\mu}^{(t)}$ is unbiased] We prove by induction on $t \ge 1$.

\medskip\noindent\textit{Base case $(t=1)$.}
By~\eqref{eq:init}, $\boldsymbol{\mu}^{(1)} = \frac{1}{n_r}\sum_{j=1}^{n_r} \boldsymbol{x}^{(0)}_j$ is the
sample mean of $n_r$ i.i.d.\ draws from $\mathcal{N}(\boldsymbol{\mu},\boldsymbol{\Sigma})$, so
\begin{equation*}
    \mathbb{E}\bigl[\boldsymbol{\mu}^{(1)}\bigr]
    =\frac{1}{n_r}\sum_{j=1}^{n_r}\mathbb{E}\bigl[\boldsymbol{x}^{(0)}_j\bigr]=\boldsymbol{\mu}.
\end{equation*}

\medskip\noindent\textit{Inductive step.}
Fix $t \ge 2$ and assume $\mathbb{E}\bigl[\boldsymbol{\mu}^{(t-1)}\bigr]=\boldsymbol{\mu}$. Taking expectations in \eqref{eq:mu-update} and using the linearity of expectation, we get
\begin{equation}
\label{eq:mu-unbiased-outer}
  \mathbb{E}\bigl[\boldsymbol{\mu}^{(t)}\bigr]
  = \rho_{t - 1}\,\mathbb{E}\bigl[\boldsymbol{\mu}^{(t-1)}\bigr]
  +\frac{1}{N_t}\sum_{i=1}^{n_b(t-1)}\mathbb{E}\bigl[\boldsymbol{x}^{(t-1)}_i\bigr].
\end{equation}
To evaluate $\mathbb{E}\bigl[\boldsymbol{x}^{(t-1)}_i\bigr]$, we condition on
$\mathcal{F}_{t-1}$
(recall~\eqref{eq: Ft definition}). By Assumption~\ref{ass:cond-ind}, given
$\mathcal{F}_{t-1}$, the samples $\boldsymbol{x}^{(t-1)}_i$ are i.i.d. from the mixture~\eqref{eq:mixture},
whose mean is $\alpha_{t-1}\boldsymbol{\mu}^{(t-1)}+(1-\alpha_{t-1})\boldsymbol{\mu}$. Hence, by the tower
property and the induction hypothesis,
\begin{align*}
  \mathbb{E}\bigl[\boldsymbol{x}^{(t-1)}_i\bigr]
  &= \mathbb{E}\Bigl[\mathbb{E}\bigl[\boldsymbol{x}^{(t-1)}_i\mid\mathcal{F}_{t-1}\bigr]\Bigr]
   = \mathbb{E}\bigl[\alpha_{t-1}\boldsymbol{\mu}^{(t-1)}+(1-\alpha_{t-1})\boldsymbol{\mu}\bigr] \\
  &= \alpha_{t-1}\,\mathbb{E}\bigl[\boldsymbol{\mu}^{(t-1)}\bigr]+(1-\alpha_{t-1})\boldsymbol{\mu}
   = \alpha_{t-1}\boldsymbol{\mu}+(1-\alpha_{t-1})\boldsymbol{\mu}=\boldsymbol{\mu}.
\end{align*}
Substituting $\mathbb{E}\bigl[\boldsymbol{\mu}^{(t-1)}\bigr]=\boldsymbol{\mu}$ and
$\mathbb{E}\bigl[\boldsymbol{x}^{(t-1)}_i\bigr]=\boldsymbol{\mu}$ into~\eqref{eq:mu-unbiased-outer},
\begin{equation*}
    \mathbb{E}\bigl[\boldsymbol{\mu}^{(t)}\bigr]
  = \rho_{t - 1}\boldsymbol{\mu}+\frac{n_b(t-1)}{N_t}\boldsymbol{\mu}
  = \frac{N_{t-1}+n_b(t-1)}{N_t}\,\boldsymbol{\mu}
  = \boldsymbol{\mu},
\end{equation*}
where we used $\rho_{t - 1}=N_{t-1}/N_t$ together with $N_{t-1}+n_b(t-1)=N_t$, which follows directly from \eqref{eq:counts}.
\end{proof}

\begin{proof}[Proof of part~(b): bias of $\boldsymbol{\Sigma}^{(t)}$]

\textit{Step 1: the pooled scatter identity.} From \eqref{eq:CT-def}, we have
\begin{equation}
\label{eq:second moment of delta_t}
    \mathbb{E}\bigl[\boldsymbol{\delta}^{(s)}(\boldsymbol{\delta}^{(s)})^\top\bigr]
    = \mathbb{E}\bigl[(\boldsymbol{\mu}^{(s)}-\boldsymbol{\mu})
      (\boldsymbol{\mu}^{(s)}-\boldsymbol{\mu})^\top\bigr]
    = \boldsymbol{C}_s ,
    \qquad s\ge1.
\end{equation}
Multiplying by $N_t$ and taking expectations in \eqref{eq: pooled scatter identity}, we get
\begin{equation}
\label{eq:Sigma-bias-key}
\begin{aligned}
  N_t\,\mathbb{E}\bigl[\boldsymbol{\Sigma}^{(t)}\bigr]
  &= \sum_{j=1}^{n_r}\mathbb{E}\bigl[(\boldsymbol{x}^{(0)}_j-\boldsymbol{\mu})(\boldsymbol{x}^{(0)}_j-\boldsymbol{\mu})^\top\bigr]
   +\sum_{s=1}^{t-1}\sum_{i=1}^{n_b(s)}\mathbb{E}\bigl[(\boldsymbol{x}^{(s)}_i-\boldsymbol{\mu})(\boldsymbol{x}^{(s)}_i-\boldsymbol{\mu})^\top\bigr]
   -N_t\boldsymbol{C}_t
   \\
   &= n_r\boldsymbol{\Sigma} +\sum_{s=1}^{t-1}\sum_{i=1}^{n_b(s)}\mathbb{E}\bigl[(\boldsymbol{x}^{(s)}_i-\boldsymbol{\mu})(\boldsymbol{x}^{(s)}_i-\boldsymbol{\mu})^\top\bigr]
   -N_t\boldsymbol{C}_t,
\end{aligned}
\end{equation}
where we used that $\boldsymbol{x}^{(0)}_j \sim \mathcal{N}(\boldsymbol{\mu},\boldsymbol{\Sigma})$ are i.i.d., so each contributes $\mathbb{E}[(\boldsymbol{x}^{(0)}_j-\boldsymbol{\mu})(\boldsymbol{x}^{(0)}_j-\boldsymbol{\mu})^\top]=\boldsymbol{\Sigma}$.

\medskip\noindent\textit{Step 2: induction on $t$.}
We show $\mathbb{E}[\boldsymbol{\Sigma}^{(t)}]=\boldsymbol{\Sigma}-\boldsymbol{C}_t$.

\smallskip\noindent\textit{Base case $(t=1)$.}
We have $N_1 = n_r$, and thus \eqref{eq:Sigma-bias-key} gives 
\begin{equation*}
    n_r\mathbb{E}[\boldsymbol{\Sigma}^{(1)}]=n_r\boldsymbol{\Sigma}-n_r\boldsymbol{C}_1 \Rightarrow \mathbb{E}[\boldsymbol{\Sigma}^{(1)}]=\boldsymbol{\Sigma}-\boldsymbol{C}_1.
\end{equation*}

\smallskip\noindent\textit{Inductive step.} Assume
$\mathbb{E}[\boldsymbol{\Sigma}^{(s)}]=\boldsymbol{\Sigma}-\boldsymbol{C}_s$ for all $1\le s\le t-1$. For a newly drawn sample $\boldsymbol{x}^{(s)}_i$ with $1\le s \le t-1$, the
tower property, identity \eqref{eq:cond-sq} from Lemma \ref{lem:cond-moments}, the induction hypothesis, and
$\mathbb{E}[\boldsymbol{\delta}^{(s)}(\boldsymbol{\delta}^{(s)})^\top]=\boldsymbol{C}_s$ (from \eqref{eq:second moment of delta_t}) give
\begin{align*}
  \mathbb{E}\bigl[(\boldsymbol{x}^{(s)}_i-\boldsymbol{\mu})(\boldsymbol{x}^{(s)}_i-\boldsymbol{\mu})^\top\bigr] &= \mathbb{E}\left[\mathbb{E}\bigl[(\boldsymbol{x}^{(s)}_i-\boldsymbol{\mu})(\boldsymbol{x}^{(s)}_i-\boldsymbol{\mu})^\top\mid\mathcal{F}_s\bigr]\right]
  \\
  &= \mathbb{E}\left[\alpha_s\boldsymbol{\Sigma}^{(s)}+(1-\alpha_s)\boldsymbol{\Sigma} +\alpha_s\,\boldsymbol{\delta}^{(s)}(\boldsymbol{\delta}^{(s)})^\top\right]
  \\
  &= \alpha_s\,\mathbb{E}[\boldsymbol{\Sigma}^{(s)}]+(1-\alpha_s)\boldsymbol{\Sigma}+\alpha_s\boldsymbol{C}_s \\
  &= \alpha_s(\boldsymbol{\Sigma}-\boldsymbol{C}_s)+(1-\alpha_s)\boldsymbol{\Sigma}+\alpha_s\boldsymbol{C}_s
   = \boldsymbol{\Sigma}.
\end{align*}
Substituting into~\eqref{eq:Sigma-bias-key} and using
$n_r+\sum_{s=1}^{t-1}n_b(s)=N_t$,
\begin{equation*}
    N_t\,\mathbb{E}\bigl[\boldsymbol{\Sigma}^{(t)}\bigr]
    = n_r\boldsymbol{\Sigma}+\sum_{s=1}^{t-1}n_b(s)\,\boldsymbol{\Sigma}-N_t\boldsymbol{C}_t
    = N_t\boldsymbol{\Sigma}-N_t\boldsymbol{C}_t,
\end{equation*}
so $\mathbb{E}[\boldsymbol{\Sigma}^{(t)}]=\boldsymbol{\Sigma}-\boldsymbol{C}_t$.
\end{proof}

\subsection{Steps towards proving Theorem~\ref{thm:unified_sharp_criterion}}
\label{sec:perlim_results_thm_unified_sharp_criterion}

In this subsection, we present the steps towards proving Theorem~\ref{thm:unified_sharp_criterion}. 

\subsubsection{Range of scalar sequences}
\label{sec:range_scalar_seq_thm_proof}

First, we characterize the ranges of the three sequences defined in \eqref{eq:scalars} and $\rho_t$, defined in \eqref{eq:counts}. These ranges will be used extensively in the subsequent results.

\begin{lemma}[Range of $\lambda_t$, $\gamma_t$, $\rho_t$]
\label{lem:range}
$0 < \lambda_t \le 1$, $\ 0 < \gamma_t < 1$ and $0 < \rho_t < 1$ for all $t \ge 1$.
\end{lemma}

\textbf{Remark.} For $1\le s\le t$, \eqref{eq: Gamma_(s,t) def}, and Lemma~\ref{lem:range} imply $0 < \Gamma_{s,t} < 1$.

\begin{proof}
Fix $t \ge 1$. From~\eqref{eq:counts} we have $N_{t+1} = N_t + n_b(t)$. Remember that $N_t \ge n_r \ge 1$ and $n_b(t) \ge 1$.

\medskip\noindent\emph{Range of $\lambda_t$.} By~\eqref{eq:scalars} and $\alpha_t\in[0,1]$, we have 
\begin{equation*}
    \lambda_t = \frac{N_t + n_b(t)\alpha_t}{N_{t+1}} = \frac{N_t + n_b(t)\alpha_t}{N_t + n_b(t)} \Rightarrow 0 < \lambda_t \le 1
\end{equation*}

\medskip\noindent\emph{Range of $\rho_t$.} By~\eqref{eq:counts} and $n_b(t) \ge 1$, we have
\begin{equation*}
    \rho_t = \frac{N_t}{N_{t+1}} = \frac{N_t}{N_t + n_b(t)} \Rightarrow 0 < \rho_t < 1
\end{equation*}

\medskip\noindent\emph{Positivity $\gamma_t>0$.} By~\eqref{eq:scalars} we have
\begin{equation}
\label{eq: beta_t expansion in Lemma 3.6 proof}
    \beta_t = \frac{n_b(t)}{N_{t+1}^2} = \frac{n_b(t)}{\left(N_t + n_b(t)\right)^2}
\end{equation}
Substituting~\eqref{eq: beta_t expansion in Lemma 3.6 proof} into the definition $\gamma_t=\lambda_t^2-\alpha_t^2\beta_t$ from~\eqref{eq:scalars},
\begin{equation}
\label{eq:gamma-explicit-app}
\begin{aligned}
    \gamma_t &= \left(\frac{N_t + n_b(t)\alpha_t}{N_t + n_b(t)}\right)^2 - \frac{\alpha_t^2 n_b(t)}{\left(N_t + n_b(t)\right)^2} 
    \\
    &= \frac{\left(N_t  +n_b(t)\alpha_t\right)^2 - \alpha_t^2n_b(t)}{\left(N_t + n_b(t)\right)^2}
    \\
    &= \frac{N_t^2 + 2N_tn_b(t)\alpha_t + \alpha_t^2n_b(t)\left(n_b(t) - 1\right)}{\left(N_t + n_b(t)\right)^2} 
    \\
    &\ge \frac{N_t^2}{\left(N_t + n_b(t)\right)^2} > 0,
\end{aligned}
\end{equation}
where we used $2N_tn_b(t)\alpha_t \ge 0$ and $\alpha_t^2n_b(t)\left(n_b(t) - 1\right) \ge 0$ (as $n_b(t)\ge1$ and $\alpha_t\in[0,1]$). Finally, $\gamma_t > 0$.

\medskip\noindent\emph{Upper bound $\gamma_t<1$.} From~\eqref{eq:gamma-explicit-app},
\begin{align*}
    1 - \gamma_t &= \frac{\left(N_t + n_b(t)\right)^2 - \left(N_t + n_b(t)\alpha_t\right)^2 + \alpha_t^2n_b(t)}{\left(N_t + n_b(t)\right)^2}
    \\
    &= \frac{2N_tn_b(t)\left(1-\alpha_t\right) + n_b^2(t)\left(1-\alpha_t^2\right) + \alpha_t^2n_b(t)}{\left(N_t + n_b(t)\right)^2}
    \\
    &= \frac{(1-\alpha_t)n_b(t)\cdot \left(2N_t + n_b(t)\left(1 + \alpha_t\right)\right) + \alpha_t^2n_b(t)}{\left(N_t + n_b(t)\right)^2}
    \\
    &= \frac{n_b(t)}{\left(N_t + n_b(t)\right)^2}\cdot \left((1-\alpha_t)\cdot \left(2N_t + n_b(t)\left(1 + \alpha_t\right) \right) + \alpha_t^2\right)
\end{align*}
That is,
\begin{equation}
\label{eq: final formula for 1-gamma_t proof of Lemma 3.6}
    1 - \gamma_t = \beta_t\cdot \left((1-\alpha_t)\cdot \left(2N_t + n_b(t)\left(1 + \alpha_t\right) \right) + \alpha_t^2\right)
\end{equation}
If $0 \le \alpha_t < 1$, the expression above is strictly positive. Otherwise, $\alpha_t =1$, and we have
\begin{equation*}
    1 - \gamma_t = \frac{n_b(t)}{\left(N_t + n_b(t)\right)^2} > 0
\end{equation*}
In either case, $1 - \gamma_t > 0 \Rightarrow \gamma_t < 1$.
\end{proof}

\subsubsection{Consistency of $\boldsymbol{\mu}^{(t)}$}
\label{sec:mu(t)->mu in L2 cond}

We start by characterizing when $\mathbb{E}\bigl[\|\boldsymbol{\mu}^{(t)}-\boldsymbol{\mu}\|_2^2\bigr] \xrightarrow{t\rightarrow\infty} 0$. First, we provide an equivalent condition in terms of the covariance of $\boldsymbol{\mu}^{(t)}$.

\begin{lemma}[$L^2$ consistency equivalence] \label{lem: CT->0 <=> mu(t)-> mu in L2}
Under Assumption~\ref{ass:cond-ind}, we have
\begin{equation}
\label{eq: cov->0 iff mu^(t) L2 mu}
    \boldsymbol{C}_t \to \boldsymbol{0} \Leftrightarrow \mathbb{E}\bigl[\|\boldsymbol{\mu}^{(t)}-\boldsymbol{\mu}\|_2^2\bigr] \xrightarrow{t\rightarrow\infty} 0.
\end{equation}
\end{lemma}

\textbf{Remark.} We note $\boldsymbol{C}_t \to \boldsymbol{0}$ with the Frobenius norm. This is valid because $\boldsymbol{C}_t \in \mathbb{R}^{d \times d}$ is in a finite dimensional space, which means that all norms are equivalent.
Additionally, from Theorem~\ref{thm:estimators_bias}(b),  Equation~\eqref{eq: cov->0 iff mu^(t) L2 mu} is equivalent to the asymptotic unbiasedness of the covariance estimator $\boldsymbol{\Sigma}^{(t)}$.

\begin{proof}
By \eqref{eq:MSE-trace}, $\mathbb{E}\bigl[\|\boldsymbol{\mu}^{(t)}-\boldsymbol{\mu}\|_2^2\bigr] \xrightarrow{t\rightarrow\infty} 0$ if and only if $\Tr(\boldsymbol{C}_t) \to 0$. By \eqref{eq:CT-def}, the matrix $\boldsymbol{C}_t$ is symmetric, and it is positive-semi-definite as a covariance matrix. Hence, Lemma~\ref{lem:trace-frobenius}(b) gives
$\Tr(\boldsymbol{C}_t)\to 0 \iff \boldsymbol{C}_t\to\boldsymbol{0}$. Combining the two equivalences gives the goal equivalence.
\end{proof}

Next, let us provide a closed-form formula for the covariance of $\boldsymbol{\mu}^{(t)}$.

\begin{theorem}[Covariance of $\boldsymbol{\mu}^{(t)}$]\label{thm:CT}
Under Assumption~\ref{ass:cond-ind}, with $\lambda_t,\beta_t,\gamma_t$ as in~\eqref{eq:scalars}, and $\Gamma_{s,t}$ as in~\eqref{eq: Gamma_(s,t) def}:
\begin{enumerate}
  \item[(a)] \textbf{Recursion.}
    \begin{equation}\label{eq:CT-recursion}    
    \boldsymbol{C}_t = \gamma_{t-1}\boldsymbol{C}_{t-1}+\beta_{t-1}\boldsymbol{\Sigma},\quad t\ge2,\qquad \boldsymbol{C}_1=\frac{\boldsymbol{\Sigma}}{n_r}.
    \end{equation}
  \item[(b)] \textbf{Closed form.}
    \begin{equation}\label{eq:CT-closed}
      \boldsymbol{C}_t=\Gamma_{1,t-1}\,\boldsymbol{C}_1+\boldsymbol{\Sigma}\sum_{s=1}^{t-1}\beta_s\,\Gamma_{s+1,t-1}.
    \end{equation}

  \item[(c)] \textbf{Special case $(\alpha_t \equiv 0)$.} The covariance is $\boldsymbol{C}_t = \boldsymbol{\Sigma}/N_t$. This is the covariance of the sample mean of $N_t$ i.i.d. real samples (indeed, no generated data is used).

  \item[(d)] \textbf{Special case $(\alpha_t \equiv 1)$.} The recursion becomes $\boldsymbol{C}_t=(1-\beta_{t-1})\boldsymbol{C}_{t-1}+\beta_{t-1}\boldsymbol{\Sigma}$. In this case, all new samples are synthetic.

  {\color{black}
  \item[(e)] \textbf{Scalar form.} The covariance is a scalar multiple of the true covariance, $\boldsymbol{C}_t=c_t\boldsymbol{\Sigma}$, where
  \begin{equation}\label{eq:cT-scalar}
      c_1=\frac{1}{n_r},\qquad c_t=\gamma_{t-1}c_{t-1}+\beta_{t-1} \quad (t \ge 2).
  \end{equation}}
\end{enumerate}
\end{theorem}

\textbf{Remark.} In the recursion~\eqref{eq:CT-recursion}, $\gamma_{t-1}\in(0,1)$ (Lemma~\ref{lem:range}) acts as a damping factor on the covariance from the previous iteration, while the added terms $\beta_{t-1}\boldsymbol{\Sigma}$ are summable (Lemma~\ref{lem:beta-summability}). Together, these yield the bound $\boldsymbol{C}_t\preceq\frac{2}{n_r}\boldsymbol{\Sigma}$ of Lemma~\ref{lem:bound}.

\begin{proof}[Proof of part~(a): recursion]
Apply the law of total covariance to $\boldsymbol{\mu}^{(t)}$, conditioning on
$\mathcal{F}_{t-1}$:
\begin{equation}\label{eq:tot-cov-CT}
  \boldsymbol{C}_t = \Cov\!\bigl(\mathbb{E}[\boldsymbol{\mu}^{(t)}\mid\mathcal{F}_{t-1}]\bigr)
      + \mathbb{E}\!\bigl[\Cov(\boldsymbol{\mu}^{(t)}\mid\mathcal{F}_{t-1})\bigr].
\end{equation}

\medskip\noindent\textbf{Term 1 (covariance of the conditional mean).}
By~\eqref{eq:mu-update} and~\eqref{eq:cond-mean} from Lemma~\ref{lem:cond-moments},
\begin{equation}
\label{eq: E[mu|F] expansion proof Thm 3.7}
\begin{aligned}
    \mathbb{E}[\boldsymbol{\mu}^{(t)}\mid\mathcal{F}_{t-1}] &= \mathbb{E}\left[\rho_{t - 1}\boldsymbol{\mu}^{(t-1)} + \frac{1}{N_t} \sum_{i=1}^{n_b(t-1)} \boldsymbol{x}^{(t-1)}_i  \ \middle| \ \mathcal{F}_{t-1} \right]
    \\
    &= \rho_{t - 1} \boldsymbol{\mu}^{(t-1)} + \frac{1}{N_t}\sum_{i=1}^{n_b(t-1)} \mathbb{E}[\boldsymbol{x}^{(t-1)}_i \mid \mathcal{F}_{t-1}]
    \\
    &= \rho_{t - 1}\boldsymbol{\mu}^{(t-1)} + \frac{n_b(t-1)}{N_t}\cdot \left(\alpha_{t-1}\boldsymbol{\mu}^{(t-1)} + (1-\alpha_{t-1})\boldsymbol{\mu}\right)
    \\
    &= \frac{N_{t-1} + n_b(t-1)\alpha_{t-1}}{N_t}\boldsymbol{\mu}^{(t-1)} + \frac{(1-\alpha_{t-1})n_b(t-1)}{N_t}\boldsymbol{\mu}
    \\
    &= \lambda_{t-1}\boldsymbol{\mu}^{(t-1)} + (1-\lambda_{t-1})\boldsymbol{\mu}
\end{aligned}
\end{equation}
where we used the definition of $\lambda_t$ in~\eqref{eq:scalars}, and
\begin{equation*}
    1 - \lambda_{t} = 1 - \frac{N_t + n_b(t)\alpha_t}{N_{t+1}} = \frac{N_{t+1} - N_t - n_b(t)\alpha_t}{N_{t+1}} = \frac{\left(1 - \alpha_t\right)n_b(t)}{N_{t+1}}.
\end{equation*}
As $\lambda_{t-1}$ is deterministic and $(1-\lambda_{t-1})\boldsymbol{\mu}$ is constant, from~\eqref{eq: E[mu|F] expansion proof Thm 3.7} we have
\begin{equation}
\label{eq: first part in C_T formula - proof Thm 3.7}
    \Cov\!\bigl(\mathbb{E}[\boldsymbol{\mu}^{(t)}\mid\mathcal{F}_{t-1}]\bigr)
  = \lambda_{t-1}^2\,\Cov(\boldsymbol{\mu}^{(t-1)})=\lambda_{t-1}^2\,\boldsymbol{C}_{t-1}.
\end{equation}
\medskip\noindent\textbf{Term 2 (expectation of the conditional covariance).}
By \eqref{eq:mu-update}, the conditional covariance reads:
\begin{align*}
    \Cov\!\bigl(\boldsymbol{\mu}^{(t)} \mid \mathcal{F}_{t-1}\bigr) &= \Cov\!\left(\rho_{t - 1}\boldsymbol{\mu}^{(t-1)} + \frac{1}{N_t}\sum_{i=1}^{n_b(t-1)} \boldsymbol{x}^{(t-1)}_i \ \middle| \ \mathcal{F}_{t-1}\right)
    \\
    &= \frac{n_b(t-1)}{N_t^2}\,\Cov\!\bigl(\boldsymbol{x}^{(t-1)}_1\mid\mathcal{F}_{t-1}\bigr)
    \\
    &= \beta_{t-1}\,\Cov\!\bigl(\boldsymbol{x}^{(t-1)}_1\mid\mathcal{F}_{t-1}\bigr)
\end{align*}

where we used that given $\mathcal{F}_{t-1}$, the term $\rho_{t - 1}\boldsymbol{\mu}^{(t-1)}$ is deterministic, and the
$n_b(t-1)$ new samples are i.i.d. (Assumption \ref{ass:cond-ind}), so the cross-covariances vanish. We also used the definition of $\beta_t$ in \eqref{eq:scalars}. Using \eqref{eq:cond-cov} from Lemma~\ref{lem:cond-moments} with $t-1$, the conditional covariance is:
\begin{equation}
\label{eq: conditional covariance expansion - proof Thm 3.7}
    \Cov\!\bigl(\boldsymbol{\mu}^{(t)} \mid \mathcal{F}_{t-1}\bigr) = \beta_{t-1}\left(\alpha_{t-1}\boldsymbol{\Sigma}^{(t-1)} + (1-\alpha_{t-1})\boldsymbol{\Sigma} + \alpha_{t-1}(1-\alpha_{t-1})\boldsymbol{\delta}^{(t-1)}(\boldsymbol{\delta}^{(t-1)})^\top\right).
\end{equation}
Taking the outer expectation and using
$\mathbb{E}[\boldsymbol{\Sigma}^{(t-1)}]=\boldsymbol{\Sigma}-\boldsymbol{C}_{t-1}$ (Theorem~\ref{thm:estimators_bias}(b)) together
with $\mathbb{E}[\boldsymbol{\delta}^{(t-1)}(\boldsymbol{\delta}^{(t-1)})^\top]=\boldsymbol{C}_{t-1}$
($\boldsymbol{\delta}^{(t-1)}=\boldsymbol{\mu}^{(t-1)}-\boldsymbol{\mu}$ and \eqref{eq:CT-def}), from \eqref{eq: conditional covariance expansion - proof Thm 3.7} we have
\begin{equation}
\label{eq: second part in C_T formula - proof Thm 3.7}
\begin{aligned}
    \mathbb{E}\!\bigl[\Cov(\boldsymbol{\mu}^{(t)}\mid\mathcal{F}_{t-1})\bigr]
      &= \beta_{t-1}\Bigl(\alpha_{t-1}(\boldsymbol{\Sigma}-\boldsymbol{C}_{t-1})+(1-\alpha_{t-1})\boldsymbol{\Sigma}
         +\alpha_{t-1}(1-\alpha_{t-1})\boldsymbol{C}_{t-1}\Bigr) \\
      &= \beta_{t-1}\Bigl(\alpha_{t-1}\boldsymbol{\Sigma}-\alpha_{t-1}\boldsymbol{C}_{t-1}+(1-\alpha_{t-1})\boldsymbol{\Sigma}
         +\alpha_{t-1}\boldsymbol{C}_{t-1}-\alpha_{t-1}^2\boldsymbol{C}_{t-1}\Bigr) \\
      &= \beta_{t-1}\bigl(\boldsymbol{\Sigma}-\alpha_{t-1}^2\boldsymbol{C}_{t-1}\bigr).
\end{aligned}
\end{equation}

\medskip\noindent\textbf{Combining.} From \eqref{eq:tot-cov-CT}, \eqref{eq: first part in C_T formula - proof Thm 3.7} and \eqref{eq: second part in C_T formula - proof Thm 3.7}, the covariance $\boldsymbol{C}_t$ reads:
\begin{align*}
    \boldsymbol{C}_t
    &= \lambda_{t-1}^2\boldsymbol{C}_{t-1}+\beta_{t-1}(\boldsymbol{\Sigma}-\alpha_{t-1}^2\boldsymbol{C}_{t-1})
    \\
    &= (\lambda_{t-1}^2-\alpha_{t-1}^2\beta_{t-1})\boldsymbol{C}_{t-1}+\beta_{t-1}\boldsymbol{\Sigma}
    \\
    &= \gamma_{t-1}\boldsymbol{C}_{t-1}+\beta_{t-1}\boldsymbol{\Sigma}
\end{align*}
where we used the definition of $\gamma_t$ in \eqref{eq:scalars}. The initial condition is
\begin{equation}
\label{eq: C_1 proof Thm 3.7}
    \boldsymbol{C}_1 = \Cov\!\left(\frac{1}{n_r}\sum_{j=1}^{n_r} \boldsymbol{x}^{(0)}_j\right) = \frac{1}{n_r^2}\cdot n_r\boldsymbol{\Sigma} = \frac{\boldsymbol{\Sigma}}{n_r}.
\end{equation}
\end{proof}

\begin{proof}[Proof of part~(b): closed form]
We prove \eqref{eq:CT-closed} via induction on $t\ge1$. 
\begin{itemize}
    \item For $t=1$, the right-hand side of~\eqref{eq:CT-closed} is $\Gamma_{1,0}\boldsymbol{C}_1=\boldsymbol{C}_1$ (empty sum; $\Gamma_{1,0}=1$ by~\eqref{eq: Gamma_(s,t) def}).

    \item Assuming $\boldsymbol{C}_{t-1}=\Gamma_{1,t-2}\boldsymbol{C}_1+\boldsymbol{\Sigma}\sum_{s=1}^{t-2}\beta_s\Gamma_{s+1,t-2}$ and
    substituting into the recursion~\eqref{eq:CT-recursion},

    \begin{align*}
      \boldsymbol{C}_t
      &= \gamma_{t-1}\Bigl(\Gamma_{1,t-2}\boldsymbol{C}_1
         +\boldsymbol{\Sigma}\sum_{s=1}^{t-2}\beta_s\Gamma_{s+1,t-2}\Bigr)+\beta_{t-1}\boldsymbol{\Sigma} \\
      &= \underbrace{\gamma_{t-1}\Gamma_{1,t-2}}_{\Gamma_{1,t-1}}\boldsymbol{C}_1
         +\boldsymbol{\Sigma}\sum_{s=1}^{t-2}\beta_s\underbrace{\gamma_{t-1}\Gamma_{s+1,t-2}}_{\Gamma_{s+1,t-1}}
         +\beta_{t-1}\boldsymbol{\Sigma}\underbrace{\Gamma_{t,t-1}}_{=1} \\
      &= \Gamma_{1,t-1}\boldsymbol{C}_1+\boldsymbol{\Sigma}\sum_{s=1}^{t-1}\beta_s\Gamma_{s+1,t-1},
    \end{align*}
    using $\gamma_{t-1}\Gamma_{s,t-2}=\Gamma_{s,t-1}$ and the empty product $\Gamma_{t,t-1}=1$, so
    that $\beta_{t-1}\boldsymbol{\Sigma}$ becomes the $s=t-1$ term of the sum. We used the definition of $\Gamma_{s,t}$ in \eqref{eq: Gamma_(s,t) def} throughout.
    
\end{itemize}
\end{proof}

\begin{proof}[Proof of part~(c): $\alpha_t\equiv0$] Assume $\alpha_t \equiv 0$. From \eqref{eq:scalars} we have:
\begin{equation}
\label{eq: gamma, beta for alpha=0 proof Thm 3.7}
    \gamma_t = \lambda_t^2 = \frac{N_t^2}{N_{t+1}^2}, \qquad \beta_t = \frac{n_b(t)}{N_{t+1}^2}.
\end{equation}
We prove that $\boldsymbol{C}_t=\boldsymbol{\Sigma}/N_t$ by induction on $t \ge 1$:
\begin{itemize}
    \item For $t = 1$, using $N_1 = n_r$, from \eqref{eq: C_1 proof Thm 3.7} we have
    \begin{equation*}
        \boldsymbol{C}_1 = \frac{\boldsymbol{\Sigma}}{n_r} = \frac{\boldsymbol{\Sigma}}{N_1}
    \end{equation*}

    \item Assuming $\boldsymbol{C}_{t-1}=\boldsymbol{\Sigma}/N_{t-1}$, the
    recursion~\eqref{eq:CT-recursion} gives
    \begin{align*}
        \boldsymbol{C}_t &= \gamma_{t-1}\boldsymbol{C}_{t-1}+\beta_{t-1}\boldsymbol{\Sigma}
        \\
        & =\frac{N_{t-1}^2}{N_t^2}\cdot\frac{\boldsymbol{\Sigma}}{N_{t-1}}+\frac{n_b(t-1)}{N_t^2}\boldsymbol{\Sigma}
        \\
        &=\frac{N_{t-1}+n_b(t-1)}{N_t^2}\boldsymbol{\Sigma}
        \\
        &=\frac{\boldsymbol{\Sigma}}{N_t}.
    \end{align*}
    where we used the formulas for $\gamma_t, \beta_t$ from \eqref{eq: gamma, beta for alpha=0 proof Thm 3.7}, and $N_{t-1} + n_b(t - 1) = N_t$ from \eqref{eq:counts}.
\end{itemize}
\end{proof}

\begin{proof}[Proof of part~(d): $\alpha_t\equiv1$] Assume $\alpha_t \equiv 1$. From $N_t+n_b(t)=N_{t+1}$ (follows from \eqref{eq:counts}) we have 
\begin{equation*}
    \lambda_t = \frac{N_t + n_b(t)\alpha_t}{N_{t+1}} = \frac{N_t + n_b(t)}{N_{t+1}} = 1
\end{equation*}
Hence, $\gamma_t=\lambda_t^2-\alpha_t^2\beta_t=1-\beta_t$. The recursion~\eqref{eq:CT-recursion}
then reads $\boldsymbol{C}_t = (1-\beta_{t-1})\boldsymbol{C}_{t-1}+\beta_{t-1}\boldsymbol{\Sigma}$.
\end{proof}

{\color{black}
\begin{proof}[Proof of part~(e): scalar form]
We prove by induction on $t\ge1$. For $t=1$, $\boldsymbol{C}_1=\boldsymbol{\Sigma}/n_r=c_1\boldsymbol{\Sigma}$ by~\eqref{eq: C_1 proof Thm 3.7}. Assuming $\boldsymbol{C}_{t-1}=c_{t-1}\boldsymbol{\Sigma}$, the recursion~\eqref{eq:CT-recursion} gives
\begin{equation*}
    \boldsymbol{C}_t=\gamma_{t-1}c_{t-1}\boldsymbol{\Sigma}+\beta_{t-1}\boldsymbol{\Sigma}=c_t\boldsymbol{\Sigma}.
\end{equation*}
\end{proof}}

{\color{black} We can now prove Lemma~\ref{lem:bound}, stated in the main text.

\begin{proof}[Proof of Lemma~\ref{lem:bound}]
The lower bound $\boldsymbol{C}_t\succeq\boldsymbol{0}$ holds since $\boldsymbol{C}_t$ is a covariance matrix~\eqref{eq:mu_cov_first_def}.

For the upper bound, by Theorem~\ref{thm:CT}(e) we have $\boldsymbol{C}_t=c_t\boldsymbol{\Sigma}$, with $c_t$ given by~\eqref{eq:cT-scalar}. First, $c_t\ge0$ for every $t\ge1$ by induction, since $c_1=1/n_r>0$, and $\gamma_t>0$ and $\beta_t>0$ for every $t\ge1$ (Lemma~\ref{lem:range} and~\eqref{eq:scalars}). Hence, using $\gamma_{t-1}<1$ (Lemma~\ref{lem:range}),
\begin{equation*}
    c_t=\gamma_{t-1}c_{t-1}+\beta_{t-1}\le c_{t-1}+\beta_{t-1},\qquad t\ge2.
\end{equation*}
Unrolling this inequality and applying Lemma~\ref{lem:beta-summability},
\begin{equation*}
    c_t\le c_1+\sum_{s=1}^{t-1}\beta_s\le\frac{1}{n_r}+\frac{1}{n_r}=\frac{2}{n_r},\qquad t\ge1,
\end{equation*}
where for $t=1$ the sum is empty. Since $\boldsymbol{\Sigma}\succ0$, we conclude
\begin{equation*}
    \boldsymbol{C}_t=c_t\boldsymbol{\Sigma}\preceq\frac{2}{n_r}\boldsymbol{\Sigma}.
\end{equation*}
\end{proof}}

We need to characterize when the estimated mean converges to the truth. By Lemma~\ref{lem: CT->0 <=> mu(t)-> mu in L2}, this is equivalent to characterizing
when $\boldsymbol{C}_t \xrightarrow{t\rightarrow\infty} \boldsymbol{0}$. The following theorem addresses this.

\begin{theorem}[General necessary and sufficient condition]\label{thm:NS-theorem-for-CT->0} Under Assumption~\ref{ass:cond-ind}, with $\lambda_t,\beta_t,\gamma_t$ as in~\eqref{eq:scalars}, $\Gamma_{s,t}$ as in~\eqref{eq: Gamma_(s,t) def}, $\boldsymbol{C}_t\to0$ as $t\to\infty$
if and only if
\begin{equation}\label{eq:NS-condition for CT->0}
  \Gamma_{1,t - 1}\to0
  \qquad\text{and}\qquad
  S_t \coloneqq \sum_{s=1}^{t-1}\beta_s\,\Gamma_{s+1,t-1}\to0.
\end{equation}
\end{theorem}

\textbf{Remark.} From \eqref{eq: cov->0 iff mu^(t) L2 mu}, it follows that Theorem~\ref{thm:NS-theorem-for-CT->0} provides a necessary and sufficient condition for $\mathbb{E}\bigl[\|\boldsymbol{\mu}^{(t)}-\boldsymbol{\mu}\|_2^2\bigr] \xrightarrow{t\rightarrow\infty} 0$.

\begin{proof}
By the closed form of Theorem~\ref{thm:CT}(b), \eqref{eq:CT-closed}, we have
\begin{equation}\label{eq:CT-split-app}
  \boldsymbol{C}_t=\Gamma_{1,t-1}\boldsymbol{C}_1+S_t\,\boldsymbol{\Sigma},
  \qquad S_t = \sum_{s=1}^{t-1}\beta_s\,\Gamma_{s+1,t-1}.
\end{equation}

\medskip\noindent\textit{Sufficiency.} Suppose $\Gamma_{1,t-1}\to0$ and $S_t\to0$. By the triangle inequality and $\|c\boldsymbol{A}\|_F=|c|\,\|\boldsymbol{A}\|_F$ for a scalar $c$,
\begin{equation*}
    0 \le \|\boldsymbol{C}_t\|_F
    \le\Gamma_{1,t-1}\|\boldsymbol{C}_1\|_F+S_t\|\boldsymbol{\Sigma}\|_F \longrightarrow 0,
\end{equation*}
since $\|\boldsymbol{C}_1\|_F$ and $\|\boldsymbol{\Sigma}\|_F$ are finite constants. We also used that $\Gamma_{1,t-1} \ge 0, S_t \ge 0$. Hence $\boldsymbol{C}_t\to\boldsymbol{0}$.

\medskip\noindent\textit{Necessity.} Suppose $\boldsymbol{C}_t\to\boldsymbol{0}$. By~\eqref{eq:CT-def}, the matrix $\boldsymbol{C}_t$ is symmetric, and it is positive-semi-definite as a covariance matrix. Hence, Lemma~\ref{lem:trace-frobenius}(b) gives $\Tr(\boldsymbol{C}_t) \to 0$. From \eqref{eq:CT-split-app}, we have
\begin{equation*}
    \Tr(\boldsymbol{C}_t) = \Gamma_{1, t -1}\Tr(\boldsymbol{C}_1) + S_t\Tr(\boldsymbol{\Sigma}).
\end{equation*}
Both summands are nonnegative: $\Gamma_{1,t -1} \ge 0$ and $S_t \ge 0$ by
\eqref{eq: Gamma_(s,t) def} together with $\gamma_t \in (0,1)$ and $\beta_t > 0$
(Lemma~\ref{lem:range}), while $\Tr(\boldsymbol{C}_1)=\Tr(\boldsymbol{\Sigma})/n_r>0$
and $\Tr(\boldsymbol{\Sigma})>0$ since $\boldsymbol{\Sigma}\succ0$. Discarding one
summand at a time and dividing by these two positive constants gives
\begin{equation*}
    0 \le \Gamma_{1,t - 1} \le \frac{\Tr(\boldsymbol{C}_t)}{\Tr(\boldsymbol{C}_1)}
    \longrightarrow 0,
    \qquad
    0 \le S_t \le \frac{\Tr(\boldsymbol{C}_t)}{\Tr(\boldsymbol{\Sigma})}
    \longrightarrow 0 .
\end{equation*}
Hence, $\Gamma_{1,t-1} \to 0$ and $S_t \to 0$.
\end{proof}

Next, we show that the condition for $\boldsymbol{C}_t \xrightarrow{t\rightarrow\infty} \boldsymbol{0}$ collapses to a single elementary series in $\{\alpha_t\}_{t \ge 1}$, $\{n_b(t)\}_{t \ge 1}$, and $n_r$.

\begin{theorem}[Sharp elementary criterion for consistency]
\label{thm:sharp-criterion-mu}
Under Assumption~\ref{ass:cond-ind}, with $\rho_t$ as in~\eqref{eq:counts},
\begin{equation}\label{eq:sharp-criterion-mu}
  \boldsymbol{C}_t\to\boldsymbol{0}
  \quad\Longleftrightarrow\quad
  \sum_{t=1}^{\infty}(1-\alpha_t)(1-\rho_t)
  =\sum_{t=1}^{\infty}(1-\alpha_t)\,\frac{n_b(t)}{N_{t+1}}=\infty ,
\end{equation}
and equivalently for $\mathbb{E}\bigl[\|\boldsymbol{\mu}^{(t)}-\boldsymbol{\mu}\|_2^2\bigr] \xrightarrow{t\rightarrow\infty} 0$. In particular, $S_t \to 0$ whenever $\Gamma_{1,t-1} \to 0$, so the second condition
in~\eqref{eq:NS-condition for CT->0} is redundant.
\end{theorem}

\begin{proof}
It is convenient to route the argument through two intermediate conditions. Write
\begin{equation*}
    \text{(a)}\ \boldsymbol{C}_t\to\boldsymbol{0},\qquad
    \text{(b)}\ \Gamma_{1,t-1}\to0,\qquad
    \text{(c)}\ \sum_{t=1}^\infty(1-\gamma_t)=\infty,\qquad
    \text{(d)}\ \sum_{t=1}^\infty(1-\alpha_t)(1 - \rho_t)=\infty,
\end{equation*}
so that the assertion of the theorem is (a)$\Leftrightarrow$(d). Theorem~\ref{thm:NS-theorem-for-CT->0}
already gives (a)$\Rightarrow$(b); we prove (b)$\Rightarrow$(a), (b)$\Leftrightarrow$(c) and
(c)$\Leftrightarrow$(d), which together yield (a)$\Leftrightarrow$(d) as well as the
equivalences among (b), (c) and (d). The final restatement in terms of $\mathbb{E}\bigl[\|\boldsymbol{\mu}^{(t)}-\boldsymbol{\mu}\|_2^2\bigr] \xrightarrow{t\rightarrow\infty} 0$ is
Lemma~\ref{lem: CT->0 <=> mu(t)-> mu in L2}.

\medskip\noindent\textit{Step 1: (b)$\Rightarrow$(a).} We first prove that $\Gamma_{1,t - 1} \to 0$ implies $S_t \to 0$.

Assume that $\Gamma_{1,t-1} \to 0$, and fix $\varepsilon>0$. By Lemma~\ref{lem:beta-summability} choose $s_0$ with
$\sum_{s>s_0}\beta_s<\varepsilon$, and for $t-1>s_0$ split
\begin{equation*}
  S_t = \sum_{s=1}^{t-1} \beta_s\Gamma_{s+1,t-1} =\underbrace{\sum_{s=1}^{s_0}\beta_s\Gamma_{s+1,t-1}}_{A_t} +\underbrace{\sum_{s=s_0+1}^{t-1}\beta_s\Gamma_{s+1,t-1}}_{B_t}.
\end{equation*}

By Lemma~\ref{lem:range} every $\gamma_k\in(0,1)$, so each $\Gamma_{s+1,t-1} = \prod_{k=s+1}^{t-1} \gamma_k \le 1$ and
$B_t \le \sum_{s>s_0}\beta_s < \varepsilon$. 

In addition, for $s \le s_0$, from the definition of $\Gamma_{s,t}$ in \eqref{eq: Gamma_(s,t) def}:
\begin{equation*}
    \Gamma_{s+1,t-1} = \Gamma_{s_0+1,t-1}\cdot \Gamma_{s+1, s_0} \le \Gamma_{s_0+1,t-1}
\end{equation*}

Thus,
\begin{equation*}
    0 \le A_t \le \Gamma_{s_0+1, t-1}\cdot \sum_{s=1}^{s_0} \beta_s = \frac{\sum_{s=1}^{s_0} \beta_s}{\Gamma_{1,s_0}}\cdot \Gamma_{1,t-1} \to 0
\end{equation*}

That is, $A_t \to 0$. Therefore, $\limsup_t S_t \le \varepsilon$, and as $\varepsilon > 0$ was arbitrary and $S_t \ge 0$ we get $S_t \to 0$. Both conditions of~\eqref{eq:NS-condition for CT->0} now hold, so Theorem~\ref{thm:NS-theorem-for-CT->0} gives $\boldsymbol{C}_t\to\boldsymbol{0}$. This also proves the final assertion of the theorem.

\medskip\noindent\textit{Step 2: (b)$\Leftrightarrow$(c).}
Assume~(c). Since $\gamma_t\in(0,1)$, the bound $\ln u\le u-1$ gives
\begin{equation*}
    \Gamma_{1,t-1} = \prod_{s=1}^{t-1} \gamma_s = \exp\!\left(\sum_{s=1}^{t-1}\ln\gamma_s\right)
    \le\exp\!\left(-\sum_{s=1}^{t-1}(1-\gamma_s)\right)\xrightarrow[t\to\infty]{}0,
\end{equation*}
which, together with $\Gamma_{1, t-1} \ge 0$, is~(b). Conversely,~(b) together with $\gamma_t\in(0,1)$ is precisely the hypothesis of Lemma~\ref{lem:product-decay}, whose conclusion is~(c).

\medskip\noindent\textit{Step 3: (c)$\Leftrightarrow$(d).} Remember that from \eqref{eq:counts} and \eqref{eq:scalars}, we have $N_{t+1} = N_t + n_b(t)$ and $\beta_t = \frac{n_b(t)}{N_{t+1}^2}$.

Using the identity~\eqref{eq: final formula for 1-gamma_t proof of Lemma 3.6}, we have

\begin{equation}\label{eq:1-gamma-beta}
  1-\gamma_t=\beta_t\bigl[(1-\alpha_t)K_t+\alpha_t^2\bigr],
  \qquad K_t \coloneqq 2N_t+n_b(t)(1+\alpha_t).
\end{equation}
Because $0\le\alpha_t^2\le1$, Lemma~\ref{lem:beta-summability} yields
\begin{equation*}
    \sum_{t = 1}^\infty \beta_t\alpha_t^2 \le \sum_{t = 1}^\infty \beta_t \le \frac{1}{n_r} <\infty.
\end{equation*}
Thus, the second term in~\eqref{eq:1-gamma-beta} is summable and can never cause divergence. Hence
\begin{equation}\label{eq:crit-reduce}
  \sum_{t=1}^\infty(1-\gamma_t)=\infty
  \iff
  \sum_{t=1}^\infty\beta_t(1-\alpha_t)K_t=\infty .
\end{equation}
We bound $K_t$ on both sides by a multiple of $N_{t+1}$. Since $N_t \ge 0$ and $\alpha_t\ge0$,
\begin{equation*}
    K_t=2N_t+n_b(t)(1+\alpha_t) \ge N_t+n_b(t)=N_{t+1},
\end{equation*}
and since $\alpha_t\le1$,
\begin{equation*}
  K_t \le 2N_t+2n_b(t)=2N_{t+1}.
\end{equation*}
That is, 
\begin{equation*}
    (1-\alpha_t) \beta_tN_{t+1} \le \beta_t(1-\alpha_t)K_t \le 2(1-\alpha_t) \beta_tN_{t+1}.
\end{equation*}
By \eqref{eq:counts}, we have $\beta_tN_{t+1} = \frac{n_b(t)}{N_{t+1}} = 1 - \frac{N_t}{N_{t+1}} = 1 - \rho_t$. Thus:
\begin{equation*}
    (1-\alpha_t)(1-\rho_t) \le \beta_t(1-\alpha_t)K_t \le 2 (1-\alpha_t)(1-\rho_t).
\end{equation*}
The two series therefore diverge together, which with~\eqref{eq:crit-reduce} gives (c)$\Leftrightarrow$(d). 
\end{proof}

\subsubsection{Consistency of $\boldsymbol{\Sigma}^{(t)}$}
\label{sec:Sigma(T)->Sigma in F norm cond}
Next, we characterize when $\mathbb{E}\!\left[\|\boldsymbol{\Sigma}^{(t)}-\boldsymbol{\Sigma}\|_F\right] \xrightarrow{t\rightarrow\infty} 0$. We start with eight technical lemmas. First, we denote the Gaussian fourth-moment constant
\begin{equation}\label{eq:CSigma-def}
  C_{\boldsymbol{\Sigma}} \coloneqq \Tr(\boldsymbol{\Sigma})^2+2\|\boldsymbol{\Sigma}\|_F^2.
\end{equation}
It equals $\mathbb{E}\|\boldsymbol{g}\|_2^4$ for $\boldsymbol{g}\sim\mathcal{N}(0,\boldsymbol{\Sigma})$: by Isserlis' theorem $\mathbb{E}[g_i^2g_j^2]=\boldsymbol{\Sigma}_{ii}\boldsymbol{\Sigma}_{jj}+2\boldsymbol{\Sigma}_{ij}^2$, so
$\mathbb{E}\|\boldsymbol{g}\|_2^4=\sum_{i,j}(\boldsymbol{\Sigma}_{ii}\boldsymbol{\Sigma}_{jj}+2\boldsymbol{\Sigma}_{ij}^2)
=\Tr(\boldsymbol{\Sigma})^2+2\|\boldsymbol{\Sigma}\|_F^2.$

First, we isolate the dependence of the covariance estimator on the mean error.

\begin{lemma}[Real-centered second-moment statistic]\label{lem:Shat-props}
Define
\begin{equation}
\label{eq:Shat-def}
  \hat{\boldsymbol{S}}_t \coloneqq \frac{1}{N_t}\Biggl(\sum_{j=1}^{n_r}(\boldsymbol{x}^{(0)}_j-\boldsymbol{\mu})(\boldsymbol{x}^{(0)}_j-\boldsymbol{\mu})^\top
   +\sum_{s=1}^{t-1}\sum_{i=1}^{n_b(s)}(\boldsymbol{x}^{(s)}_i-\boldsymbol{\mu})(\boldsymbol{x}^{(s)}_i-\boldsymbol{\mu})^\top\Biggr).
\end{equation}
Then, using \eqref{eq:delta-def} $(\boldsymbol{\delta}^{(t)} = \boldsymbol{\mu}^{(t)} - \boldsymbol{\mu})$:
\begin{enumerate}
  \item[(i)] $\hat{\boldsymbol{S}}_t\succeq0$ almost surely;
  \item[(ii)] $\boldsymbol{\Sigma}^{(t)}=\hat{\boldsymbol{S}}_t-\boldsymbol{\delta}^{(t)}(\boldsymbol{\delta}^{(t)})^\top$. Thus,
    $\Tr(\boldsymbol{\Sigma}^{(t)})=\Tr(\hat{\boldsymbol{S}}_t)-\|\boldsymbol{\delta}^{(t)}\|_2^2$;
  \item[(iii)] for $t\ge2$,
    $\displaystyle\hat{\boldsymbol{S}}_t=\rho_{t - 1}\hat{\boldsymbol{S}}_{t-1}
      +\frac{1}{N_t}\sum_{i=1}^{n_b(t-1)}(\boldsymbol{x}^{(t-1)}_i-\boldsymbol{\mu})(\boldsymbol{x}^{(t-1)}_i-\boldsymbol{\mu})^\top$;
  \item[(iv)] for $t \ge 2$, $\mathbb{E}[\hat{\boldsymbol{S}}_t\mid\mathcal{F}_{t-1}] =\lambda_{t-1}\hat{\boldsymbol{S}}_{t-1}+(1-\lambda_{t-1})\boldsymbol{\Sigma}$, and consequently $\mathbb{E}[\hat{\boldsymbol{S}}_t]=\boldsymbol{\Sigma}$ for all $t\ge1$.
\end{enumerate}
\end{lemma}

\begin{proof}
\emph{(i)} $\hat{\boldsymbol{S}}_t$ is a non-negatively weighted sum of outer products. Hence, it is symmetric PSD.

\smallskip\noindent\emph{(ii)} Using the pooled-scatter identity~\eqref{eq: pooled scatter identity}
and recalling \eqref{eq:delta-def} ($\boldsymbol{\delta}^{(t)}=\boldsymbol{\mu}^{(t)}-\boldsymbol{\mu}$) gives
$\boldsymbol{\Sigma}^{(t)}=\hat{\boldsymbol{S}}_t-\boldsymbol{\delta}^{(t)}(\boldsymbol{\delta}^{(t)})^\top$; the trace identity follows from
$\Tr(\boldsymbol{\delta}^{(t)}(\boldsymbol{\delta}^{(t)})^\top)=\|\boldsymbol{\delta}^{(t)}\|_2^2$.

\smallskip\noindent\emph{(iii)} Fix $t \ge 2$. From \eqref{eq:Shat-def}, we have
\begin{align*}
    \hat{\boldsymbol{S}}_t &= \frac{1}{N_t}\Biggl(\sum_{j=1}^{n_r}(\boldsymbol{x}^{(0)}_j-\boldsymbol{\mu})(\boldsymbol{x}^{(0)}_j-\boldsymbol{\mu})^\top
    +\sum_{s=1}^{t-1}\sum_{i=1}^{n_b(s)}(\boldsymbol{x}^{(s)}_i-\boldsymbol{\mu})(\boldsymbol{x}^{(s)}_i-\boldsymbol{\mu})^\top\Biggr)
    \\
    &= \frac{1}{N_t}\Biggl(\sum_{j=1}^{n_r}(\boldsymbol{x}^{(0)}_j-\boldsymbol{\mu})(\boldsymbol{x}^{(0)}_j-\boldsymbol{\mu})^\top + \sum_{s=1}^{t-2}\sum_{i=1}^{n_b(s)}(\boldsymbol{x}^{(s)}_i-\boldsymbol{\mu})(\boldsymbol{x}^{(s)}_i-\boldsymbol{\mu})^\top + \sum_{i=1}^{n_b(t-1)} (\boldsymbol{x}^{(t-1)}_i-\boldsymbol{\mu})(\boldsymbol{x}^{(t-1)}_i-\boldsymbol{\mu})^\top\Biggr)
    \\
    &=\frac{N_{t-1}}{N_t}\hat{\boldsymbol{S}}_{t-1} + \frac{1}{N_t}\sum_{i=1}^{n_b(t-1)} (\boldsymbol{x}^{(t-1)}_i-\boldsymbol{\mu})(\boldsymbol{x}^{(t-1)}_i-\boldsymbol{\mu})^\top
    \\
    &=\rho_{t - 1}\hat{\boldsymbol{S}}_{t-1} + \frac{1}{N_t}\sum_{i=1}^{n_b(t-1)} (\boldsymbol{x}^{(t-1)}_i-\boldsymbol{\mu})(\boldsymbol{x}^{(t-1)}_i-\boldsymbol{\mu})^\top
\end{align*}
We used the definition of $\rho_t$ from \eqref{eq:counts}.

\smallskip\noindent\emph{(iv)} From (iii), we have
\begin{align*}
    \mathbb{E}\bigl[\hat{\boldsymbol{S}}_t \mid \mathcal{F}_{t-1}\bigr] &= \rho_{t - 1}\mathbb{E}\bigl[\hat{\boldsymbol{S}}_{t-1} \mid \mathcal{F}_{t-1}\bigr] + \frac{1}{N_t}\sum_{i=1}^{n_b(t-1)} \mathbb{E}\bigl[(\boldsymbol{x}^{(t-1)}_i-\boldsymbol{\mu})(\boldsymbol{x}^{(t-1)}_i-\boldsymbol{\mu})^\top \mid \mathcal{F}_{t - 1} \bigr]
    \\
    &= \rho_{t - 1}\hat{\boldsymbol{S}}_{t - 1} + \frac{1}{N_t}\sum_{i=1}^{n_b(t - 1)} \bigl(\alpha_{t-1}\boldsymbol{\Sigma}^{(t-1)} + (1-\alpha_{t-1})\boldsymbol{\Sigma} + \alpha_{t-1}\boldsymbol{\delta}^{(t-1)}(\boldsymbol{\delta}^{(t-1)})^\top\bigr)
    \\
    &= \frac{N_{t-1}}{N_t}\hat{\boldsymbol{S}}_{t-1} + \frac{1}{N_t}\sum_{i=1}^{n_b(t-1)} \bigl(\alpha_{t-1}\hat{\boldsymbol{S}}_{t-1} + (1-\alpha_{t-1})\boldsymbol{\Sigma}\bigr)
    \\
    &= \frac{N_{t-1} + \alpha_{t-1} n_b(t-1)}{N_t}\hat{\boldsymbol{S}}_{t-1} + \frac{(1-\alpha_{t-1})n_b(t-1)}{N_t}\boldsymbol{\Sigma}
    \\
    &= \lambda_{t-1}\hat{\boldsymbol{S}}_{t-1} + (1 - \lambda_{t-1})\boldsymbol{\Sigma}.
\end{align*}

We used 
\begin{itemize}

    \item $\hat{\boldsymbol{S}}_{t-1}$ is $\mathcal{F}_{t-1}$-measurable.

    \item $\rho_{t-1} = N_{t-1}/N_t$ from \eqref{eq:counts}.

    \item From \eqref{eq:cond-sq} (Lemma~\ref{lem:cond-moments}), we have
    \begin{equation*}
        \mathbb{E}\bigl[(\boldsymbol{x}^{(t-1)}_i-\boldsymbol{\mu})(\boldsymbol{x}^{(t-1)}_i-\boldsymbol{\mu})^\top\mid\mathcal{F}_{t-1}\bigr]
        =\alpha_{t-1}\boldsymbol{\Sigma}^{(t-1)}+(1-\alpha_{t-1})\boldsymbol{\Sigma}
        +\alpha_{t-1}\boldsymbol{\delta}^{(t-1)}(\boldsymbol{\delta}^{(t-1)})^\top
    \end{equation*}
    
    \item $\boldsymbol{\Sigma}^{(t-1)}+\boldsymbol{\delta}^{(t-1)}(\boldsymbol{\delta}^{(t-1)})^\top
    =\hat{\boldsymbol{S}}_{t-1}$ from (ii) at index $t-1$.

    \item The definition of $\lambda_t$ from \eqref{eq:scalars}, and
    \begin{equation*}
        1 - \lambda_{t} = 1 - \frac{N_t + n_b(t)\alpha_t}{N_{t+1}} = \frac{N_{t+1} - N_t - n_b(t)\alpha_t}{N_{t+1}} = \frac{\left(1 - \alpha_t\right)n_b(t)}{N_{t+1}}.
    \end{equation*}
    
\end{itemize}
That is, we proved that
\begin{equation}
\label{eq: E(S_T | F_T-1) final formula}    \mathbb{E}\bigl[\hat{\boldsymbol{S}}_t \mid \mathcal{F}_{t-1}\bigr] = \lambda_{t-1}\hat{\boldsymbol{S}}_{t-1} + (1 - \lambda_{t-1})\boldsymbol{\Sigma}.
\end{equation}

Finally, we prove unbiasedness by induction on $t \ge 1$. At $t = 1$, we have
\begin{equation*}
    \hat{\boldsymbol{S}}_1 = \frac{1}{n_r}\sum_{j=1}^{n_r}(\boldsymbol{x}^{(0)}_j-\boldsymbol{\mu})(\boldsymbol{x}^{(0)}_j-\boldsymbol{\mu})^\top \Rightarrow \mathbb{E}[\hat{\boldsymbol{S}}_1] = \frac{1}{n_r}\sum_{j=1}^{n_r}\mathbb{E}\bigl[(\boldsymbol{x}^{(0)}_j-\boldsymbol{\mu})(\boldsymbol{x}^{(0)}_j-\boldsymbol{\mu})^\top\bigr] = \boldsymbol{\Sigma}
\end{equation*}
Assume that $\mathbb{E}[\hat{\boldsymbol{S}}_{t-1}] = \boldsymbol{\Sigma}$. From \eqref{eq: E(S_T | F_T-1) final formula} and the tower property, we have
\begin{align*}
    \mathbb{E}[\hat{\boldsymbol{S}}_t] &= \mathbb{E}\bigl[\mathbb{E}[\hat{\boldsymbol{S}}_t \mid \mathcal{F}_{t - 1}]\bigr] 
    \\
    &= \mathbb{E}[\lambda_{t - 1}\hat{\boldsymbol{S}}_{t-1} + (1-\lambda_{t-1})\boldsymbol{\Sigma}] 
    \\
    &= \lambda_{t-1}\mathbb{E}[\hat{\boldsymbol{S}}_{t-1}] + (1- \lambda_{t-1})\boldsymbol{\Sigma} 
    \\
    &= \boldsymbol{\Sigma}. 
\end{align*}
\end{proof}

We now state a technical lemma that will be helpful later on.

\begin{lemma}[Rayleigh bound]
\label{lem: rayleigh_bound}
Let $\boldsymbol{A}\in\mathbb{R}^{d\times d}$ be symmetric with $\boldsymbol{A}\succeq0$.
Then for every $\boldsymbol{x}\in\mathbb{R}^d$,
\begin{equation*}
    \boldsymbol{x}^\top\boldsymbol{A}\boldsymbol{x}\;\le\;\Tr(\boldsymbol{A})\,\|\boldsymbol{x}\|_2^2 .
\end{equation*}
\end{lemma}

\begin{proof}
By the spectral theorem, $\boldsymbol{A}=\boldsymbol{Q}\boldsymbol{D}\boldsymbol{Q}^\top$
with $\boldsymbol{Q}$ orthogonal and $\boldsymbol{D}=\operatorname{diag}(\nu_1,\dots,\nu_d)$,
where $\nu_1,\dots,\nu_d$ are the eigenvalues of $\boldsymbol{A}$: real because
$\boldsymbol{A}$ is symmetric, and nonnegative because $\boldsymbol{A}\succeq0$, since
$\nu_i=\boldsymbol{q}_i^\top\boldsymbol{A}\boldsymbol{q}_i\ge0$ for the corresponding
unit eigenvector $\boldsymbol{q}_i$.

Fix $\boldsymbol{x}\in\mathbb{R}^d$ and set $\boldsymbol{y} \coloneqq \boldsymbol{Q}^\top\boldsymbol{x}$.
Since $\boldsymbol{Q}$ is orthogonal, $\|\boldsymbol{y}\|_2=\|\boldsymbol{x}\|_2$, and
\begin{equation*}
    \boldsymbol{x}^\top\boldsymbol{A}\boldsymbol{x}
    =\boldsymbol{x}^\top\boldsymbol{Q}\boldsymbol{D}\boldsymbol{Q}^\top\boldsymbol{x}
    =\boldsymbol{y}^\top\boldsymbol{D}\boldsymbol{y}
    =\sum_{i=1}^{d}\nu_i\,y_i^2 .
\end{equation*}
Every $y_i^2\ge0$, so bounding each $\nu_i$ by
$\nu_{\max} \coloneqq \max_{1\le j\le d}\nu_j$ gives
\begin{equation*}
    \sum_{i=1}^{d}\nu_i\,y_i^2
    \;\le\;\nu_{\max}\sum_{i=1}^{d}y_i^2
    \;=\;\nu_{\max}\,\|\boldsymbol{y}\|_2^2
    \;=\;\nu_{\max}\,\|\boldsymbol{x}\|_2^2 .
\end{equation*}
Finally, all eigenvalues are nonnegative, so discarding the remaining $d-1$ of them
from the trace yields
\begin{equation*}
    \nu_{\max}\;\le\;\sum_{i=1}^{d}\nu_i\;=\;\Tr(\boldsymbol{A}),
\end{equation*}
and combining the last two displays gives the claim.
\end{proof}

\begin{lemma}[Conditional fourth-moment bound]\label{lem:fourth-moment}
For every $t\ge1$ and each newly drawn sample $\boldsymbol{x}^{(t)}_i$,
\begin{equation}\label{eq:fourth-bound}
  \mathbb{E}\bigl[\|\boldsymbol{x}^{(t)}_i-\boldsymbol{\mu}\|_2^4\mid\mathcal{F}_t\bigr]
  \le 3\alpha_t\,\Tr(\hat{\boldsymbol{S}}_t)^2+C_{\boldsymbol{\Sigma}} ,
\end{equation}
while for the real batch $\mathbb{E}\bigl[\|\boldsymbol{x}^{(0)}_j-\boldsymbol{\mu}\|_2^4\bigr]=C_{\boldsymbol{\Sigma}}$.
\end{lemma}

\begin{proof}  Given $\mathcal{F}_t$, the sample $\boldsymbol{x}^{(t)}_i$ is drawn from the mixture~\eqref{eq:mixture}, so from the total expectation theorem
\begin{equation}
\label{eq:fourth-mixture}
  \mathbb{E}\bigl[\|\boldsymbol{x}^{(t)}_i-\boldsymbol{\mu}\|_2^4\mid\mathcal{F}_t\bigr]
  = \alpha_t\,\mathbb{E}\bigl[\|\boldsymbol{Y}^{(t)} -\boldsymbol{\mu}\|_2^4\mid\mathcal{F}_t\bigr]+(1-\alpha_t)\,C_{\boldsymbol{\Sigma}},
\end{equation}
where $\boldsymbol{Y}^{(t)}\sim\mathcal{N}(\boldsymbol{\mu}^{(t)},\boldsymbol{\Sigma}^{(t)})$ given $\mathcal{F}_t$, and the real
component contributes $\mathbb{E}\!\bigl[\|\boldsymbol{W}-\boldsymbol{\mu}\|_2^4\bigr]=C_{\boldsymbol{\Sigma}}$ for
$\boldsymbol{W}\sim\mathcal{N}(\boldsymbol{\mu},\boldsymbol{\Sigma})$ (by~\eqref{eq:CSigma-def}). It remains to bound the
synthetic term.

From \eqref{eq:delta-def}, we have $\boldsymbol{\delta}^{(t)} = \boldsymbol{\mu}^{(t)} - \boldsymbol{\mu}$ is $\mathcal{F}_t$-measurable. Thus, write $\boldsymbol{Y}^{(t)} -\boldsymbol{\mu}=\boldsymbol{y}^{(t)} + \boldsymbol{\delta}^{(t)}$ with $\boldsymbol{y}^{(t)} \coloneqq \boldsymbol{Y}^{(t)} -\boldsymbol{\mu}^{(t)}\sim\mathcal{N}(\boldsymbol{0},\boldsymbol{\Sigma}^{(t)})$
given $\mathcal{F}_t$. Since $\boldsymbol{y}^{(t)}$ is
conditionally symmetric about $0$, the odd terms in $\boldsymbol{y}^{(t)}$ vanish, and
\begin{equation}
\label{eq:Y-expand}
\begin{aligned}
  \mathbb{E}[\|\boldsymbol{Y}^{(t)} - \boldsymbol{\mu}\|_2^4 \mid \mathcal{F}_t] &=  \mathbb{E}[\|\boldsymbol{y}^{(t)} + \boldsymbol{\delta}^{(t)}\|_2^4\mid\mathcal{F}_t]
  \\
  &= \mathbb{E}[\|\boldsymbol{y}^{(t)}\|_2^4\mid\mathcal{F}_t] + 4\,\mathbb{E}\!\left[\left((\boldsymbol{y}^{(t)})^\top\boldsymbol{\delta}^{(t)}\right)^2 \ \Big| \ \mathcal{F}_t\right] + 2\|\boldsymbol{\delta}^{(t)}\|_2^2\,\mathbb{E}[\|\boldsymbol{y}^{(t)}\|_2^2\mid\mathcal{F}_t]+\|\boldsymbol{\delta}^{(t)}\|_2^4
  \\
  &= \mathbb{E}[\|\boldsymbol{y}^{(t)}\|_2^4\mid\mathcal{F}_t] + 4\,\mathbb{E}\!\left[\left((\boldsymbol{y}^{(t)})^\top\boldsymbol{\delta}^{(t)}\right)^2 \ \Big| \ \mathcal{F}_t\right] + 2p\cdot \mathbb{E}[\|\boldsymbol{y}^{(t)}\|_2^2\mid\mathcal{F}_t] + p^2,
\end{aligned}
\end{equation}
where we denote $u \coloneqq \Tr(\hat{\boldsymbol{S}}_t)$ and $p \coloneqq \|\boldsymbol{\delta}^{(t)}\|_2^2$. In order to further calculate and bound the above expression, we will use $\boldsymbol{y}^{(t)} \mid \mathcal{F}_t \sim \mathcal{N}(\boldsymbol{0},\boldsymbol{\Sigma}^{(t)})$, and use $\Tr(\boldsymbol{\Sigma}^{(t)}) = \Tr(\hat{\boldsymbol{S}}_t) - \|\boldsymbol{\delta}^{(t)}\|_2^2 = u - p$ (Lemma~\ref{lem:Shat-props}(ii)).

First, conditionally $\boldsymbol{\Sigma}^{(t)}$ is fixed, so by~\eqref{eq:CSigma-def} applied to $\boldsymbol{\Sigma}^{(t)}$,
\begin{equation*}
    \mathbb{E}[\|\boldsymbol{y}^{(t)}\|_2^4\mid\mathcal{F}_t]=\Tr(\boldsymbol{\Sigma}^{(t)})^2+2\|\boldsymbol{\Sigma}^{(t)}\|_F^2
    \le3\Tr(\boldsymbol{\Sigma}^{(t)})^2 = 3(u-p)^2,
\end{equation*}
the inequality using $\|\boldsymbol{A}\|_F^2\le\Tr(\boldsymbol{A})^2$ for symmetric positive-semi-definite $\boldsymbol{A}$ (Lemma~\ref{lem:trace-frobenius}(a), applied to
$\boldsymbol{\Sigma}^{(t)}\succeq0$ by Lemma~\ref{lem:sample covariance is PSD}).

Second, $\mathbb{E}[ \|\boldsymbol{y}^{(t)}\|_2^2 \mid \mathcal{F}_t]=\Tr(\boldsymbol{\Sigma}^{(t)}) = u - p$.

Finally, 
\begin{align*}
    \mathbb{E}\!\left[\left((\boldsymbol{y}^{(t)})^\top\boldsymbol{\delta}^{(t)}\right)^2 \ \Big| \ \mathcal{F}_t\right] &= \mathbb{E}\!\left[\left(\boldsymbol{\delta}^{(t)}\right)^\top\boldsymbol{y}^{(t)}\left(\boldsymbol{y}^{(t)}\right)^\top\boldsymbol{\delta}^{(t)} \ \Big| \ \mathcal{F}_t\right]
    \\
    &= (\boldsymbol{\delta}^{(t)})^\top \mathbb{E}\!\left[\boldsymbol{y}^{(t)}\left(\boldsymbol{y}^{(t)}\right)^\top \Big| \ \mathcal{F}_t \right]\boldsymbol{\delta}^{(t)}
    \\
    &= \left(\boldsymbol{\delta}^{(t)}\right)^\top\boldsymbol{\Sigma}^{(t)}\boldsymbol{\delta}^{(t)}
    \\
    &= \left(\boldsymbol{\delta}^{(t)}\right)^\top \hat{\boldsymbol{S}}_t \boldsymbol{\delta}^{(t)} - \|\boldsymbol{\delta}^{(t)}\|_2^4
    \\
    &\le \Tr(\hat{\boldsymbol{S}}_t)\|\boldsymbol{\delta}^{(t)}\|_2^2 - \|\boldsymbol{\delta}^{(t)}\|_2^4
    \\
    &= up - p^2
    \\
    &= p(u - p).
\end{align*}
We used $\boldsymbol{\Sigma}^{(t)}=\hat{\boldsymbol{S}}_t-\boldsymbol{\delta}^{(t)}\left(\boldsymbol{\delta}^{(t)}\right)^\top$
(Lemma~\ref{lem:Shat-props}(ii)), and Lemma~\ref{lem: rayleigh_bound} with $\hat{\boldsymbol{S}}_t \succeq 0$ by Lemma \ref{lem:Shat-props}(i). Collecting terms in~\eqref{eq:Y-expand},
\begin{align*}
    \mathbb{E}[\|\boldsymbol{Y}^{(t)} -\boldsymbol{\mu}\|_2^4 \mid \mathcal{F}_t]
    &= \mathbb{E}[\|\boldsymbol{y}^{(t)}\|_2^4\mid\mathcal{F}_t] + 4\,\mathbb{E}\!\left[\left((\boldsymbol{y}^{(t)})^\top\boldsymbol{\delta}^{(t)}\right)^2 \ \Big| \ \mathcal{F}_t\right] + 2p\cdot \mathbb{E}[\|\boldsymbol{y}^{(t)}\|_2^2\mid\mathcal{F}_t] + p^2
    \\
    &\le 3(u - p)^2 + 4p(u - p) + 2p(u - p) + p^2
    \\
    &= 3(u - p)^2 + 6p(u - p) + p^2
    \\
    &= 3u^2 - 3p^2 + p^2
    \\
    &= 3u^2 - 2p^2
    \\
    &\le 3u^2
    \\
    &= 3\Tr(\hat{\boldsymbol{S}}_t)^2.
\end{align*}
We used $-2p^2 \le 0$, and
\begin{equation*}
    3(u - p)^2 + 6p(u - p) = 3(u-p)\left(u - p + 2p\right) = 3(u - p)(u + p) = 3u^2 - 3p^2.
\end{equation*}
Substituting
into~\eqref{eq:fourth-mixture} and bounding $1-\alpha_t\le1$ gives~\eqref{eq:fourth-bound}.
\end{proof}

\begin{lemma}[Recursion unrolling]\label{lem:rec-unroll-frob}
Let $\{a_t\}_{t\ge1}$, $\{r_t\}_{t\ge1}$, $\{q_t\}_{t\ge1}$ be nonnegative sequences with
$a_t\le r_{t-1}a_{t-1}+q_{t-1}$ for all $t\ge2$. Then for every $t\ge2$,
\begin{equation*}
    a_t \le \left(\prod_{k=1}^{t-1}r_k \right) a_1 + \sum_{s=1}^{t-1}\left(\prod_{k=s+1}^{t-1}r_k\right) q_s,
\end{equation*}
with the empty product equal to $1$.
\end{lemma}

\begin{proof}
We prove by induction on $t$. The base case $t = 2$ is the hypothesis $a_2 \le r_1a_1 + q_1$. Assuming the bound at
$t - 1$, we get
\begin{align*}
    a_t\le r_{t-1}a_{t-1} + q_{t-1}
    &\le r_{t-1} \left[\left(\prod_{k = 1}^{t-2}r_k\right) a_1
    + \sum_{s=1}^{t-2}\left(\prod_{k=s+1}^{t-2}r_k\right) q_s\right] + q_{t-1}
    \\
    &= \left(\prod_{k=1}^{t-1}r_k\right) a_1
    + \sum_{s=1}^{t-2}\left(\prod_{k=s+1}^{t-1}r_k\right) q_s + q_{t - 1}
    \\
    &= \left(\prod_{k=1}^{t-1}r_k\right) a_1
    + \sum_{s=1}^{t-1}\left(\prod_{k=s+1}^{t-1}r_k\right) q_s,
\end{align*}
where $q_{t-1}$ is the $s=t-1$ term (empty product $=1$).
\end{proof}

\begin{lemma}[Uniform bound on the amplification products]\label{lem:uniform-product-bound}
Let $r_t \coloneqq \lambda_t^2+3\alpha_t\beta_t$ for $t\ge1$, with $\lambda_t,\beta_t$ as
in~\eqref{eq:scalars}. Then $r_t\ge0$ for every $t \ge 1$, and for all integers $1\le a\le b<\infty$,
\begin{equation}\label{eq:Rstarstar}
  \prod_{t=a}^{b}r_t \le \;\exp\!\left(\frac{3}{n_r}\right) < \infty .
\end{equation}
In particular the bound is uniform in $a$ and $b$, and holds for \emph{every}
$\{\alpha_t\}\subset[0,1]$ and $\{n_b(t)\ge1\}$, with no further hypothesis.
\end{lemma}

\begin{proof} Fix $t\ge1$. By Lemma~\ref{lem:range} we have $\lambda_t\in(0,1]$,
hence $\lambda_t^2\le1$; and $\alpha_t\in[0,1]$ gives $3\alpha_t\beta_t\le3\beta_t$. Adding,
\begin{equation}\label{eq:r-pointwise}
  0\;\le\;r_t\;\le\;1+3\beta_t ,
\end{equation}
the lower bound because both summands defining $r_t$ are nonnegative. Fix integers $1 \le a \le b$. From \eqref{eq:r-pointwise}, we have
\begin{align*}
    \prod_{t=a}^b r_t \le \prod_{t = a}^b \left(1 + 3\beta_t\right) &\le \prod_{t=1}^\infty (1 + 3\beta_t)
    \\
    &= \exp\left(\sum_{t=1}^\infty \ln(1+3\beta_t)\right)
    \\
    &\le \exp\left(3\sum_{t=1}^\infty \beta_t \right)
    \\
    &\le \exp\left(\frac{3}{n_r}\right)
\end{align*}
Where we used $1+3\beta_t \ge 1$, $\ln(1+x) \le x$ for $x \ge 0$, and $\sum_{t=1}^\infty \beta_t \le 1/n_r$ from Lemma \ref{lem:beta-summability}.
\end{proof}

\begin{lemma}[Uniform second-moment bound on $\hat{\boldsymbol{S}}_t$]\label{lem:uniform-L2-Shat}
Under Assumption~\ref{ass:cond-ind},
\begin{equation*}
    B^\ast \coloneqq \sup_{t \ge 1} \mathbb{E}\bigl[\Tr(\hat{\boldsymbol{S}}_t)^2\bigr] < \infty.
\end{equation*}
\end{lemma}

\begin{proof}
Write $u_t \coloneqq \Tr(\hat{\boldsymbol{S}}_t)$, and denote
\begin{equation}
\label{eq:v_T def}
    v_t \coloneqq \Var(u_t) = \mathbb{E}\bigl[\Tr(\hat{\boldsymbol{S}}_t)^2\bigr] - \Tr(\boldsymbol{\Sigma})^2.
\end{equation}
We used $\mathbb{E}[u_t] = \mathbb{E}\!\bigl[\Tr(\hat{\boldsymbol{S}}_t)\bigr] = \Tr(\mathbb{E}[\hat{\boldsymbol{S}}_t]) = \Tr(\boldsymbol{\Sigma})$
(Lemma~\ref{lem:Shat-props}(iv)).  We establish a recursion on $u_t$. From Lemma~\ref{lem:Shat-props}(iii), we have for every $t \ge 2$
\begin{equation}
\label{eq:u_T rec}
\begin{aligned}
    u_t &= \rho_{t - 1}u_{t-1} + \frac{1}{N_t}\sum_{i=1}^{n_b(t-1)} \Tr\bigl((\boldsymbol{x}^{(t-1)}_i - \boldsymbol{\mu})(\boldsymbol{x}^{(t-1)}_i - \boldsymbol{\mu})^\top\bigr)
    \\
    &= \rho_{t - 1}u_{t-1} + \frac{1}{N_t}\sum_{i=1}^{n_b(t-1)} \|\boldsymbol{x}^{(t - 1)}_i - \boldsymbol{\mu}\|_2^2
\end{aligned}
\end{equation}
and $u_1 = \frac{1}{n_r}\sum_{j=1}^{n_r} \|\boldsymbol{x}^{(0)}_j - \boldsymbol{\mu}\|_2^2$. From \eqref{eq:v_T def}, we need to prove that $\sup_{t \ge 1} v_t < \infty$. We will use the law of total variance, which presupposes $u_t \in L^2$. We verify that by induction on $t \ge 1$.

\medskip\noindent\emph{Square-integrability at every finite $t$.} For $t = 1$, the samples $\{\boldsymbol{x}_i^{(0)}\}_{i=1}^{n_r}$ are i.i.d. from $\mathcal{N}(\boldsymbol{\mu},\boldsymbol{\Sigma})$, and thus
\begin{equation}
\label{eq: v1 comp proof unif L2 bound}
    v_1 = \frac{1}{n_r}\Var(\|\boldsymbol{x}^{(0)}_1 - \boldsymbol{\mu}\|_2^2) \le \frac{1}{n_r}\mathbb{E}\!\bigl[\|\boldsymbol{x}^{(0)}_1 - \boldsymbol{\mu}\|_2^4\bigr] = \frac{1}{n_r}C_{\boldsymbol{\Sigma}} < \infty,
\end{equation}
where we used $\Var(X) \le \mathbb{E}[X^2]$. From \eqref{eq:v_T def}, we have $\mathbb{E}[u_1^2] = v_1 + \Tr(\boldsymbol{\Sigma})^2 < \infty$, thus $u_1 \in L^2$. Assume $u_{t-1} \in L^2$, then
\begin{align*}
    \mathbb{E}[u_t^2] &\le 2\!\left(\rho_{t - 1}^2 \mathbb{E}[u_{t-1}^2] + \frac{n_b(t-1)}{N_t^2}\sum_{i=1}^{n_b(t-1)} \mathbb{E}\!\bigl[\|\boldsymbol{x}^{(t - 1)}_i - \boldsymbol{\mu}\|_2^4\bigr]\right) 
    \\
    &\le 2\!\left(\rho_{t - 1}^2 \mathbb{E}\!\bigl[u_{t-1}^2\bigr] + \frac{n_b(t-1)}{N_t^2}\sum_{i=1}^{n_b(t-1)} (3\alpha_{t-1} \mathbb{E}\!\bigl[u_{t - 1}^2\bigr] + C_{\boldsymbol{\Sigma}})\right) \\
    & < \infty,
\end{align*}
where the first inequality applies Lemma~\ref{lem:l1-l2} twice: with $n=2$ to the two summands of~\eqref{eq:u_T rec}, i.e., $(a+b)^2\le2(a^2+b^2)$, and with $n=n_b(t-1)$ to the sum $\sum_{i=1}^{n_b(t-1)}\|\boldsymbol{x}^{(t-1)}_i-\boldsymbol{\mu}\|_2^2$, which gives
\begin{equation*}
    \left(\sum_{i=1}^{n_b(t-1)}\|\boldsymbol{x}^{(t-1)}_i-\boldsymbol{\mu}\|_2^2\right)^2 \le n_b(t-1)\sum_{i=1}^{n_b(t-1)}\|\boldsymbol{x}^{(t-1)}_i-\boldsymbol{\mu}\|_2^4.
\end{equation*}
The second inequality uses the following bound: from Lemma~\ref{lem:fourth-moment} and the tower property, we have
\begin{equation*}
    \mathbb{E}\!\bigl[\|\boldsymbol{x}^{(t - 1)}_i - \boldsymbol{\mu}\|_2^4\bigr] = \mathbb{E}\!\bigl[\mathbb{E}\!\bigl[\|\boldsymbol{x}^{(t - 1)}_i - \boldsymbol{\mu}\|_2^4 \mid \mathcal{F}_{t - 1}\bigr]\bigr] \le \mathbb{E}\!\bigl[3\alpha_{t-1}\Tr(\hat{\boldsymbol{S}}_{t-1})^2 + C_{\boldsymbol{\Sigma}}\bigr] = 3\alpha_{t-1}\mathbb{E}\!\bigl[u_{t-1}^2\bigr] + C_{\boldsymbol{\Sigma}}.
\end{equation*}

\medskip\noindent\emph{The law of total variance.} Fix $t \ge 2$. The law of total variance dictates that
\begin{equation}
\label{eq: vT-law of total variance}
    v_t = \Var\!\left(\mathbb{E}\!\bigl[u_t \mid \mathcal{F}_{t-1} \bigr]\right)  + \mathbb{E}\!\bigl[\Var\!\left(u_t \mid \mathcal{F}_{t-1} \right)\bigr].
\end{equation}
Taking the trace of Lemma~\ref{lem:Shat-props}(iv),
$\mathbb{E}[u_t\mid\mathcal{F}_{t-1}]=\lambda_{t-1}u_{t-1}+(1-\lambda_{t-1})\Tr(\boldsymbol{\Sigma})$. Thus, the first summand reads
\begin{equation}
\label{eq: var of cond exp u_T|F_(T-1)}
    \Var\bigl(\mathbb{E}[u_t\mid\mathcal{F}_{t-1}]\bigr)=\lambda_{t-1}^2\,v_{t-1}.
\end{equation}
For the second summand, we use \eqref{eq:u_T rec}:
\begin{equation}
\label{eq:conditional u var}
\begin{aligned}
  \Var(u_t\mid\mathcal{F}_{t-1})
  &=\frac{n_b(t-1)}{N_t^2}\Var\bigl(\|\boldsymbol{x}^{(t-1)}_1 -\boldsymbol{\mu}\|_2^2\mid\mathcal{F}_{t-1}\bigr)
   \le\beta_{t-1}\,\mathbb{E}\bigl[\|\boldsymbol{x}^{(t-1)}_1 -\boldsymbol{\mu}\|_2^4\mid\mathcal{F}_{t-1}\bigr]\\
  &\le\beta_{t-1}\bigl(3\alpha_{t-1}u_{t-1}^2+C_{\boldsymbol{\Sigma}}\bigr).
\end{aligned}
\end{equation}
We used that $u_{t-1}$ is $\mathcal{F}_{t-1}$-measurable, and conditionally on $\mathcal{F}_{t-1}$ the samples $\{\boldsymbol{x}^{(t-1)}_i\}_{i=1}^{n_b(t-1)}$ are i.i.d. (Assumption~\ref{ass:cond-ind}), $\Var \le $ second moment together with $\beta_t$ definition from \eqref{eq:scalars}, and Lemma~\ref{lem:fourth-moment}. From \eqref{eq: vT-law of total variance}, \eqref{eq: var of cond exp u_T|F_(T-1)} and \eqref{eq:conditional u var}, we have
\begin{equation}
\label{eq:v_T first rec}
\begin{aligned}
    v_t &\le \lambda_{t-1}^2v_{t-1} + \mathbb{E}\!\bigl[3\beta_{t-1}\alpha_{t-1}u_{t-1}^2 + \beta_{t-1}C_{\boldsymbol{\Sigma}}\bigr] 
    \\
    &= \left(\lambda_{t-1}^2 + 3\alpha_{t-1}\beta_{t-1}\right)v_{t-1} + \beta_{t-1}(3\alpha_{t-1} \Tr(\boldsymbol{\Sigma})^2 + C_{\boldsymbol{\Sigma}}),
\end{aligned}
\end{equation}
where we used $\mathbb{E}[u_{t-1}^2] = v_{t-1} + \Tr(\boldsymbol{\Sigma})^2$ from \eqref{eq:v_T def}. From \eqref{eq:v_T first rec}, we have:
\begin{equation}\label{eq:v-recursion}
  v_t\le r_{t-1}\,v_{t-1}+q_{t-1},\qquad
  \begin{cases}
    r_{t-1} \coloneqq \lambda_{t-1}^2+3\alpha_{t-1}\beta_{t-1},\\[2pt]
    q_{t-1} \coloneqq \beta_{t-1}\bigl(3\alpha_{t-1}\Tr(\boldsymbol{\Sigma})^2+C_{\boldsymbol{\Sigma}}\bigr).
  \end{cases}
\end{equation}

\medskip\noindent\emph{Uniform bound.} Applying Lemma~\ref{lem:rec-unroll-frob} to~\eqref{eq:v-recursion} and using Lemma~\ref{lem:uniform-product-bound}, we have
\begin{equation}
\label{eq:v_T final rec}
\begin{aligned}
    v_t &\le \left(\prod_{k=1}^{t-1} r_k\right)v_1 + \sum_{s=1}^{t-1}\left(\prod_{k=s+1}^{t-1} r_k\right)q_s
    \\
    &\le \exp\!\left(\frac{3}{n_r}\right)v_1 + \exp\!\left(\frac{3}{n_r}\right)\sum_{s=1}^{t-1} q_s 
    \\
    &\le \exp\!\left(\frac{3}{n_r}\right)(v_1 + D_0)
\end{aligned}
\end{equation}
uniformly in $t$. Using \eqref{eq:v_T first rec}, we defined
\begin{equation}
\label{eq:D_0 def}
\begin{aligned}
    D_0 \coloneqq \sum_{s=1}^\infty q_s &\le \left(3\Tr(\boldsymbol{\Sigma})^2 + C_{\boldsymbol{\Sigma}}\right)\sum_{t=1}^\infty \beta_t
    \\
    &\le \frac{1}{n_r}\left(C_{\boldsymbol{\Sigma}} + 3\Tr(\boldsymbol{\Sigma})^2\right).
\end{aligned}
\end{equation}
The first inequality follows from $\alpha_{t-1} \le 1$, and the second inequality follows from Lemma~\ref{lem:beta-summability}. Finally, from \eqref{eq:v_T def}, \eqref{eq:v_T final rec} and \eqref{eq:D_0 def}, we have
\begin{equation}
\label{eq:final sup Tr(ST^2) bound proof}
    \sup_{t \ge 1} \mathbb{E}\bigl[\Tr(\hat{\boldsymbol{S}}_t)^2\bigr] = \sup_{t \ge 1} \!\left(v_t + \Tr(\boldsymbol{\Sigma})^2\right) \le \exp\!\left(\frac{3}{n_r}\right)(v_1 + D_0) + \Tr(\boldsymbol{\Sigma})^2 < \infty.
\end{equation}
\end{proof}

\begin{lemma}[Divergence of $1-\lambda_t^2$]\label{lem:lambda2-div}
If $\sum_{t=1}^{\infty}(1-\alpha_t)(1-\rho_t) = \infty$, then
$\sum_{t=1}^{\infty}(1-\lambda_t^2) = \infty$.
\end{lemma}

\begin{proof}
From the definition of $\lambda_t$ \eqref{eq:scalars} and the definition of $\rho_t$ in \eqref{eq:counts}, we have
\begin{equation*}
    1 - \lambda_t = 1 - \frac{N_t + n_b(t)\alpha_t}{N_{t+1}} = \frac{N_{t+1} - N_t - n_b(t)\alpha_t}{N_{t+1}}
    = \frac{n_b(t)(1-\alpha_t)}{N_{t+1}} = (1-\rho_t)(1-\alpha_t),
\end{equation*}
where we used $N_{t+1} = N_t + n_b(t)$ and $\rho_t = N_t/N_{t+1} \Rightarrow 1 - \rho_t = n_b(t)/N_{t+1}$. As $\lambda_t\in(0,1]$
(Lemma~\ref{lem:range}), $1-\lambda_t^2=(1-\lambda_t)(1+\lambda_t)\ge1-\lambda_t$, thus
\begin{equation*}
    \sum_{t=1}^{\infty}(1-\lambda_t^2) \ge \sum_{t=1}^{\infty}(1-\lambda_t)
    = \sum_{t=1}^{\infty}(1-\alpha_t)(1-\rho_t)\;=\;\infty
\end{equation*}
by hypothesis.
\end{proof}

\begin{lemma}[A useful upper bound] \label{lem: useful_upper_bound} Let $\boldsymbol{X}$ be a random matrix with $\mathbb{E}[\|\boldsymbol{X}\|_F^2]< \infty$, and $\mathcal{G}$ be a sub-$\sigma$-algebra. Define $\boldsymbol{Z} \coloneqq \boldsymbol{X} - \mathbb{E}[\boldsymbol{X} \mid \mathcal{G}]$. Then,
\begin{equation*}
    \mathbb{E}[\|\boldsymbol{Z}\|_F^2 \mid \mathcal{G}] \le \mathbb{E}[\|\boldsymbol{X}\|_F^2 \mid \mathcal{G}].
\end{equation*}
\end{lemma}

\begin{proof} Indeed, we have
\begin{align*}
    \mathbb{E}[\|\boldsymbol{Z}\|_F^2 \mid \mathcal{G}] &= \mathbb{E}[\Tr(\boldsymbol{Z}^\top\boldsymbol{Z}) \mid \mathcal{G}]
    \\
    &= \Tr\!\left(\mathbb{E}[\boldsymbol{Z}^\top\boldsymbol{Z} \mid \mathcal{G}]\right)
    \\
    &= \Tr\!\left(\mathbb{E}[\left(\boldsymbol{X} - \mathbb{E}[\boldsymbol{X} \mid \mathcal{G}]\right)^\top\!\left(\boldsymbol{X} - \mathbb{E}[\boldsymbol{X} \mid \mathcal{G}]\right) \mid \mathcal{G}]\right)
    \\
    &= \Tr\!\left(\mathbb{E}\!\left[\boldsymbol{X}^\top\boldsymbol{X} \ \big| \ \mathcal{G}\right] - \mathbb{E}\!\left[\boldsymbol{X}^\top \mathbb{E}[\boldsymbol{X} \mid \mathcal{G}] \ \big| \ \mathcal{G} \right] - \mathbb{E}\!\left[\mathbb{E}[\boldsymbol{X} \mid \mathcal{G}]^\top \boldsymbol{X} \mid \mathcal{G} \right] + \mathbb{E}\!\left[\mathbb{E}[\boldsymbol{X} \mid \mathcal{G}]^\top \mathbb{E}[\boldsymbol{X} \mid \mathcal{G}] \mid \mathcal{G} \right]\right)
    \\
    &= \mathbb{E}[\Tr(\boldsymbol{X}^\top\boldsymbol{X}) \mid \mathcal{G}] - \Tr\!\left(\mathbb{E}[\boldsymbol{X} \mid \mathcal{G}]^\top \mathbb{E}[\boldsymbol{X} \mid \mathcal{G}] + \mathbb{E}[\boldsymbol{X} \mid \mathcal{G}]^\top \mathbb{E}[\boldsymbol{X} \mid \mathcal{G}]- \mathbb{E}[\boldsymbol{X} \mid \mathcal{G}]^\top \mathbb{E}[\boldsymbol{X} \mid \mathcal{G}]\right)
    \\
    &= \mathbb{E}[\Tr(\boldsymbol{X}^\top\boldsymbol{X}) \mid \mathcal{G}] - \Tr\!\left(\mathbb{E}[\boldsymbol{X} \mid \mathcal{G}]^\top \mathbb{E}[\boldsymbol{X} \mid \mathcal{G}]\right)
    \\
    &= \mathbb{E}[\|\boldsymbol{X}\|_F^2 \mid \mathcal{G}] - \|\mathbb{E}[\boldsymbol{X} \mid \mathcal{G}]\|_F^2
    \\
    &\le \mathbb{E}[\|\boldsymbol{X}\|_F^2 \mid \mathcal{G}].
\end{align*}
We used $\|\boldsymbol{A}\|_F^2 = \Tr(\boldsymbol{A}^\top\boldsymbol{A})$, $\mathbb{E}[\Tr(\boldsymbol{A}) \mid \mathcal{G}] = \Tr(\mathbb{E}[\boldsymbol{A} \mid \mathcal{G}])$ and that $\mathbb{E}[\boldsymbol{X}\mid\mathcal{G}]$ is $\mathcal{G}$-measurable.
\end{proof}

Finally, we characterize when $\boldsymbol{\Sigma}^{(t)}$ is consistent (in expected Frobenius norm). 

\begin{theorem}[Frobenius norm consistency]
\label{thm:Frobenius norm consistency of Sigma}
Under Assumption~\ref{ass:cond-ind},
\begin{equation*}
    \mathbb{E}\!\left[\|\boldsymbol{\Sigma}^{(t)}-\boldsymbol{\Sigma}\|_F\right] \xrightarrow{t\rightarrow\infty} 0
  \quad\Longleftrightarrow\quad
  \sum_{t=1}^{\infty}(1-\alpha_t)(1-\rho_t)=\infty .
\end{equation*}
\end{theorem}

\begin{proof} Let $P_t \coloneqq \mathbb{E}\bigl[\|\hat{\boldsymbol{S}}_t-\boldsymbol{\Sigma}\|_F^2\bigr]$. 

\medskip\noindent\emph{$P_t$ is finite.} Indeed,
\begin{equation*}
    P_t \le 2\!\left(\mathbb{E}\!\left[\|\hat{\boldsymbol{S}}_t\|_F^2\right] + \|\boldsymbol{\Sigma}\|_F^2\right) \le 2\!\left(\mathbb{E}\!\left[\Tr(\hat{\boldsymbol{S}}_t)^2\right] + \|\boldsymbol{\Sigma}\|_F^2\right) < \infty.
\end{equation*}
We used $\|\boldsymbol{A}-\boldsymbol{B}\|_F^2\le2\!\left(\|\boldsymbol{A}\|_F^2+\|\boldsymbol{B}\|_F^2\right)$
and the monotonicity of the expectation. Then, since $\hat{\boldsymbol{S}}_t$ is symmetric positive-semi-definite (Lemma~\ref{lem:Shat-props}(i)), we have
$\|\hat{\boldsymbol{S}}_t\|_F^2\le\Tr(\hat{\boldsymbol{S}}_t)^2$
(Lemma~\ref{lem:trace-frobenius}(a)). The final bound inequality follows from Lemma~\ref{lem:uniform-L2-Shat}, together with $\|\boldsymbol{\Sigma}\|_F < \infty$.

Steps~1--3 establish sufficiency, i.e.\ that $\sum_{t = 1}^\infty (1 - \alpha_t)(1 - \rho_t) = \infty$ implies $\mathbb{E}[\|\boldsymbol{\Sigma}^{(t)}-\boldsymbol{\Sigma}\|_F] \to 0$. Step~4
establishes necessity.

\medskip\noindent\textbf{Step 1 (A recursion for $P_t$).} For $t \ge 2$, define
\begin{equation}
\label{eq:V tilde, xi tilde def}
    \tilde{\boldsymbol{V}}_t \coloneqq \hat{\boldsymbol{S}}_t - \boldsymbol{\Sigma}, \quad \tilde{\boldsymbol{\xi}}_t \coloneqq \hat{\boldsymbol{S}}_t - \mathbb{E}\!\bigl[\hat{\boldsymbol{S}}_t \mid \mathcal{F}_{t-1}\bigr].
\end{equation}

\medskip\noindent\emph{A recursion for $\tilde{\boldsymbol{V}}_t$.} From Lemma~\ref{lem:Shat-props}(iv), we have
\begin{equation*}
    \tilde{\boldsymbol{V}}_t - \tilde{\boldsymbol{\xi}}_t = \mathbb{E}\!\bigl[\hat{\boldsymbol{S}}_t \mid \mathcal{F}_{t-1}\bigr] - \boldsymbol{\Sigma} = \lambda_{t-1}\!\left(\hat{\boldsymbol{S}}_{t-1} - \boldsymbol{\Sigma}\right) = \lambda_{t-1}\tilde{\boldsymbol{V}}_{t-1}.
\end{equation*}
Thus,
\begin{equation}
    \tilde{\boldsymbol{V}}_t = \lambda_{t-1}\tilde{\boldsymbol{V}}_{t-1} + \tilde{\boldsymbol{\xi}}_t.
\end{equation}

\medskip\noindent\emph{Conditional expectation of $\tilde{\boldsymbol{\xi}}_t$.} From \eqref{eq:V tilde, xi tilde def}, the following identity holds:
\begin{equation}
\label{eq: E(xi | F) = 0}
    \mathbb{E}[\tilde{\boldsymbol{\xi}}_t\mid\mathcal{F}_{t-1}] = \mathbb{E}[\hat{\boldsymbol{S}}_t \mid \mathcal{F}_{t-1}] - \mathbb{E}[\hat{\boldsymbol{S}}_t \mid \mathcal{F}_{t-1}] = \boldsymbol{0}.
\end{equation}

\medskip\noindent\emph{Final computation.} From \eqref{eq:V tilde, xi tilde def}, we have $P_t = \mathbb{E}\bigl[\|\tilde{\boldsymbol{V}}_t\|_F^2\bigr]$. Thus,
\begin{equation}
\label{eq:first rec for P_T}
\begin{aligned}
    P_t &= \mathbb{E}\bigl[\langle \tilde{\boldsymbol{V}}_t, \tilde{\boldsymbol{V}}_t \rangle_F\bigr]
    \\
    &= \mathbb{E}\bigl[\langle \lambda_{t-1}\tilde{\boldsymbol{V}}_{t-1} + \tilde{\boldsymbol{\xi}}_t, \lambda_{t-1}\tilde{\boldsymbol{V}}_{t-1} + \tilde{\boldsymbol{\xi}}_t \rangle_F\bigr]
    \\
    &= \lambda_{t-1}^2 \mathbb{E}\bigl[\langle \tilde{\boldsymbol{V}}_{t-1}, \tilde{\boldsymbol{V}}_{t-1} \rangle_F\bigr] + 2\lambda_{t-1} \mathbb{E}\bigl[\langle \tilde{\boldsymbol{V}}_{t-1}, \tilde{\boldsymbol{\xi}}_t \rangle_F\bigr] + \mathbb{E}\bigl[\langle \tilde{\boldsymbol{\xi}}_t, \tilde{\boldsymbol{\xi}}_t \rangle_F\bigr]
    \\
    &= \lambda_{t-1}^2 \mathbb{E}\bigl[\|\tilde{\boldsymbol{V}}_{t-1}\|_F^2\bigr] + \mathbb{E}\bigl[\|\tilde{\boldsymbol{\xi}}_t\|_F^2\bigr]
    \\
    &= \lambda_{t-1}^2 P_{t-1} + \mathbb{E}\bigl[\|\tilde{\boldsymbol{\xi}}_t\|_F^2\bigr].
\end{aligned}
\end{equation}
We defined the inner product $\langle \boldsymbol{A}, \boldsymbol{B} \rangle_F \coloneqq \Tr(\boldsymbol{A}^\top\boldsymbol{B}) = \sum_{i=1}^d \sum_{j=1}^d A_{ij}B_{ij}$, and used $\mathbb{E}\bigl[\langle \tilde{\boldsymbol{V}}_{t-1}, \tilde{\boldsymbol{\xi}}_t \rangle_F\bigr] = 0$. Indeed, $\tilde{\boldsymbol{V}}_{t-1}$ is $\mathcal{F}_{t-1}$-measurable, and from \eqref{eq: E(xi | F) = 0} we have:
\begin{align*}
    \mathbb{E}\bigl[\langle \tilde{\boldsymbol{V}}_{t-1}, \tilde{\boldsymbol{\xi}}_t \rangle_F\bigr] &= \mathbb{E}\bigl[\mathbb{E}\bigl[\Tr\bigl(\tilde{\boldsymbol{V}}_{t-1}^\top\tilde{\boldsymbol{\xi}}_t\bigr) \mid \mathcal{F}_{t-1}\bigr]\bigr]
    \\
    &= \mathbb{E}\bigl[\Tr\bigl(\mathbb{E}\bigl[\tilde{\boldsymbol{V}}_{t-1}^\top\tilde{\boldsymbol{\xi}}_t \mid \mathcal{F}_{t-1}\bigr]\bigr)\bigr]
    \\
    &= \mathbb{E}\bigl[\Tr\bigl(\tilde{\boldsymbol{V}}_{t-1}^\top\mathbb{E}\bigl[\tilde{\boldsymbol{\xi}}_t \mid \mathcal{F}_{t-1}\bigr]\bigr)\bigr]
    \\
    &= \mathbb{E}\bigl[\Tr\bigl(\tilde{\boldsymbol{V}}_{t-1}^\top\boldsymbol{0}\bigr)\bigr]
    \\
    &= 0.
\end{align*}
Finally, from \eqref{eq:first rec for P_T}, we have
\begin{equation}\label{eq:PT-rec}
  P_t = \lambda_{t-1}^2\,P_{t-1} + \mathbb{E}\bigl[\|\tilde{\boldsymbol{\xi}}_t\|_F^2\bigr].
\end{equation}

\medskip\noindent\emph{Bounding $\mathbb{E}\bigl[\|\tilde{\boldsymbol{\xi}}_t\|_F^2\bigr]$.} For every $s \ge 1$ and $1 \le i \le n_b(s)$, define 
\begin{equation}
\label{eq: z_i,t def}
    \boldsymbol{z}^{(s)}_i \coloneqq \boldsymbol{x}^{(s)}_i - \boldsymbol{\mu}.
\end{equation}
Fix $t \ge 2$. For every $1 \le s \le t - 2$, the random vector $\boldsymbol{z}^{(s)}_i$ is $\mathcal{F}_{t-1}$-measurable, and so is $\boldsymbol{x}^{(0)}_j-\boldsymbol{\mu}$ for $1\le j\le n_r$. Hence their outer products equal their conditional expectations given $\mathcal{F}_{t-1}$, e.g.,
\begin{equation*}
    \mathbb{E}\!\left[\boldsymbol{z}^{(s)}_i\left(\boldsymbol{z}^{(s)}_i\right)^\top \Big| \ \mathcal{F}_{t-1}\right] = \boldsymbol{z}^{(s)}_i\left(\boldsymbol{z}^{(s)}_i\right)^\top.
\end{equation*}
From \eqref{eq:Shat-def} and \eqref{eq:V tilde, xi tilde def}, these terms cancel, and we get
\begin{equation}
\label{eq: xi_T first decomp}
    \tilde{\boldsymbol{\xi}}_t = \frac{1}{N_t}\sum_{i=1}^{n_b(t-1)} \!\left(\boldsymbol{z}^{(t-1)}_i \!\bigl(\boldsymbol{z}^{(t-1)}_i\bigr)^\top - \mathbb{E}\!\left[\boldsymbol{z}^{(t-1)}_i \!\bigl(\boldsymbol{z}^{(t-1)}_i\bigr)^\top \ \Big| \  \mathcal{F}_{t-1}\right]\right).
\end{equation}
Let us define
\begin{equation}
\label{eq: Z_i def}
    \boldsymbol{Z}^{(t-1)}_i \coloneqq \boldsymbol{z}^{(t-1)}_i \!\bigl(\boldsymbol{z}^{(t-1)}_i\bigr)^\top - \mathbb{E}\!\left[\boldsymbol{z}^{(t-1)}_i \!\bigl(\boldsymbol{z}^{(t-1)}_i\bigr)^\top \ \Big| \  \mathcal{F}_{t-1}\right].
\end{equation}
Finally, from \eqref{eq: xi_T first decomp} and \eqref{eq: Z_i def}, we have
\begin{equation}
\label{eq: final exi_T decomp}
    \tilde{\boldsymbol{\xi}}_t = \frac{1}{N_t}\sum_{i=1}^{n_b(t-1)} \boldsymbol{Z}^{(t-1)}_i.
\end{equation}
Conditionally on $\mathcal{F}_{t-1}$, the samples $\boldsymbol{x}^{(t-1)}_i$ are i.i.d. Thus, from \eqref{eq: z_i,t def} and \eqref{eq: Z_i def}, the random matrices $\boldsymbol{Z}^{(t-1)}_i$ are also i.i.d. as a deterministic transformation of $\boldsymbol{x}^{(t-1)}_i$. That is,
\begin{align*}
    \mathbb{E}[\|\tilde{\boldsymbol{\xi}}_t\|_F^2\mid\mathcal{F}_{t-1}]
    &= \frac{n_b(t-1)}{N_t^2}\,\mathbb{E}[\|\boldsymbol{Z}^{(t-1)}_i\|_F^2\mid\mathcal{F}_{t-1}]
    \\
    &\le \beta_{t-1}\,\mathbb{E}[\|\boldsymbol{z}^{(t-1)}_i\|_2^4\mid\mathcal{F}_{t-1}]
    \\
    &= \beta_{t-1}\,\mathbb{E}[\|\boldsymbol{x}^{(t-1)}_i-\boldsymbol{\mu}\|_2^4\mid\mathcal{F}_{t-1}]
    \\
    &\le \beta_{t-1}\bigl(3\alpha_{t-1}\Tr(\hat{\boldsymbol{S}}_{t-1})^2+C_{\boldsymbol{\Sigma}}\bigr).
\end{align*}
We used the definition of $\beta_t$ from~\eqref{eq:scalars} and
Lemma~\ref{lem: useful_upper_bound} with $\mathcal{G}=\mathcal{F}_{t-1}$ and
$\boldsymbol{X} = \boldsymbol{z}^{(t-1)}_i\bigl(\boldsymbol{z}^{(t-1)}_i\bigr)^\top$,
for which
\begin{equation*}
    \|\boldsymbol{X}\|_F^2
    = \left\|\boldsymbol{z}^{(t-1)}_i\bigl(\boldsymbol{z}^{(t-1)}_i\bigr)^\top\right\|_F^2
    = \left\|\boldsymbol{z}^{(t-1)}_i\right\|_2^4 ,
\end{equation*}
so that its hypothesis $\mathbb{E}[\|\boldsymbol{X}\|_F^2]<\infty$ holds: taking
expectations in Lemma~\ref{lem:fourth-moment} (using the tower property) and using
Lemma~\ref{lem:uniform-L2-Shat}, $\mathbb{E}\bigl[\|\boldsymbol{z}^{(t-1)}_i\|_2^4\bigr]
\le 3\alpha_{t-1}\,\mathbb{E}\bigl[\Tr(\hat{\boldsymbol{S}}_{t-1})^2\bigr] + C_{\boldsymbol{\Sigma}} < \infty$. We also used~\eqref{eq: z_i,t def} and Lemma~\ref{lem:fourth-moment} for the conditional fourth moment. This gives
\begin{equation}
\label{eq:final cond exp bound of xi_T}
    \mathbb{E}[\|\tilde{\boldsymbol{\xi}}_t\|_F^2\mid\mathcal{F}_{t-1}] \le \beta_{t-1}\bigl(3\alpha_{t-1}\Tr(\hat{\boldsymbol{S}}_{t-1})^2+C_{\boldsymbol{\Sigma}}\bigr).
\end{equation}
Using the tower property in \eqref{eq:final cond exp bound of xi_T}, we have
\begin{equation}
\label{eq: final E{xi^2} bound}
\begin{aligned}
    \mathbb{E}[\|\tilde{\boldsymbol{\xi}}_t\|_F^2] &= \mathbb{E}\!\left[\mathbb{E}\!\left[\|\tilde{\boldsymbol{\xi}}_t\|_F^2\mid\mathcal{F}_{t-1}\right]\right]
    \\
    &\le \mathbb{E}\!\left[\beta_{t-1}\!\left(3\alpha_{t-1}\Tr(\hat{\boldsymbol{S}}_{t-1})^2 + C_{\boldsymbol{\Sigma}}\right)\right]
    \\
    &= \beta_{t-1}\!\left(3\alpha_{t-1}\mathbb{E}\!\left[\Tr(\hat{\boldsymbol{S}}_{t-1})^2\right] + C_{\boldsymbol{\Sigma}}\right)
    \\
    &\le \left(3B^\ast+C_{\boldsymbol{\Sigma}}\right)\beta_{t-1}.
\end{aligned}
\end{equation}
We used $\alpha_{t-1} \le 1$ and the uniform bound $\mathbb{E}\!\bigl[\Tr(\hat{\boldsymbol{S}}_{t-1})^2\bigr] \le B^\ast$ from Lemma~\ref{lem:uniform-L2-Shat}. Finally, from \eqref{eq: final E{xi^2} bound} we have:
\begin{equation}
\label{eq:xi-tilde-bound}
  \mathbb{E}[\|\tilde{\boldsymbol{\xi}}_t\|_F^2]\le D\,\beta_{t-1},\qquad
  D \coloneqq 3B^\ast+C_{\boldsymbol{\Sigma}}<\infty.
\end{equation}

\medskip\noindent\textbf{Step 2 ($P_t\to0$).}
From \eqref{eq:PT-rec} and \eqref{eq:xi-tilde-bound}, we have $P_t\le\lambda_{t-1}^2P_{t-1}+D\beta_{t-1}$.
Unrolling by Lemma~\ref{lem:rec-unroll-frob} with $r_{t-1}=\lambda_{t-1}^2\le1$
(Lemma \ref{lem:range}) and $q_{t-1}=D\beta_{t-1}$,
\begin{equation}
\label{eq:PT-unroll}
  P_t \le \underbrace{\left(\prod_{k=1}^{t-1}\lambda_k^2\right)}_{(\mathrm{I})}P_1
   +D\underbrace{\sum_{s=1}^{t-1}\left(\prod_{k=s+1}^{t-1}\lambda_k^2\right)\beta_s}_{\mathrm{(II)}}.
\end{equation}

\emph{Term $(\mathrm{I})$.} By the hypothesis $\sum_{t = 1}^\infty (1 - \alpha_t)(1 - \rho_t) = \infty$ and Lemma~\ref{lem:lambda2-div},
\begin{equation}
\label{eq:prod of lambda_s^2 -> 0}
    \prod_{k=1}^{t-1}\lambda_k^2
    =\exp\!\left(\sum_{k=1}^{t-1}\ln\lambda_k^2\right)
    \le\exp\!\left(-\sum_{k=1}^{t-1}(1-\lambda_k^2)\right)\xrightarrow[t\to\infty]{}0.
\end{equation}
We used $\ln(x) \le x-1$ with $x = \lambda_k^2 \ge 0$ (indeed, $ 0 \le \lambda_k \le 1$ from Lemma~\ref{lem:range})

\emph{Term $(\mathrm{II})$.} Fix $\varepsilon > 0$. By Lemma~\ref{lem:beta-summability}, choose $s_0 \ge 2$ with
\begin{equation}
\label{eq: beta_t tail sum}
    \sum_{s = s_0}^\infty \beta_s < \frac{\varepsilon}{2},
\end{equation}
and from \eqref{eq:prod of lambda_s^2 -> 0}, choose $s_1 \ge 1$ such that for every $t \ge s_1$,
\begin{equation}
\label{eq: lambda_s^2 tail prod}
    \prod_{k = s_0}^{t - 1} \lambda_k^2 \le \frac{1}{\sum_{s = 1}^{s_0 - 1} \beta_s}\frac{\varepsilon}{2},
\end{equation}
where we used $\prod_{k = s_0}^{t - 1} \lambda_k^2 = \frac{\prod_{k = 1}^{t - 1} \lambda_k^2}{\prod_{k = 1}^{s_0 - 1} \lambda_k^2} \xrightarrow[t \rightarrow \infty]{} 0$.

From \eqref{eq:PT-unroll}, \eqref{eq: beta_t tail sum} and \eqref{eq: lambda_s^2 tail prod}, for all $t \ge \max(s_0 + 2, s_1)$:
\begin{equation}
\label{eq: final bound for (II)}
\begin{aligned}
    \mathrm{II} &= \sum_{s=1}^{s_0 - 1} \left(\prod_{k = s+1}^{t-1} \lambda_k^2\right)\beta_s + \sum_{s = s_0}^{t-1} \left(\prod_{k = s+1}^{t-1} 
    \lambda_k^2\right)\beta_s
    \\
    &\le \prod_{k = s_0}^{t - 1}\lambda_k^2 \cdot \sum_{s=1}^{s_0 - 1}\beta_s + \sum_{s=s_0}^{t-1} \beta_s
    \\
    &\le \prod_{k = s_0}^{t - 1}\lambda_k^2 \cdot \sum_{s=1}^{s_0 - 1}\beta_s + \sum_{s = s_0}^\infty \beta_s
    \\
    &\le \frac{\varepsilon}{2} + \frac{\varepsilon}{2}
    \\
    &= \varepsilon,
\end{aligned}
\end{equation}
where we used $\lambda_k^2 \le 1$ for all $k \ge 1$ (Lemma~\ref{lem:range}). Finally, \eqref{eq: final bound for (II)} holds for every $\varepsilon > 0$, and thus $(\mathrm{II}) \to 0$. Together with $(\mathrm{I}) \to 0$ from \eqref{eq:prod of lambda_s^2 -> 0}, we have $P_t \to 0$ (indeed, from \eqref{eq:PT-unroll}, $P_t \le (\mathrm{I})\cdot P_1 + D\cdot (\mathrm{II}) \to 0$). That is,
\begin{equation}
\label{eq:PT->0 final}
    P_t \xrightarrow{t \rightarrow \infty} 0.
\end{equation}

\medskip\noindent\textbf{Step 3 (from $\hat{\boldsymbol{S}}_t$ to $\boldsymbol{\Sigma}^{(t)}$).} The triangle inequality gives
\begin{equation}
\label{eq: final upper bound for E[||Sigma(T) - Sigma||]}
\begin{aligned}
    \mathbb{E}\!\bigl[\|\boldsymbol{\Sigma}^{(t)} - \boldsymbol{\Sigma}\|_F\bigr] &\le \mathbb{E}\!\bigl[\|\hat{\boldsymbol{S}}_t - \boldsymbol{\Sigma}\|_F\bigr] + \mathbb{E}\!\bigl[\|\boldsymbol{\Sigma}^{(t)} - \hat{\boldsymbol{S}}_t\|_F\bigr]
    \\
    &= \mathbb{E}\!\bigl[\|\hat{\boldsymbol{S}}_t - \boldsymbol{\Sigma}\|_F\bigr] + \mathbb{E}\!\bigl[\|\boldsymbol{\delta}^{(t)}(\boldsymbol{\delta}^{(t)})^\top\|_F\bigr]  
    \\
    &\le \sqrt{\mathbb{E}\!\bigl[\|\hat{\boldsymbol{S}}_t - \boldsymbol{\Sigma}\|_F^2\bigr]} + \mathbb{E}\!\bigl[\|\boldsymbol{\delta}^{(t)}\|_2^2\bigr]
    \\
    &= \sqrt{P_t} + \Tr(\boldsymbol{C}_t).
\end{aligned}
\end{equation}
We used Lemma~\ref{lem:Shat-props}(ii) and the identity $\|\boldsymbol{x}\boldsymbol{x}^\top\|_F = \|\boldsymbol{x}\|_2^2$. Additionally, $(\cdot)^2$ is convex, and Jensen's inequality with
$X = \|\hat{\boldsymbol{S}}_t - \boldsymbol{\Sigma}\|_F$ gives
\begin{equation*}
    \mathbb{E}[X]^2 \leq \mathbb{E}[X^2]
    \quad\Rightarrow\quad
    \mathbb{E}[X] \leq \sqrt{\mathbb{E}[X^2]},
\end{equation*}
where $X \geq 0$ a.s. Then, we used $P_t = \mathbb{E}\!\bigl[\|\hat{\boldsymbol{S}}_t - \boldsymbol{\Sigma}\|_F^2\bigr]$ and $\mathbb{E}\!\bigl[\|\boldsymbol{\delta}^{(t)}\|_2^2\bigr] = \Tr(\boldsymbol{C}_t)$, which follows from \eqref{eq:MSE-trace}. That is, from \eqref{eq: final upper bound for E[||Sigma(T) - Sigma||]}, we have shown that
\begin{equation}
\label{eq:L1-final-decomp}
    \mathbb{E}\!\bigl[\|\boldsymbol{\Sigma}^{(t)} - \boldsymbol{\Sigma}\|_F\bigr] \le \sqrt{P_t} + \Tr(\boldsymbol{C}_t).
\end{equation}
By~\eqref{eq:PT->0 final}, we have $P_t \to 0$. Additionally, from our hypothesis $\sum_{t=1}^\infty (1 - \alpha_t)(1 - \rho_t) = \infty$ and Theorem~\ref{thm:sharp-criterion-mu}, we have $\boldsymbol{C}_t \to \boldsymbol{0}$. Hence, $\Tr(\boldsymbol{C}_t) \to 0$ (Lemma~\ref{lem: CT->0 <=> mu(t)-> mu in L2}). Thus, both terms
of~\eqref{eq:L1-final-decomp} vanish, so $\mathbb{E}[\|\boldsymbol{\Sigma}^{(t)}-\boldsymbol{\Sigma}\|_F] \to 0$.

\medskip\noindent\textbf{Step 4 (necessity).}
Conversely, suppose
\begin{equation}
\label{eq:hypothesis_sigma_thm}
    \sum_{t=1}^{\infty}(1-\alpha_t)(1-\rho_t) < \infty.
\end{equation}
Since $\|\cdot\|_F$ is convex, Jensen's inequality and Theorem~\ref{thm:estimators_bias}(b) give
\begin{equation}
\label{eq: L1-decomp-lower-bound}
    \mathbb{E}\bigl[\|\boldsymbol{\Sigma}^{(t)}-\boldsymbol{\Sigma}\|_F\bigr]
    \ge \bigl\|\mathbb{E}[\boldsymbol{\Sigma}^{(t)}]-\boldsymbol{\Sigma}\bigr\|_F
    = \|{-\boldsymbol{C}_t}\|_F = \|\boldsymbol{C}_t\|_F,
\end{equation}
where we used Theorem~\ref{thm:estimators_bias}(b). By \eqref{eq:hypothesis_sigma_thm} and Theorem \ref{thm:sharp-criterion-mu}, $\boldsymbol{C}_t \not\to \boldsymbol{0}$. Hence, $\|\boldsymbol{C}_t\|_F\not\to0$
(Remark after Lemma~\ref{lem: CT->0 <=> mu(t)-> mu in L2}). From \eqref{eq: L1-decomp-lower-bound}, we have $\mathbb{E}[\|\boldsymbol{\Sigma}^{(t)}-\boldsymbol{\Sigma}\|_F]\not\to 0$. This establishes the stated equivalence.
\end{proof}

\subsubsection{Convergence of the fitted distribution to the true distribution}
\label{sec:W2->0 cond}
We now characterize when the fitted distribution $\mathcal{D}_t$ converges to the true distribution $\mathcal{D}_{\mathrm{real}}$ (see~\eqref{eq:model-def}) in the $2$-Wasserstein distance $W_2$, a standard metric between probability distributions \citep{villani2009optimal}.

\begin{definition}[$2$-Wasserstein distance]\label{def:W2-def}
For probability distributions $P,Q$ on $\mathbb{R}^d$ with finite second moments,
\begin{equation*}
    W_2^2(P,Q) \coloneqq \inf_{\pi\in\Pi(P,Q)}\ \mathbb{E}_{(\boldsymbol{x},\boldsymbol{y})\sim\pi}\bigl[\|\boldsymbol{x}-\boldsymbol{y}\|_2^2\bigr],
\end{equation*}
where $\Pi(P,Q)$ is the set of couplings of $P$ and $Q$, i.e., joint distributions with marginals $P$ and $Q$.
\end{definition}

First, we prove a bound on a metric between two symmetric positive-semi-definite matrices. It is used in the proof of Theorem~\ref{thm:W2->0}.

\begin{lemma}[Bures--Frobenius bound]
\label{lem:bures-frobenius}
Let $\boldsymbol A,\boldsymbol B\in\mathbb{R}^{d\times d}$ be symmetric with
$\boldsymbol A\succeq0$ and $\boldsymbol B\succeq0$. Define the Bures metric $\mathcal{B}$ as
\begin{equation}
\label{eq:bures metric def}
    \mathcal{B}^2(\boldsymbol A,\boldsymbol B)
    =\Tr(\boldsymbol A)+\Tr(\boldsymbol B)
     -2\Tr\bigl((\boldsymbol A^{1/2}\boldsymbol B\boldsymbol A^{1/2})^{1/2}\bigr).
\end{equation}
Then
\begin{equation}
\label{eq:bures-frobenius}
  \mathcal{B}^2(\boldsymbol A,\boldsymbol B)
  \;\le\;\bigl\|\boldsymbol A^{1/2}-\boldsymbol B^{1/2}\bigr\|_F^2
  \;\le\;\|\boldsymbol A-\boldsymbol B\|_*
  \;\le\;\sqrt{d}\,\|\boldsymbol A-\boldsymbol B\|_F ,
\end{equation}
where $\|\cdot\|_*$ denotes the nuclear (trace) norm, i.e. the sum of the singular values.
\end{lemma}

\begin{proof}
\emph{First inequality.} Expanding the Frobenius norm, we get
\begin{equation}
\label{eq: ||A^(1/2) - B^(1/2)||_F^2 identity Lemma}
\begin{aligned}
    \bigl\|\boldsymbol A^{1/2}-\boldsymbol B^{1/2}\bigr\|_F^2 &= \Tr\bigl((\boldsymbol A^{1/2}-\boldsymbol B^{1/2})(\boldsymbol A^{1/2}-\boldsymbol B^{1/2})^\top\bigr)
    \\
    &= \Tr\bigl(\boldsymbol{A} - \boldsymbol{A}^{1/2}\boldsymbol{B}^{1/2} - \boldsymbol{B}^{1/2}\boldsymbol{A}^{1/2} + \boldsymbol{B}\bigr)
    \\
    &= \Tr(\boldsymbol{A}) + \Tr(\boldsymbol{B}) - 2\Tr(\boldsymbol{A}^{1/2}\boldsymbol{B}^{1/2})
\end{aligned}
\end{equation}
Indeed, for any $\boldsymbol{C}\succeq 0$, its symmetric square root $\boldsymbol{C}^{1/2}$ satisfies $\boldsymbol{C}^{1/2}\boldsymbol{C}^{1/2}=\boldsymbol{C}$. Finally, we have $\Tr(\boldsymbol{A}\boldsymbol{B}) = \Tr(\boldsymbol{B}\boldsymbol{A})$. Using the definition of $\mathcal{B}^2$ in \eqref{eq:bures metric def} and \eqref{eq: ||A^(1/2) - B^(1/2)||_F^2 identity Lemma}, the claim is equivalent to
\begin{equation}
\label{eq: ineq to prove Tr(A^1/2B^1/2) <= Tr((A^1/2 B A^1/2)^1/2)}
    \Tr(\boldsymbol{A}^{1/2}\boldsymbol {B}^{1/2})
    \le\Tr\bigl((\boldsymbol{A}^{1/2}\boldsymbol{B} \boldsymbol{A}^{1/2})^{1/2}\bigr).
\end{equation}
Set $\boldsymbol{M} \coloneqq \boldsymbol{B}^{1/2}\boldsymbol{A}^{1/2}$. Then
\begin{equation}
\label{eq: M^TM = A^1/2BA^1/2}
    \boldsymbol{M}^\top\boldsymbol{M} = \boldsymbol{A}^{1/2}\boldsymbol{B}^{1/2}\boldsymbol{B}^{1/2}\boldsymbol{A}^{1/2} = \boldsymbol{A}^{1/2}\boldsymbol{B}\boldsymbol{A}^{1/2}.
\end{equation}
Since the singular values of $\boldsymbol{M}$ are the eigenvalues of the positive-semi-definite matrix $(\boldsymbol{M}^\top\boldsymbol{M})^{1/2}$, the nuclear norm satisfies, by~\eqref{eq: M^TM = A^1/2BA^1/2},
\begin{equation}
\label{eq: first eq to ||M||_*}
    \|\boldsymbol{M}\|_*=\Tr\bigl((\boldsymbol{M}^\top\boldsymbol{M})^{1/2}\bigr) =\Tr\bigl((\boldsymbol{A}^{1/2}\boldsymbol{B}\boldsymbol{A}^{1/2})^{1/2}\bigr).
\end{equation}
By cyclicity,
\begin{equation}
\label{eq: Tr(M)=Tr(A^1/2B^1/2)}
    \Tr(\boldsymbol{M}) = \Tr(\boldsymbol{A}^{1/2}\boldsymbol{B}^{1/2}).
\end{equation}
Using \eqref{eq: ineq to prove Tr(A^1/2B^1/2) <= Tr((A^1/2 B A^1/2)^1/2)}, \eqref{eq: first eq to ||M||_*}, \eqref{eq: Tr(M)=Tr(A^1/2B^1/2)}, the claim is $\Tr(\boldsymbol{M})\le\|\boldsymbol{M}\|_*$, {\color{black} which follows immediately from von Neumann's trace inequality \citep{horn2012matrix}.}

\smallskip\noindent\emph{Second inequality.} This is the Powers--St\o rmer inequality \citep{powers1970free}; see also \citet[Lemma~1]{santoro2025large}.

\smallskip\noindent\emph{Third inequality.} Let $s_1,\dots,s_d\ge 0$ be the singular values of $\boldsymbol{A}-\boldsymbol{B}$, so that $\|\boldsymbol{A}-\boldsymbol{B}\|_* = \sum_{i=1}^d s_i$ and $\|\boldsymbol{A}-\boldsymbol{B}\|_F = \left(\sum_{i=1}^d s_i^2\right)^{1/2}$. Lemma~\ref{lem:l1-l2} gives
\begin{equation*}
    \|\boldsymbol{A}-\boldsymbol{B}\|_* = \sum_{i=1}^d s_i
    \le\sqrt{d}\,\left(\sum_{i=1}^d s_i^2\right)^{1/2}
    =\sqrt{d}\,\|\boldsymbol{A}-\boldsymbol{B}\|_F .
\end{equation*}
\end{proof}

{\color{black}
We now characterize when the fitted distribution converges to the true distribution in the $2$-Wasserstein distance.

\begin{theorem}[$W_2$ consistency]
\label{thm:W2->0}
Under Assumption~\ref{ass:cond-ind},
\begin{equation*}
    \mathbb{E}\bigl[W_2^2(\mathcal{D}_t,\mathcal{D}_{\mathrm{real}})\bigr]
    \xrightarrow{t\rightarrow\infty} 0
    \quad\Longleftrightarrow\quad
    \sum_{t=1}^{\infty}(1-\alpha_t)(1-\rho_t)=\infty .
\end{equation*}
\end{theorem}

\begin{proof}
Recall Definition~\ref{def:W2-def}. For Gaussians, it admits the closed form \citep{dowson1982frechet, gelbrich1990}
\begin{equation}
\label{eq: gaussian W2 formula}
  W_2^2\bigl(\mathcal N(\boldsymbol\mu_1,\boldsymbol\Sigma_1),
    \mathcal N(\boldsymbol\mu_2,\boldsymbol\Sigma_2)\bigr)
  = \|\boldsymbol\mu_1-\boldsymbol\mu_2\|_2^2
    + \mathcal{B}^2(\boldsymbol\Sigma_1,\boldsymbol\Sigma_2),
\end{equation}
where $\mathcal{B}$ is the Bures metric~\eqref{eq:bures metric def}.
Applying~\eqref{eq: gaussian W2 formula} with
$\boldsymbol\mu_1=\boldsymbol\mu^{(t)}$,
$\boldsymbol\Sigma_1=\boldsymbol\Sigma^{(t)}$,
$\boldsymbol\mu_2=\boldsymbol\mu$, $\boldsymbol\Sigma_2=\boldsymbol\Sigma$, and
writing $\mathcal{B}_t\coloneqq\mathcal{B}(\boldsymbol\Sigma^{(t)},\boldsymbol\Sigma)$,
we obtain from~\eqref{eq: gaussian W2 formula} the pointwise identity
\begin{equation}
\label{eq:W2-split}
  W_2^2(\mathcal{D}_t,\mathcal{D}_{\mathrm{real}})
  = \|\boldsymbol{\mu}^{(t)} - \boldsymbol{\mu}\|_2^2 + \mathcal{B}_t^2 ,
\end{equation}
in which both summands are nonnegative. Taking expectations
in~\eqref{eq:W2-split} and using
$\mathbb{E}\bigl[\|\boldsymbol\delta^{(t)}\|_2^2\bigr]=\Tr(\boldsymbol C_t)$
from~\eqref{eq:MSE-trace},
\begin{equation}
\label{eq:EW2sq-identity}
  \mathbb{E}\bigl[W_2^2(\mathcal{D}_t,\mathcal{D}_{\mathrm{real}})\bigr]
  = \mathbb{E}\bigl[\|\boldsymbol{\mu}^{(t)} - \boldsymbol{\mu}\|_2^2\bigr] + \mathbb{E}\bigl[\mathcal{B}_t^2\bigr].
\end{equation}

\medskip\noindent\textit{Sufficiency.}
Suppose $\sum_{t=1}^{\infty}(1-\alpha_t)(1-\rho_t)=\infty$. By
Lemma~\ref{lem:sample covariance is PSD} we have
$\boldsymbol\Sigma^{(t)}\succeq 0$, and $\boldsymbol\Sigma\succ 0$ by assumption,
so Lemma~\ref{lem:bures-frobenius} applies and gives the pointwise bound
\begin{equation*}
    \mathcal{B}_t^2\;\le\;\sqrt d\,\bigl\|\boldsymbol\Sigma^{(t)}-\boldsymbol\Sigma\bigr\|_F .
\end{equation*}
Substituting into~\eqref{eq:EW2sq-identity},
\begin{equation}
\label{eq:EW2sq-bound}
  \mathbb{E}\bigl[W_2^2(\mathcal{D}_t,\mathcal{D}_{\mathrm{real}})\bigr] \le \mathbb{E}\bigl[\|\boldsymbol{\mu}^{(t)} - \boldsymbol{\mu}\|_2^2\bigr] + \sqrt d\;\mathbb{E}\bigl[\|\boldsymbol\Sigma^{(t)}-\boldsymbol\Sigma\|_F\bigr].
\end{equation}
By Theorem~\ref{thm:sharp-criterion-mu}, the hypothesis gives
$\mathbb{E}\bigl[\|\boldsymbol{\mu}^{(t)} - \boldsymbol{\mu}\|_2^2\bigr] \to 0$. By Theorem~\ref{thm:Frobenius norm consistency of Sigma}, the hypothesis also gives
$\mathbb{E}\bigl[\|\boldsymbol\Sigma^{(t)}-\boldsymbol\Sigma\|_F\bigr]\to0$.
Both terms on the right-hand side of~\eqref{eq:EW2sq-bound} therefore vanish, so
$\mathbb{E}\bigl[W_2^2(\mathcal{D}_t,\mathcal{D}_{\mathrm{real}})\bigr]\to0$.

\medskip\noindent\textit{Necessity.}
Conversely, suppose
$\mathbb{E}\bigl[W_2^2(\mathcal{D}_t,\mathcal{D}_{\mathrm{real}})\bigr]\to0$. Since
$\mathbb{E}\bigl[\mathcal{B}_t^2\bigr]\ge0$, identity~\eqref{eq:EW2sq-identity} gives
\begin{equation*}
    0 \le \mathbb{E}\bigl[\|\boldsymbol{\mu}^{(t)} - \boldsymbol{\mu}\|_2^2\bigr] \le \mathbb{E}\bigl[W_2^2(\mathcal{D}_t,\mathcal{D}_{\mathrm{real}})\bigr]
    \xrightarrow{t\to\infty} 0 .
\end{equation*}
Thus, $\mathbb{E}\bigl[\|\boldsymbol{\mu}^{(t)} - \boldsymbol{\mu}\|_2^2\bigr] \to 0$. Using Theorem~\ref{thm:sharp-criterion-mu} and Lemma~\ref{lem: CT->0 <=> mu(t)-> mu in L2}, we have  $\sum_{t=1}^{\infty}(1-\alpha_t)(1-\rho_t)=\infty$. This establishes the stated equivalence.
\end{proof}

\subsection{Proof of Theorem~\ref{thm:unified_sharp_criterion}}
\label{sec:unified_sharp_criterion_proof}
Theorem~\ref{thm:unified_sharp_criterion} characterizes in closed form when this iterative estimation process converges to the true distribution. We now assemble
it from Lemma~\ref{lem: CT->0 <=> mu(t)-> mu in L2} and Theorems~\ref{thm:sharp-criterion-mu}, \ref{thm:Frobenius norm consistency of Sigma},
and~\ref{thm:W2->0}.

\begin{proof}
It suffices to show that (a) is equivalent to each of (b), (c), and (d). First, (a)$\Leftrightarrow$(b) follows by combining
Theorem~\ref{thm:sharp-criterion-mu} with Lemma~\ref{lem: CT->0 <=> mu(t)-> mu in L2}. Second, (a)$\Leftrightarrow$(c) is exactly Theorem~\ref{thm:Frobenius norm consistency of Sigma}. Finally, (a)$\Leftrightarrow$(d) is precisely Theorem~\ref{thm:W2->0}.
\end{proof}

\subsection{Proof of Lemma~\ref{lem: sup_t alpha_t < 1 => cond holds}}
\label{app:sup_t alpha_t < 1 => cond holds thm}

Before proving Lemma~\ref{lem: sup_t alpha_t < 1 => cond holds}, we prove an important technical lemma.

\begin{lemma}[Diverging series]\label{lem:diverging_series} Let $0 < a_s \rightarrow \infty$ be strictly increasing; then
\begin{equation*}
    \sum_{s=1}^\infty \left(1 - \frac{a_s}{a_{s+1}}\right) = \infty
\end{equation*}
\end{lemma}

\begin{proof}
Take $\{a_s\}_{s=1}^\infty$ strictly increasing such that $0 < a_s \rightarrow \infty$. We now use Lemma~\ref{lem:product-decay} with $r_s = \frac{a_s}{a_{s+1}} \in (0,1)$ (where we used $a_s > 0$ and $a_s < a_{s+1}$). The condition of the lemma holds:

\begin{equation*}
    \prod_{s=1}^t r_s = \prod_{s=1}^t \frac{a_s}{a_{s+1}} = \frac{a_1}{a_{t+1}} \xrightarrow[t\rightarrow\infty]{} 0,
\end{equation*}
where we used $a_t \to \infty$. This proves 
\begin{equation*}
    \sum_{s=1}^\infty \left(1 - \frac{a_s}{a_{s+1}}\right) = \sum_{s=1}^\infty \left(1 - r_s\right) = \infty
\end{equation*}
\end{proof}

\begin{proof}[Proof of Lemma~\ref{lem: sup_t alpha_t < 1 => cond holds}]

Let $\bar\alpha \coloneqq \sup_{t \ge 1}\alpha_t \in [0, 1)$. We prove that $\sum_{t = 1}^\infty (1 - \alpha_t)(1 - \rho_t) = \infty$. Indeed, note that
\begin{equation*}
    \sum_{t=1}^\infty (1 - \alpha_t)(1-\rho_t) \ge (1 - \bar\alpha)\sum_{t=1}^\infty (1 - \rho_t) = (1 - \bar\alpha)\sum_{t=1}^\infty \left(1 - \frac{N_t}{N_{t+1}}\right) = \infty.
\end{equation*}
We used $1 - \bar\alpha > 0$ and
\begin{equation*}
    \sum_{t=1}^\infty \left(1 - \frac{N_t}{N_{t+1}}\right) = \infty.
\end{equation*}
This directly follows from Lemma~\ref{lem:diverging_series} with $a_t = N_t$. Indeed, $N_t = N_{t-1} + n_b(t - 1) > N_{t-1}$ is strictly increasing, and $N_t = n_r + \sum_{s=1}^{t-1} n_b(s) \ge t - 1 \to \infty \Rightarrow N_t \to \infty$.
\end{proof}

\subsection{Proof of Proposition~\ref{prop: alpha = 1 perm bias}}
\label{app:alpha = 1 perm bias thm}

\begin{proof}
Assume $\alpha_t\equiv1$ throughout. We start by bounding $\mathbb{E}\bigl[\|\boldsymbol{\mu}^{(t)} - \boldsymbol{\mu}\|_2^2\bigr]$.

\medskip\noindent\textbf{Step 1 (lower bound).} By Theorem~\ref{thm:CT}(d), the recursion becomes
\begin{equation}
\label{eq:curr CT rec prop}
    \boldsymbol{C}_t=(1-\beta_{t-1})\boldsymbol{C}_{t-1}+\beta_{t-1}\boldsymbol{\Sigma},
\end{equation}
and hence
\begin{equation*}
    \boldsymbol{C}_t-\boldsymbol{C}_{t-1}
    =\beta_{t-1}\bigl(\boldsymbol{\Sigma}-\boldsymbol{C}_{t-1}\bigr),
    \qquad t \ge 2.
\end{equation*}
By Theorem~\ref{thm:estimators_bias}(b) and Lemma~\ref{lem:sample covariance is PSD}, $\boldsymbol{C}_{t-1}=\boldsymbol{\Sigma}-\mathbb{E}\bigl[\boldsymbol{\Sigma}^{(t-1)}\bigr]\preceq\boldsymbol{\Sigma}$, and
$\beta_{t-1}>0$ by~\eqref{eq:scalars}. Hence, $\boldsymbol{C}_t-\boldsymbol{C}_{t-1} \succeq 0$, i.e., $\boldsymbol{C}_t \succeq \boldsymbol{C}_{t-1}$ for every $t \ge 2$. By induction, 
\begin{equation}
\label{eq: C_T>C_1 prop}   \boldsymbol{C}_t\succeq\boldsymbol{C}_1
\end{equation}
for all $t \ge 1$. Since the trace is monotone with respect to the positive semidefinite order,
\begin{equation*}
    \Tr(\boldsymbol{C}_t) \ge \Tr(\boldsymbol{C}_1) = \frac{\Tr(\boldsymbol{\Sigma})}{n_r},
\end{equation*}
where we used \eqref{eq: C_1 proof Thm 3.7}. Therefore, by \eqref{eq:MSE-trace},
\begin{equation}
\label{eq:MSE lower bound prop}
    \frac{\Tr(\boldsymbol{\Sigma})}{n_r} \le \mathbb{E}\bigl[\|\boldsymbol{\mu}^{(t)} - \boldsymbol{\mu}\|_2^2\bigr].
\end{equation}

\medskip\noindent\textbf{Step 2 (upper bound).} Applying trace in \eqref{eq:curr CT rec prop}, we get
\begin{align*}
    \Tr(\boldsymbol{C}_t) &= (1 - \beta_{t-1})\Tr(\boldsymbol{C}_{t-1}) + \beta_{t-1}\Tr(\boldsymbol{\Sigma})
    \\
    &\le \Tr(\boldsymbol{C}_{t-1}) + \beta_{t -1}\Tr(\boldsymbol{\Sigma}),
\end{align*}
where we used $\beta_{t-1} > 0$ by \eqref{eq:scalars}. By induction,
\begin{equation}
\label{eq:Tr(CT) upper bound prop}
\begin{aligned}
    \Tr(\boldsymbol{C}_t) &\le \Tr(\boldsymbol{C}_1) + \left(\sum_{s = 2}^{t} \beta_{s-1}\right) \Tr(\boldsymbol{\Sigma})
    \\
    &\le \frac{\Tr(\boldsymbol{\Sigma})}{n_r} + \left(\sum_{s = 1}^\infty \beta_s\right)\Tr(\boldsymbol{\Sigma})
    \\
    &\le \frac{2\Tr(\boldsymbol{\Sigma})}{n_r},
\end{aligned}
\end{equation}
where we used \eqref{eq: C_1 proof Thm 3.7}, $\beta_t > 0$ from \eqref{eq:scalars} and Lemma~\ref{lem:beta-summability}. Therefore, by \eqref{eq:MSE-trace}, 
\begin{equation}
\label{eq:MSE upper bound prop}
    \mathbb{E}\bigl[\|\boldsymbol{\mu}^{(t)} - \boldsymbol{\mu}\|_2^2\bigr] \le \frac{2\Tr(\boldsymbol{\Sigma})}{n_r}.
\end{equation}
Combining \eqref{eq:MSE lower bound prop} and \eqref{eq:MSE upper bound prop}, we get
\begin{equation}
\label{eq:MSE final bounds prop}
    \frac{\Tr(\boldsymbol{\Sigma})}{n_r} \le \mathbb{E}\bigl[\|\boldsymbol{\mu}^{(t)} - \boldsymbol{\mu}\|_2^2\bigr] \le \frac{2\Tr(\boldsymbol{\Sigma})}{n_r}.
\end{equation}

Finally, we bound $\mathbb{E}\bigl[\|\boldsymbol{\Sigma}^{(t)}-\boldsymbol{\Sigma}\|_F\bigr]$.

\medskip\noindent\textbf{Step 3 (lower bound).} Using \eqref{eq: L1-decomp-lower-bound} and \eqref{eq: C_T>C_1 prop}
\begin{equation}
\label{eq:Cov lower bound prop}
\begin{aligned}
    \mathbb{E}\bigl[\|\boldsymbol{\Sigma}^{(t)}-\boldsymbol{\Sigma}\|_F\bigr] &\ge \|\boldsymbol{C}_t\|_F
    \\
    &\ge \|\boldsymbol{C}_1\|_F
    \\
    &\ge \frac{\|\boldsymbol{\Sigma}\|_F}{n_r}
\end{aligned}
\end{equation}
where we used the monotonicity of the Frobenius norm with respect to the positive semidefinite order and \eqref{eq: C_1 proof Thm 3.7}.

\medskip\noindent\textbf{Step 4 (upper bound).} Note that \eqref{eq:PT-rec}, \eqref{eq:xi-tilde-bound}, \eqref{eq:PT-unroll} and \eqref{eq: final upper bound for E[||Sigma(T) - Sigma||]} in the proof of Theorem~\ref{thm:Frobenius norm consistency of Sigma} do not use the hypothesis $\sum_{s=1}^\infty(1-\alpha_s)(1-\rho_s)=\infty$. Hence, they hold for $\alpha_t \equiv 1$. Combining \eqref{eq: final upper bound for E[||Sigma(T) - Sigma||]} and \eqref{eq:Tr(CT) upper bound prop}, we get
\begin{equation}
\label{eq:Cov first upper bound prop}
\begin{aligned}
    \mathbb{E}\bigl[\|\boldsymbol{\Sigma}^{(t)}-\boldsymbol{\Sigma}\|_F\bigr] &\le \sqrt{P_t} + \Tr(\boldsymbol{C}_t)
    \\
    &\le \sqrt{P_t} + \frac{2\Tr(\boldsymbol{\Sigma})}{n_r}.
\end{aligned}
\end{equation}
Using \eqref{eq:scalars} and $\alpha_t \equiv 1$, we have for every $t \ge 1$,
\begin{equation*}
    \lambda_t = \frac{N_t + n_b(t)\alpha_t}{N_{t + 1}} = \frac{N_t + n_b(t)}{N_{t + 1}} = \frac{N_{t + 1}}{N_{t + 1}} = 1.
\end{equation*}
Thus, $\lambda_t \equiv 1$. Thus, \eqref{eq:PT-unroll} reads
\begin{equation}
\label{eq: P_T first bound prop}
\begin{aligned}
    P_t &\le \left(\prod_{k = 1}^{t - 1} \lambda_k^2\right)P_1 + D\sum_{s = 1}^{t - 1}\left(\prod_{k = s + 1}^{t - 1} \lambda_k^2\right)\beta_s 
    \\
    &= P_1 + D\sum_{s = 1}^{t - 1} \beta_s
    \\
    &\le P_1 + D\sum_{s = 1}^\infty \beta_s
    \\
    &\le P_1 + \frac{D}{n_r},
\end{aligned}
\end{equation}
where we used $\beta_t > 0$ from \eqref{eq:scalars} and Lemma~\ref{lem:beta-summability}. It is left to bound $P_1$ and $D$. 

\medskip\noindent\textit{Bounding $D$.} From \eqref{eq:xi-tilde-bound}, we have
\begin{equation}
\label{eq: D-def proof prop}
    D = 3B^\ast + C_{\boldsymbol{\Sigma}},
\end{equation}
where $C_{\boldsymbol{\Sigma}}$ is defined in \eqref{eq:CSigma-def}, and $B^\ast = \sup_{t \ge 1} \mathbb{E}\!\bigl[\Tr(\hat{\boldsymbol{S}}_t)^2\bigr]$ is finite from Lemma~\ref{lem:uniform-L2-Shat}. 

We now bound $B^\ast$. Using \eqref{eq:final sup Tr(ST^2) bound proof}, we have
\begin{equation}
\label{eq: B^* first decomp prop}
    B^\ast \le \exp\!\left(\frac{3}{n_r}\right)(v_1 + D_0) + \Tr(\boldsymbol{\Sigma})^2.
\end{equation}
Using \eqref{eq: v1 comp proof unif L2 bound} and \eqref{eq:D_0 def},
\begin{equation}
\label{eq: v_1 bound prop}
    v_1 \le \frac{1}{n_r}C_{\boldsymbol{\Sigma}},
\end{equation}
and
\begin{equation}
\label{eq: D_0 bound prop}
    D_0 \le \frac{1}{n_r}\left(C_{\boldsymbol{\Sigma}} + 3\Tr(\boldsymbol{\Sigma})^2\right).
\end{equation}
Combining \eqref{eq: D-def proof prop}, \eqref{eq: B^* first decomp prop}, \eqref{eq: v_1 bound prop}, and \eqref{eq: D_0 bound prop}, we get
\begin{equation}
\label{eq: D final bound prop}
\begin{aligned}
    D &\le 3\left(\exp\!\left(\frac{3}{n_r}\right)(v_1 + D_0) + \Tr(\boldsymbol{\Sigma})^2\right) + C_{\boldsymbol{\Sigma}}
    \\
    &\le 3\left(\exp\!\left(\frac{3}{n_r}\right)\left(\frac{1}{n_r}C_{\boldsymbol{\Sigma}} + \frac{1}{n_r}\left(C_{\boldsymbol{\Sigma}} + 3\Tr(\boldsymbol{\Sigma})^2\right)\right) + \Tr(\boldsymbol{\Sigma})^2\right) + C_{\boldsymbol{\Sigma}}
    \\
    &= \frac{3}{n_r}\exp\!\left(\frac{3}{n_r}\right)\left(2C_{\boldsymbol{\Sigma}} + 3\Tr(\boldsymbol{\Sigma})^2\right) + 3\Tr(\boldsymbol{\Sigma})^2 + C_{\boldsymbol{\Sigma}}.
\end{aligned}
\end{equation}

\medskip\noindent\textit{Bounding $P_1$.} For a random matrix $\boldsymbol{X}\in\mathbb{R}^{d\times d}$, we write $\Cov(\boldsymbol{X}) \coloneqq \Cov(\operatorname{vec}(\boldsymbol{X}))\in\mathbb{R}^{d^2\times d^2}$. As for random vectors, $\Cov$ is additive over independent summands. We have
\begin{equation}
\label{eq: P_1 final bound prop}
\begin{aligned}
    P_1 &= \mathbb{E}\bigl[\|\hat{\boldsymbol{S}}_1 -\boldsymbol{\Sigma}\|_F^2\bigr]
    \\
    &= \mathbb{E}\bigl[\|\hat{\boldsymbol{S}}_1 -\mathbb{E}[\hat{\boldsymbol{S}}_1]\|_F^2\bigr]
    \\
    &= \Tr\!\left(\Cov(\hat{\boldsymbol{S}}_1)\right)
    \\
    &= \Tr\!\left(\Cov\!\left(\frac{1}{n_r}\sum_{j = 1}^{n_r} (\boldsymbol{x}^{(0)}_j - \boldsymbol{\mu})(\boldsymbol{x}^{(0)}_j - \boldsymbol{\mu})^\top\right)\right)
    \\
    &= \Tr\!\left(\frac{1}{n_r^2}\sum_{j = 1}^{n_r} \Cov\!\left((\boldsymbol{x}^{(0)}_j - \boldsymbol{\mu})(\boldsymbol{x}^{(0)}_j - \boldsymbol{\mu})^\top\right) \right)
    \\
    &= \frac{1}{n_r}\Tr\!\left(\Cov\!\left((\boldsymbol{x}^{(0)}_1 - \boldsymbol{\mu})(\boldsymbol{x}^{(0)}_1 - \boldsymbol{\mu})^\top\right)\right)
    \\
    &= \frac{1}{n_r}\left(\mathbb{E}\bigl[\|(\boldsymbol{x}^{(0)}_1 - \boldsymbol{\mu})(\boldsymbol{x}^{(0)}_1 - \boldsymbol{\mu})^\top\|_F^2\bigr] - \|\mathbb{E}[(\boldsymbol{x}^{(0)}_1 - \boldsymbol{\mu})(\boldsymbol{x}^{(0)}_1 - \boldsymbol{\mu})^\top]\|_F^2 \right)
    \\
    &= \frac{1}{n_r}\left(\mathbb{E}\!\bigl[\|\boldsymbol{x}^{(0)}_1 - \boldsymbol{\mu}\|_2^4\bigr] - \|\boldsymbol{\Sigma}\|_F^2\right)
    \\
    &= \frac{1}{n_r}\left(C_{\boldsymbol{\Sigma}} - \|\boldsymbol{\Sigma}\|_F^2\right).
\end{aligned}
\end{equation}
First, we used the definitions of $P_t$ and $\hat{\boldsymbol{S}}_t$ from Theorem~\ref{thm:Frobenius norm consistency of Sigma} and \eqref{eq:Shat-def}, respectively. Then, we used Lemma~\ref{lem:Shat-props}(iv), the additivity of the covariance for the i.i.d. samples $\{\boldsymbol{x}^{(0)}_j\}_{j=1}^{n_r}$ and the following property
\begin{equation*}
    \Tr\!\left(\Cov\!\left(\boldsymbol{X}\right)\right) = \mathbb{E}[\|\boldsymbol{X} - \mathbb{E}[\boldsymbol{X}]\|_F^2] = \mathbb{E}[\|\boldsymbol{X}\|_F^2] - \|\mathbb{E}[\boldsymbol{X}]\|_F^2.
\end{equation*}
Finally, we used the identities $\mathbb{E}[(\boldsymbol{x}^{(0)}_1 - \boldsymbol{\mu})(\boldsymbol{x}^{(0)}_1 - \boldsymbol{\mu})^\top] = \boldsymbol{\Sigma}$,  $\|\boldsymbol{x}\boldsymbol{x}^\top\|_F^2 = \|\boldsymbol{x}\|_2^4$, and $C_{\boldsymbol{\Sigma}} = \mathbb{E}\!\bigl[\|\boldsymbol{x}^{(0)}_1 - \boldsymbol{\mu}\|_2^4\bigr]$.

Combining \eqref{eq: P_T first bound prop}, \eqref{eq: D final bound prop} and \eqref{eq: P_1 final bound prop}, we get
\begin{equation}
\label{eq: P_T final bound prop}
\begin{aligned}
    P_t &\le \frac{1}{n_r}\left(C_{\boldsymbol{\Sigma}} - \|\boldsymbol{\Sigma}\|_F^2\right) + \frac{3}{n_r^2}\exp\!\left(\frac{3}{n_r}\right)\left(2C_{\boldsymbol{\Sigma}} + 3\Tr(\boldsymbol{\Sigma})^2\right) + \frac{1}{n_r}\left(3\Tr(\boldsymbol{\Sigma})^2 + C_{\boldsymbol{\Sigma}}\right)
    \\
    &= \frac{3}{n_r^2}\exp\!\left(\frac{3}{n_r}\right)\left(2C_{\boldsymbol{\Sigma}} + 3\Tr(\boldsymbol{\Sigma})^2\right) + \frac{1}{n_r}\left(2C_{\boldsymbol{\Sigma}} + 3\Tr(\boldsymbol{\Sigma})^2 - \|\boldsymbol{\Sigma}\|_F^2\right)
    \\
    &= \frac{3}{n_r^2}\exp\!\left(\frac{3}{n_r}\right)\left(5\Tr(\boldsymbol{\Sigma})^2 + 4\|\boldsymbol{\Sigma}\|_F^2\right) + \frac{1}{n_r}\left(5\Tr(\boldsymbol{\Sigma})^2 + 3\|\boldsymbol{\Sigma}\|_F^2\right),
\end{aligned}
\end{equation}
where we used \eqref{eq:CSigma-def}.

Then, combining \eqref{eq:Cov first upper bound prop} and \eqref{eq: P_T final bound prop}, we get
\begin{equation}
\label{eq: final bound on E(Sigma(t)-Sigma F) prop}
\begin{aligned}
    \mathbb{E}\bigl[\|\boldsymbol{\Sigma}^{(t)} -\boldsymbol{\Sigma}\|_F\bigr] &\le \sqrt{\frac{3}{n_r^2}\exp\!\left(\frac{3}{n_r}\right)\left(5\Tr(\boldsymbol{\Sigma})^2 + 4\|\boldsymbol{\Sigma}\|_F^2\right) + \frac{1}{n_r}\left(5\Tr(\boldsymbol{\Sigma})^2 + 3\|\boldsymbol{\Sigma}\|_F^2\right)} + \frac{2\Tr(\boldsymbol{\Sigma})}{n_r}
    \\
    &= \frac{2\Tr(\boldsymbol{\Sigma})}{n_r} + \frac{1}{\sqrt{n_r}} \sqrt{\left(5\Tr(\boldsymbol{\Sigma})^2 + 3\|\boldsymbol{\Sigma}\|_F^2\right) + \frac{3}{n_r}\exp\!\left(\frac{3}{n_r}\right)\left(5\Tr(\boldsymbol{\Sigma})^2 + 4\|\boldsymbol{\Sigma}\|_F^2\right)}.
\end{aligned}
\end{equation}

Finally, defining $A_{\boldsymbol{\Sigma}} = 5\Tr(\boldsymbol{\Sigma})^2 + 3\|\boldsymbol{\Sigma}\|_F^2$ and $B_{\boldsymbol{\Sigma}} = 5\Tr(\boldsymbol{\Sigma})^2 + 4\|\boldsymbol{\Sigma}\|_F^2$, and combining \eqref{eq:Cov lower bound prop} and \eqref{eq: final bound on E(Sigma(t)-Sigma F) prop}, we get

\begin{equation}
\label{eq: Cov lower and upper bound prop}
    \frac{\|\boldsymbol{\Sigma}\|_F}{n_r} \le \mathbb{E}\bigl[\|\boldsymbol{\Sigma}^{(t)}-\boldsymbol{\Sigma}\|_F\bigr] \le \frac{2\Tr(\boldsymbol{\Sigma})}{n_r} + \frac{1}{\sqrt{n_r}} \sqrt{A_{\boldsymbol{\Sigma}} + \frac{3}{n_r}\exp\!\left(\frac{3}{n_r}\right)B_{\boldsymbol{\Sigma}}}.
\end{equation}
From \eqref{eq:MSE final bounds prop} and \eqref{eq: Cov lower and upper bound prop}, the proposition follows.

\end{proof}

\subsection{Proof of Theorem~\ref{prop:rate}}
\label{app:rate prop}

We first state a lemma that will be used in the following proof to compare sums and integrals.

\begin{lemma}[Sum and integral equivalence]\label{lem:sum_integral_equivalence}
Let $a \in \mathbb{Z}$, let $f : [a, \infty) \to (0, \infty)$ be non-increasing,
and fix $h \ge 0$. Then for every integer $b > a$:
\begin{enumerate}
  \item[(a)] $\displaystyle \int_a^b f(x)\,dx \le \sum_{t=a}^{b-1} f(t)
             \le f(a) + \int_a^b f(x)\,dx$;
             in particular $\sum_{t=a}^{b-1} f(t) = \int_a^b f(x)\,dx + O(1)$.
  \item[(b)] $\displaystyle \sum_{t=a}^{b-1} f(t) \asymp \int_a^b f(x)\,dx$.
  \item[(c)] $\displaystyle \int_{a+h}^{b+h} f(x)\,dx \asymp \int_a^b f(x)\,dx$.
\end{enumerate}
All implied constants are independent of $b$ and depend only on $f$ and $h$.
If moreover $\int_a^\infty f(x)\,dx = \infty$, then the ratio of the two sides in
(b) and (c) tends to $1$ as $b \to \infty$, so $\asymp$ may be replaced by $\sim$.
\end{lemma}

\begin{proof}
Since $f$ is monotone on $[a,\infty)$, it is Riemann integrable on every compact
sub-interval, so all integrals below are well defined and finite. Note also that
$f > 0$ everywhere, so every quantity appearing in a denominator is nonzero.

Throughout, we use the following elementary comparison: for every real $y \ge a$,
monotonicity gives $f(y) \ge f(x) \ge f(y+1)$ for all $x \in [y, y+1]$, and hence
\begin{equation}\label{eq:one_step}
    f(y) \;\ge\; \int_y^{y+1} f(x)\,dx \;\ge\; f(y+1).
\end{equation}

\medskip
\noindent\textbf{(a)} Fix an integer $b > a$. Applying the first inequality in
\eqref{eq:one_step} at $y = t$ and summing over $t = a, \dots, b-1$,
\begin{equation*}
    \sum_{t=a}^{b-1} f(t) \;\ge\; \sum_{t=a}^{b-1} \int_t^{t+1} f(x)\,dx
    \;=\; \int_a^b f(x)\,dx,
\end{equation*}
which is the lower bound. For the upper bound, apply the second inequality in
\eqref{eq:one_step} at $y = t-1$, i.e. $f(t) \le \int_{t-1}^{t} f(x)\,dx$, and sum
over $t = a+1, \dots, b-1$ (an empty sum if $b = a+1$):
\begin{equation*}
    \sum_{t=a}^{b-1} f(t)
    \;=\; f(a) + \sum_{t=a+1}^{b-1} f(t)
    \;\le\; f(a) + \sum_{t=a+1}^{b-1} \int_{t-1}^t f(x)\,dx = f(a) + \int_a^{b-1} f(x)\,dx
    \;\le\; f(a) + \int_a^{b} f(x)\,dx,
\end{equation*}
the last step because $f > 0$. Consequently, writing
$D(b) \coloneqq \sum_{t=a}^{b-1} f(t) - \int_a^b f(x)\,dx$, we have shown
\begin{equation}
\label{eq:discrepancy}
    0 \;\le\; D(b) \;\le\; f(a) \qquad \text{for every integer } b > a,
\end{equation}
which is precisely the assertion $\sum_{t=a}^{b-1} f(t) = \int_a^b f(x)\,dx + O(1)$,
with implied constant $f(a)$ independent of $b$.

\medskip
\noindent\textbf{(b)} Fix an integer $b > a$; then $b \ge a+1$, so by
\eqref{eq:one_step} applied at $y = a$,
\begin{equation}
\label{eq:int_lower}
    \int_a^b f(x)\,dx \;\ge\; \int_a^{a+1} f(x)\,dx \;\ge\; f(a+1) \;>\; 0 \Rightarrow \frac{1}{f(a+1)}\int_a^b f(x) \,dx \ge 1.
\end{equation}
Combining \eqref{eq:discrepancy} and \eqref{eq:int_lower},
\begin{equation*}
    \sum_{t=a}^{b-1} f(t)
    \;\le\; f(a) + \int_a^b f(x)\,dx
    \;\le\; \frac{f(a)}{f(a+1)} \int_a^b f(x)\,dx + \int_a^b f(x)\,dx ,
\end{equation*}
so that, together with the lower bound from (a),
\begin{equation*}
    \int_a^b f(x)\,dx \;\le\; \sum_{t=a}^{b-1} f(t)
    \;\le\; \Big(1 + \tfrac{f(a)}{f(a+1)}\Big) \int_a^b f(x)\,dx .
\end{equation*}
Both constants are independent of $b$, proving (b).

\medskip
\noindent\textbf{(c)} Fix $h \ge 0$ and an integer $b > a$. The change of variables
$x = y + h$ gives $\int_{a+h}^{b+h} f(x)\,dx = \int_a^b f(y+h)\,dy$, and since
$f(y+h) \le f(y)$ for $h \ge 0$,
\begin{equation}\label{eq:shift_upper}
    \int_{a+h}^{b+h} f(x)\,dx \;\le\; \int_a^b f(x)\,dx .
\end{equation}
For the reverse bound, additivity of the integral yields
\begin{equation}\label{eq:shift_diff}
    \int_a^b f(x)\,dx - \int_{a+h}^{b+h} f(x)\,dx
    \;=\; \int_a^{a+h} f(x)\,dx - \int_b^{b+h} f(x)\,dx
    \;\le\; \int_a^{a+h} f(x)\,dx \;\le\; h f(a),
\end{equation}
using $f > 0$ and then $f \le f(a)$ on $[a, a+h]$, because $f$ is non-increasing. Since $b \ge a+1$, the interval
$[a+h, b+h]$ contains $[a+h, a+h+1]$, so by \eqref{eq:one_step} at $y = a+h$,
\begin{equation*}
    \int_{a+h}^{b+h} f(x)\,dx \;\ge\; \int_{a+h}^{a+h+1} f(x)\,dx
    \;\ge\; f(a+h+1) \;>\; 0 \Rightarrow \frac{1}{f(a+h+1)}\int_{a+h}^{b+h} f(x)\,dx \ge 1.
\end{equation*}
Feeding this into \eqref{eq:shift_diff},
\[
    \int_a^b f(x)\,dx \;\le\; \int_{a+h}^{b+h} f(x)\,dx + h f(a)
    \;\le\; \Big(1 + \tfrac{h f(a)}{f(a+h+1)}\Big) \int_{a+h}^{b+h} f(x)\,dx ,
\]
which together with \eqref{eq:shift_upper} proves (c) with constants independent
of $b$.

\medskip
\noindent\textbf{The case $\int_a^\infty f(x) \,dx = \infty$.} Here
$\int_a^b f(x)\,dx \to \infty$ as $b \to \infty$, so we may divide by it for all
large $b$. By \eqref{eq:discrepancy},
\[
    \frac{\sum_{t=a}^{b-1} f(t)}{\int_a^b f(x)\,dx}
    \;=\; 1 + \frac{D(b)}{\int_a^b f(x)\,dx}
    \;\in\; \left[\,1,\; 1 + \frac{f(a)}{\int_a^b f(x)\,dx}\,\right]
    \;\xrightarrow[b \to \infty]{}\; 1,
\]
giving $\sum_{t=a}^{b-1} f(t) \sim \int_a^b f(x)\,dx$. Similarly, both
$\int_a^{a+h} f$ and $\int_b^{b+h} f$ lie in $[0, h f(a)]$, so by
\eqref{eq:shift_diff} the difference
$\int_a^b f - \int_{a+h}^{b+h} f$ lies in $[-h f(a), h f(a)]$; dividing by
$\int_a^b f(x)\,dx \to \infty$ gives
\begin{equation*}
    \left| 1 - \frac{\int_{a+h}^{b+h} f(x) \,dx}{\int_a^b f(x)\, dx} \right| \le \frac{hf(a)}{\int_a^b f(x) \, dx} \xrightarrow{b\rightarrow \infty} 0 \Rightarrow \frac{\int_{a+h}^{b+h} f(x) \,dx}{\int_a^b f(x)\, dx} \xrightarrow{b\rightarrow \infty} 1.
\end{equation*}
\qedhere
\end{proof}

\begin{proof}[Proof of Theorem~\ref{prop:rate}]
Throughout $\alpha_t\equiv\alpha\in[0,1]$ and $n_b(t)\equiv n_b$ are fixed, so that
$N_t=n_r+(t-1)n_b$ and $\kappa=2(1-\alpha)\in[0,2]$ (note that $\kappa = 0$ when $\alpha = 1$). Since $n_b = N_{t+1}-N_t$ and $N_t\ge n_r$, we have
\begin{equation*}
    \frac{N_{t + 1}}{N_t} = \frac{N_t + n_b}{N_t} = 1 + \frac{n_b}{N_t} \le 1+\frac{n_b}{n_r},
\end{equation*}
so $N_t\asymp N_{t+1}\asymp tn_b$; all implied constants below depend only on $\alpha,n_b,n_r$ and $\Tr(\boldsymbol{\Sigma})$. 

We use that for strictly positive sequences, an $\asymp$ relation valid for all large indices extends to all of them by enlarging the constants over the remaining finite set. Additionally, by~\eqref{eq:MSE-trace}, note that $\mathbb{E}\bigl[\|\boldsymbol{\mu}^{(t)} - \boldsymbol{\mu}\|_2^2\bigr] = \Tr(\boldsymbol{C}_t)$. Thus, giving an asymptotic analysis for $\Tr(\boldsymbol{C}_t)$ suffices.

\medskip\noindent\textbf{Step 1: $\sum_{t=1}^m(1-\gamma_t)=\kappa\ln N_{m+1}+O(1)$.}
By~\eqref{eq:1-gamma-beta} with $\alpha_t\equiv\alpha$,
\begin{align*}
    1-\gamma_t &= \frac{n_b\bigl[(1-\alpha)\bigl(2N_t+n_b(1+\alpha)\bigr)+\alpha^2\bigr]}{N_{t+1}^2}
    \\
    &= \kappa\,n_b\,\frac{N_t}{N_{t+1}^2} +\frac{n_b\bigl[(1-\alpha)n_b(1+\alpha)+\alpha^2\bigr]}{N_{t+1}^2}
    \\
    &= \kappa n_b\frac{N_{t+1} - n_b}{N_{t+1}^2} + \frac{n_b[n_b(1-\alpha^2)+\alpha^2]}{N_{t+1}^2}
    \\
    &= \frac{\kappa n_b}{N_{t+1}} + \left(n_b(1-\alpha^2)+\alpha^2 - \kappa n_b\right)\cdot \frac{n_b}{N_{t+1}^2}
    \\
    &= \frac{\kappa n_b}{N_{t+1}} + \left(\alpha^2 - n_b(1 - \alpha)^2\right)\beta_t,
\end{align*}
where we used
\begin{equation*}
    n_b(1-\alpha^2) + \alpha^2 - \kappa n_b = n_b\left(1 - \alpha^2 - 2(1-\alpha)\right) + \alpha^2 = n_b \left(-\alpha^2 + 2\alpha - 1\right) + \alpha^2 = \alpha^2-n_b(1-\alpha)^2.
\end{equation*}
Thus,
\begin{equation}
\label{eq: ineq sum on 1 - gamma_t proof Pros 3.13}
    \sum_{t=1}^m\left(1 - \gamma_t\right) = \kappa n_b\cdot \sum_{t=1}^m \frac{1}{N_{t+1}} + \left(\alpha^2 - n_b(1 - \alpha)^2\right)\sum_{t=1}^m \beta_t = \kappa \ln N_{m+1} + O(1).
\end{equation}
We used Lemma~\ref{lem:beta-summability} ($\sum_{t=1}^\infty \beta_t < \infty$) and
\begin{equation*}
    \sum_{t=1}^m \frac{1}{N_{t+1}} = \frac{1}{n_b}\ln N_{m+1} + O(1).
\end{equation*}
It follows from Lemma~\ref{lem:sum_integral_equivalence}(a) with $a = 1, b = m + 1$, and the function $f(x) = 1/\left(n_r + n_b x\right)$. It is positive and non-increasing, thus:
\begin{align*}
    \sum_{t=1}^m \frac{1}{N_{t+1}} = \sum_{t=1}^m \frac{1}{n_r + n_b t} &=\sum_{t=1}^m f(t)
    \\
    &= \int_1^{m+1} f(x) \ dx + O(1)
    \\
    &= \int_1^{m+1} \frac{1}{n_r + n_b x}\ dx + O(1) 
    \\
    &= \left(\frac{1}{n_b}\ln(n_r + n_b x)\right)\Big|_1^{m+1} + O(1)
    \\
    &= \frac{1}{n_b} \left(\ln\left(n_r + n_b (m+1)\right) - \ln(n_r + n_b)\right) + O(1)
    \\
    &= \frac{1}{n_b}\cdot \ln(N_{m+2}) + O(1)
    \\
    &= \frac{1}{n_b} \ln\left(N_{m+1}\right) + O(1).
\end{align*}
We used $N_{m+2} \asymp N_{m+1} \Rightarrow \ln(N_{m+2}) = \ln(N_{m+1}) + O(1)$. This holds uniformly in $m$: $1 \le N_{m+2}/N_{m+1} \le 1 + n_b/n_r$.

\medskip\noindent\textbf{Step 2: $\Gamma_{1,m}\asymp N_{m+1}^{-\kappa}$, uniformly in $m\ge1$.}
From \eqref{eq:1-gamma-beta} and $\alpha_t \in [0,1]$, we have
\begin{align*}
    0 \le 1 - \gamma_t &= \frac{n_b}{N_{t+1}^2}\cdot \left[(1 - \alpha_t)\cdot \left(2N_t + n_b(1 + \alpha_t)\right) + \alpha_t^2\right]
    \\
    &\le \frac{n_b}{N_{t+1}^2}\cdot\left(2N_t + 2n_b + 1\right)
    \\
    &= \frac{n_b}{N_{t+1}^2}\cdot\left(2N_{t+1} + 1\right)
    \\
    &\le \frac{3n_b}{N_{t+1}} \xrightarrow{t\rightarrow\infty}0
\end{align*}
That is, $1 - \gamma_t \to 0$, and there is $t_1$ with $1 - \gamma_t \le \tfrac12$ for $t\ge t_1$. That is,
\begin{equation}
\label{eq: gamma_t <= 1/2 for t>=t_1 proof Pros 3.13}
    t \ge t_1 \Rightarrow \gamma_t \ge \frac{1}{2}.
\end{equation}
Finally, from \eqref{eq: Gamma_(s,t) def}, we have
\begin{equation}
\label{eq: Gamma_{1,m} as exp(sum of log) proof Prop 3.13}
    \Gamma_{1,m} = \prod_{t = 1}^m \gamma_t = \exp\left(\sum_{t=1}^m \ln \gamma_t\right) = \exp\left(-\sum_{t=1}^m -\ln \gamma_t\right) = \exp\left(-B_m\right),
\end{equation}
where we defined
\begin{equation}
\label{eq: definition of B_m proof Prop 3.13}
    B_m \coloneqq \sum_{t=1}^m -\ln\gamma_t.
\end{equation} 
Before we analyze $B_m$, we record two elementary bounds on the logarithm. First,
$-\ln u\ge0$ for every $u\in(0,1]$, by monotonicity of $\ln$. Second,
\begin{equation}
\label{eq: log bounds proof Prop 3.13}
    (1-u) \le -\ln u \le (1-u)+(1-u)^2,
    \qquad u\in\bigl[\tfrac12,1\bigr].
\end{equation}
Indeed, writing $x\coloneqq1-u\in[0,\tfrac12]$, the series
$-\ln u=-\ln(1-x)=\sum_{k=1}^\infty x^k/k$ converges, and
\begin{equation*}
    0 \le -\ln u-(1-u) = \sum_{k=2}^\infty\frac{x^k}{k} \le \frac12\sum_{k=2}^\infty x^k=\frac{x^2}{2(1-x)}=\frac{(1-u)^2}{2u} \le (1-u)^2,
\end{equation*}
the first inequality because every term is nonnegative, and the last because
$u\ge\tfrac12$. Rearranging gives~\eqref{eq: log bounds proof Prop 3.13}.

Fix $m \ge t_1$. From \eqref{eq: definition of B_m proof Prop 3.13}, an upper bound on $B_m$ reads:
\begin{align*}
    B_m &= \underbrace{\sum_{t=1}^{t_1 - 1} -\ln \gamma_t}_{O(1)} + \sum_{t = t_1}^m -\ln \gamma_t
    \\
    &\le O(1) + \sum_{t = t_1}^m (1 - \gamma_t) + \sum_{t=t_1}^{m} (1 - \gamma_t)^2
    \\
    &\le O(1) +\sum_{t=1}^m (1 - \gamma_t) + \sum_{t=1}^m \frac{9n_b^2}{N_{t+1}^2}
    \\
    &= O(1) + \kappa\ln N_{m+1} + 9n_b \sum_{t=1}^m \beta_t
    \\
    &= O(1) + \kappa\ln N_{m+1}.
\end{align*}
We used \eqref{eq: gamma_t <= 1/2 for t>=t_1 proof Pros 3.13} and \eqref{eq: log bounds proof Prop 3.13}. Then, we used \eqref{eq: ineq sum on 1 - gamma_t proof Pros 3.13}, $(1 - \gamma_t) \le 3n_b/N_{t+1}$, the definition of $\beta_t$ from \eqref{eq:scalars}, and the summability of $\beta_t$ from Lemma~\ref{lem:beta-summability}. That is,

\begin{equation}
\label{eq: upper bound on B_m - proof Prop 3.13}
    B_m \le \kappa\ln N_{m+1} + O(1)
\end{equation}

Additionally, from \eqref{eq: definition of B_m proof Prop 3.13}, a lower bound on $B_m$ is:

\begin{equation}
\label{eq: lower bound on B_m - proof Prop 3.13}
\begin{aligned}
    B_m &\ge \sum_{t=t_1}^m -\ln\gamma_t 
    \\
    &\ge \sum_{t=t_1}^m (1 - \gamma_t) 
    \\
    &= \sum_{t=1}^m (1 - \gamma_t) + O(1) 
    \\
    &= \kappa\ln N_{m+1} + O(1).
\end{aligned}
\end{equation}

In the first step, we discarded the terms $t<t_1$, which is legitimate since
$-\ln\gamma_t\ge0$ for every $t$, by $\gamma_t\in(0,1)$ (Lemma~\ref{lem:range}). In the
second, we applied the left-hand inequality of~\eqref{eq: log bounds proof Prop 3.13}, valid
because $\gamma_t\in[\tfrac12,1)$ for $t\ge t_1$
by~\eqref{eq: gamma_t <= 1/2 for t>=t_1 proof Pros 3.13}. In the third, we re-added the
finitely many terms $t<t_1$, each lying in $[0,1]$, at a cost of $O(1)$. The last step
is~\eqref{eq: ineq sum on 1 - gamma_t proof Pros 3.13}.

That is, from \eqref{eq: upper bound on B_m - proof Prop 3.13} and \eqref{eq: lower bound on B_m - proof Prop 3.13}, 
\begin{equation}
\label{eq: final B_m asymp proof Prop 3.13}
    B_m = \kappa\ln N_{m+1} + O(1).
\end{equation}

Finally, from \eqref{eq: Gamma_{1,m} as exp(sum of log) proof Prop 3.13} and \eqref{eq: final B_m asymp proof Prop 3.13}, we have
\begin{equation}
\label{eq: Gamma_(1,m) is N_(m+1)^(-c)}
\begin{aligned}
    \Gamma_{1,m} &= \exp\left(-\kappa\ln N_{m+1} + O(1)\right)
    \\
    &= \exp\left(\ln(N_{m+1}^{-\kappa}) + O(1)\right)
    \\
    &\asymp N_{m+1}^{-\kappa}.
\end{aligned}
\end{equation}
This was derived for $m\ge t_1$. For each of the finitely many $m<t_1$, the ratio
$\Gamma_{1,m}/N_{m+1}^{-\kappa}$ is a strictly positive real number, so over this finite range, it is bounded between its minimum and its maximum, both positive. Enlarging the implied constants accordingly extends the equivalence to every $m\ge1$. 

That is, $\Gamma_{1,m}\asymp N_{m+1}^{-\kappa}$, where the implied constants are uniform in $m\ge1$.

\medskip\noindent\textbf{Step 3: $\Gamma_{s+1,t-1}\asymp\bigl(N_{s+1}/N_t\bigr)^{\kappa}$.}
Indeed, from Step 2 and \eqref{eq: Gamma_(s,t) def}, we have
\begin{equation*}
    \Gamma_{s+1,t-1}=\frac{\Gamma_{1,t-1}}{\Gamma_{1,s}} \asymp \frac{N_t^{-\kappa}}{N_{s+1}^{-\kappa}} = \left(\frac{N_{s+1}}{N_t}\right)^\kappa
\end{equation*}
The implied constants are uniform in $s$ and $t$.

\medskip\noindent\textbf{Step 4: evaluation of $S_t$.}
By Step 3 and $\beta_s=n_b/N_{s+1}^2$ from \eqref{eq:scalars}, we have
\begin{equation}
\label{eq: S_T first compuation proof Prop 3.13}
\begin{aligned}
    S_t=\sum_{s=1}^{t-1}\beta_s\Gamma_{s+1,t-1}
    &\asymp \sum_{s=1}^{t-1} \frac{n_b}{N_{s+1}^2}\cdot \frac{N_{s+1}^\kappa}{N_t^\kappa} 
    \\
    &= n_bN_t^{-\kappa}\sum_{s=1}^{t-1} N_{s+1}^{\kappa-2}
    \\
    &= n_bN_t^{-\kappa}\sum_{s=1}^{t-1} \left(n_r + n_b s\right)^{\kappa-2},
\end{aligned}
\end{equation}
where the constants of Step~3 are independent of $t$ and therefore factor out of the sum
term by term. Let us now use Lemma~\ref{lem:sum_integral_equivalence}(b) with $a = 1$ and $f(s) = (n_r + n_bs)^{\kappa-2}$. Indeed, $\kappa - 2 \le 0$, and thus $f(s)$ is positive and non-increasing. 
\begin{equation}
\label{eq: first equivalence to summation of (n_r+n_bt)^(c-2) - proof Prop 3.13}
\begin{aligned}
    \sum_{s=1}^{t-1} \left(n_r + n_b s\right)^{\kappa-2} &\asymp \int_1^t (n_r+n_bx)^{\kappa-2} \ dx
    \\
    &\overset{u=n_r+n_bx}{=} \frac{1}{n_b}\int_{n_r + n_b}^{n_r + n_b t} u^p \ du
    \\
    &= \frac{1}{n_b}\int_{n_r + n_b}^{N_t + n_b} u^p \ du
    \\
    &\asymp \frac{1}{n_b}\int_{n_r}^{N_t} u^p \ du,
\end{aligned}
\end{equation}
where we defined $p \coloneqq \kappa - 2 = -2\alpha \in [-2, 0]$, and used Lemma~\ref{lem:sum_integral_equivalence}(c) (using the decreasing and positive function $u \mapsto u^p$). Let us now split into two cases. From \eqref{eq: first equivalence to summation of (n_r+n_bt)^(c-2) - proof Prop 3.13}, we have
\begin{itemize}
    \item If $p = -1$ (equivalent to $\kappa = 1$), we have
    \begin{equation}
    \label{eq: wanted comp sum for c=1, proof Prop 3.13}
        \sum_{s=1}^{t-1} \left(n_r + n_b s\right)^{\kappa-2} \asymp \frac{1}{n_b}
        \ln(u)\Big|_{n_r}^{N_t} = \frac{1}{n_b}\ln\left(\frac{N_t}{n_r}\right)
    \end{equation}

    \item If $p \ne -1$ (equivalent to $\kappa \ne 1$, i.e., $0 \le \kappa < 1$ or $1 < \kappa \le 2$), we have
    \begin{equation}
    \label{eq: wanted comp sum for c ne 1, proof Prop 3.13}
        \sum_{s=1}^{t-1} \left(n_r + n_b s\right)^{\kappa-2} \asymp \frac{1}{n_b} \frac{u^{p+1}}{p+1}\Big|_{n_r}^{N_t} = \frac{1}{n_b} \frac{N_t^{\kappa-1} - n_r^{\kappa-1}}{\kappa-1} \asymp \begin{cases}
            \frac{n_r^{\kappa-1}}{n_b(1-\kappa)} \qquad 0 \le \kappa < 1
            \\
            \frac{N_t^{\kappa-1}}{n_b(\kappa-1)} \qquad 1 < \kappa \le 2
        \end{cases}
    \end{equation}
\end{itemize}
From \eqref{eq: wanted comp sum for c=1, proof Prop 3.13} and \eqref{eq: wanted comp sum for c ne 1, proof Prop 3.13}, we have
\begin{equation}
\label{eq: final comp sum equivalence proof Prop 3.13}
    \sum_{s=1}^{t-1} \left(n_r + n_b s\right)^{\kappa-2} \asymp \frac{1}{n_b}\cdot \begin{cases}
    1 &  0 \le \kappa < 1
    \\
    \ln(N_t) & \kappa = 1
    \\
    N_t^{\kappa-1} & 1 < \kappa \le 2
    \end{cases}
\end{equation}
Let us now substitute \eqref{eq: final comp sum equivalence proof Prop 3.13} into \eqref{eq: S_T first compuation proof Prop 3.13}. We get
\begin{equation}
\label{eq: final S_T equivalence}
    S_t \asymp \begin{cases}
        N_t^{-\kappa} & 0 \le \kappa < 1
        \\
        N_t^{-1}\ln(N_t) & \kappa = 1
        \\
        N_t^{-1} & 1 < \kappa \le 2
    \end{cases}
\end{equation}

\medskip\noindent\textbf{Step 5: combination.}
Taking traces in the closed form~\eqref{eq:CT-closed} and using $\boldsymbol{C}_1=\boldsymbol{\Sigma}/n_r$,
\begin{equation}
\label{eq: first Tr(C_T) asymp, proof Prop 3.13}
  \Tr(\boldsymbol{C}_t)=\Gamma_{1,t-1}\frac{\Tr(\boldsymbol{\Sigma})}{n_r}+\Tr(\boldsymbol{\Sigma})\,S_t \asymp N_t^{-\kappa} + S_t,
\end{equation}
where the two summands are non-negative, $\Tr(\boldsymbol{\Sigma})>0$ and
$\Tr(\boldsymbol{C}_1)=\Tr(\boldsymbol{\Sigma})/n_r>0$ are constants, and
$\Gamma_{1,t-1}\asymp N_t^{-\kappa}$ by Step~2
(Equation~\eqref{eq: Gamma_(1,m) is N_(m+1)^(-c)}); a sum of two non-negative terms is
$\asymp$ the sum of any two quantities where each $\asymp$ to the corresponding term.

Let us now split into three cases, using \eqref{eq: final S_T equivalence} for $S_t$ asymptotics, and \eqref{eq: first Tr(C_T) asymp, proof Prop 3.13} for the final $\Tr(\boldsymbol{C}_t)$ asymptotics. We also use $\kappa = 2 (1-\alpha)$.
\begin{itemize}
    \item If $0 \le \kappa < 1$ (equivalent to $\tfrac12 < \alpha \le 1$), we have $S_t \asymp N_t^{-\kappa}$. Both terms in \eqref{eq: first Tr(C_T) asymp, proof Prop 3.13} have the same asymptotics, and we obtain $\Tr(\boldsymbol{C}_t) \asymp N_t^{-\kappa} = N_t^{-2(1-\alpha)}$.

    \item If $\kappa = 1$ (equivalent to $\alpha = \tfrac12$), we have $S_t \asymp N_t^{-1}\ln(N_t)$. The second term in \eqref{eq: first Tr(C_T) asymp, proof Prop 3.13} dominates, and we obtain $\Tr(\boldsymbol{C}_t) \asymp N_t^{-1}\ln(N_t)$.

    \item If $1 < \kappa \le 2$ (equivalent to $0 \le \alpha < \frac12$), we have $S_t \asymp N_t^{-1}$. Together with $N_t^{-\kappa} = o(N_t^{-1})$, the second term in \eqref{eq: first Tr(C_T) asymp, proof Prop 3.13} dominates, and we get $\Tr(\boldsymbol{C}_t) \asymp N_t^{-1}$.
\end{itemize}
Finally, we get
\begin{equation*}
    \Tr(\boldsymbol{C}_t) \asymp \begin{cases}
        N_t^{-1} & 0 \le \alpha < \frac{1}{2}
        \\
        N_t^{-1}\ln(N_t) & \alpha = \frac{1}{2}
        \\
        N_t^{-2(1-\alpha)} & \frac{1}{2} < \alpha \le 1
    \end{cases}
\end{equation*}
Which is exactly \eqref{eq:rate}. Note, that for $\alpha = 1$ we get $\Tr(\boldsymbol{C}_t) \asymp 1$, which means that $\Tr(\boldsymbol{C}_t)$ stays bounded as $t \rightarrow \infty$.
\end{proof}

\textbf{Remark.} When $\alpha = 0$, \eqref{eq:rate}
gives $\Tr(\boldsymbol{C}_t)\asymp N_t^{-1}$, in agreement with the exact identity
$\boldsymbol{C}_t = \boldsymbol{\Sigma}/N_t$ of Theorem~\ref{thm:CT}(c).

\subsection{Additional result: the error of the mean estimator increases initially for large $\alpha$}
\label{app:transient}

In Figure~\ref{fig:alpha_convergence}(a), the mean-square error of $\boldsymbol{\mu}^{(t)}$ first increases before it decreases when $\alpha$ is large. The following proposition characterizes exactly when this happens, shows that the error then has a single peak, and bounds how large the increase can be.

{\color{black}
\begin{proposition}[Initial growth of the error]\label{prop:transient}
Under Assumption~\ref{ass:cond-ind}, let $\alpha_t\equiv\alpha\in[0,1]$ and $n_b(t)\equiv n_b$. Let
\begin{equation}\label{eq:alpha-star}
  \alpha^*(n_r,n_b)\coloneqq
  \begin{cases}
    \dfrac{\sqrt{n_r^2+(n_b-1)(n_r+n_b)}-n_r}{n_b-1}, & n_b\ge2,\\[8pt]
    \dfrac{n_r+1}{2n_r}, & n_b=1,
  \end{cases}
\end{equation}
which is the unique root in $[0,1]$ of $g(\alpha)\coloneqq(n_b-1)\alpha^2+2n_r\alpha-(n_r+n_b)$ and satisfies $\tfrac12<\alpha^*\le1$. Then:
\begin{enumerate}
  \item[(a)] \textbf{Threshold.} The error increases at the first iteration, $c_2>c_1$, if and
  only if $\alpha>\alpha^*$. For $\alpha\le\alpha^*$, $c_t$ is non-increasing in $t$.
  \item[(b)] \textbf{Single peak.} There is $t^*\in\{1,2,\ldots\}\cup\{\infty\}$ such that $c_t$ is strictly increasing for $t\le t^*$ and non-increasing for $t\ge t^*$. If $\alpha<1$, then $t^*<\infty$ and $N_{t^*}\le\max\bigl\{n_r,\ \tfrac{n_r}{2(1-\alpha)}+n_b\bigr\}$.
  \item[(c)] \textbf{Bounded growth.} For every $t$, $c_t\le c_1+\sum_{s \ge 1}\beta_s \le 2c_1$.
\end{enumerate}
\end{proposition}}

{\color{black}
\textbf{Remark (interpretation).} When $\alpha>\alpha^*$, the error first \emph{rises} above its initial value $\Tr(\boldsymbol{\Sigma})/n_r$, then decreases for $\alpha<1$, and never comes back down at $\alpha=1$. Three observations follow. First, since $\alpha^*>\tfrac12$, the initial rise
occurs only where the rate of Theorem~\ref{prop:rate} is already degraded. Second, $\alpha^*$ depends only on $n_r$ and $n_b$, and not on $\boldsymbol{\Sigma}$ or $d$, as $c_t$ is independent of both. As $n_b/n_r\to\infty$, i.e., when each batch is much larger than the initial pool of real samples, we have $\alpha^*\to1$: the real samples in the batch, $(1-\alpha)n_b$ of them in expectation, outnumber the initial pool unless $\alpha$ is close to $1$, so the error rises only if almost the entire batch is generated. As $n_b/n_r\to0$, i.e., when each batch is much smaller than the initial pool, we have $\alpha^*\to\tfrac12$: the initial estimate is already accurate, so a
mostly generated batch adds more error than it removes. Third, the rise is modest: by~(c), the error at most doubles. For $n_r=20$ and $n_b=10$, $\alpha^*\approx0.654$, and at $\alpha=1$ the error rises to $1.74$ times its initial value.}

{\color{black}
\begin{proof}

Throughout, $\alpha_t\equiv\alpha$ and $n_b(t)\equiv n_b$, so that $N_t=n_r+(t-1)n_b$ is increasing in $t$, and $\beta_t,\gamma_t$ are given by~\eqref{eq:scalars}.

\textbf{Reduction to a scalar sequence.} Taking traces in Theorem~\ref{thm:CT}(e) and using~\eqref{eq:MSE-trace}, we have
$\mathbb{E}\bigl[\|\boldsymbol{\mu}^{(t)}-\boldsymbol{\mu}\|_2^2\bigr]=c_t\Tr(\boldsymbol{\Sigma})$. It therefore suffices to study the scalar sequence $c_t$ defined in \eqref{eq:cT-scalar}:
\begin{equation*}
    c_1=\frac{1}{n_r},\qquad c_{t + 1}=\gamma_tc_t+\beta_t \quad(t\ge1)
\end{equation*}

\medskip\noindent\textbf{Step 1: important definitions and identities.} We need to analyze the difference sequences:
\begin{equation}
\label{eq:c_T diff identity}
\begin{aligned}
    c_{t + 1} - c_t &= \gamma_tc_t + \beta_t - c_t 
    \\
    &= \beta_t - (1 - \gamma_t)c_t
    \\
    &= (1 - \gamma_t)\left(\frac{\beta_t}{1 - \gamma_t} - c_t\right)
    \\
    &= (1 - \gamma_t)(\theta_t - c_t).
\end{aligned}
\end{equation}
We defined
\begin{equation}
\label{eq:theta_t definition}
    \theta_t \coloneqq \frac{\beta_t}{1 - \gamma_t} = \frac{1}{(1 - \alpha)K_t + \alpha^2}, \qquad K_t = 2N_t + n_b(1 + \alpha),
\end{equation}
where we used \eqref{eq: final formula for 1-gamma_t proof of Lemma 3.6}. Using \eqref{eq:c_T diff identity}, we have
\begin{equation}
\label{eq:c_(t+1) - theta_t identity}
\begin{aligned}
    c_{t + 1} - \theta_t &= (c_{t + 1} - c_t) + (c_t - \theta_t)
    \\
    &= -(1 - \gamma_t)(c_t - \theta_t) + (c_t - \theta_t)
    \\
    &= \gamma_t(c_t - \theta_t)
\end{aligned}
\end{equation}

\medskip\noindent\textbf{Step 2: once the error stops increasing, it never increases again.}
Fix $t \ge 1$, and suppose that $c_{t + 1} \le c_t$. Then, from \eqref{eq:c_T diff identity} and $\gamma_t < 1$ (Lemma~\ref{lem:range}), we have $\theta_t - c_t \le 0 \Rightarrow c_t \ge \theta_t$. Thus, from \eqref{eq:c_(t+1) - theta_t identity} and $\gamma_t > 0$ (Lemma~\ref{lem:range}), we have
\begin{equation}
\label{eq:c_(t+1)>=theta_t first}
    c_{t + 1} - \theta_t \ge 0 \Rightarrow c_{t + 1} \ge \theta_t.
\end{equation}
From \eqref{eq:theta_t definition}, $\theta_t$ is non-increasing in $t$, thus $\theta_t \ge \theta_{t + 1}$, and from \eqref{eq:c_(t+1)>=theta_t first}, we have $c_{t + 1} \ge \theta_{t + 1}$. Using \eqref{eq:c_T diff identity}, we get $c_{t + 2} - c_{t + 1} \le 0 \Rightarrow c_{t + 2} \le c_{t + 1}$. Hence, by induction $c_{s + 1} \le c_s$ for all $s \ge t$.

\medskip\noindent\textbf{(a) Threshold.} Using~\eqref{eq:c_T diff identity} and $\gamma_t < 1$ (Lemma~\ref{lem:range}) at $t = 1$, $c_2 > c_1$ if and only if $c_1 < \theta_1$. With $c_1 = 1/n_r$, $N_1=n_r$
and~\eqref{eq:theta_t definition}, this reads
\begin{align*}
    \frac{1}{n_r} < \frac{1}{(1-\alpha)\left(2n_r + n_b(1+\alpha)\right) + \alpha^2} &\Leftrightarrow (1-\alpha)\left(2n_r + n_b(1+\alpha)\right) + \alpha^2 < n_r
    \\
    &\Leftrightarrow \alpha^2 + n_b(1 - \alpha^2) + 2n_r(1 - \alpha) - n_r < 0
    \\
    &\Leftrightarrow (1 - n_b)\alpha^2 - 2n_r\alpha + (n_r + n_b) < 0
\end{align*}
Define
\begin{equation}
\label{eq:g(alpha) def inc thm}
    g(\alpha) = (n_b - 1)\alpha^2 + 2n_r\alpha - (n_r + n_b).
\end{equation}
Thus, $c_2>c_1$ if and only if $g(\alpha) > 0$. On $[0,1]$, $g$ is strictly increasing, as
\begin{equation*}
    g'(\alpha) = 2(n_b - 1)\alpha + 2n_r > 0,
\end{equation*}
and
\begin{equation*}
  g(0) = -(n_r+n_b)<0,\qquad g\bigl(\tfrac12\bigr)=-\tfrac{3n_b+1}{4}<0,\qquad g(1)=n_r-1\ge0 .
\end{equation*}
Hence $g$ has a unique root $\alpha^*\in(\tfrac12,1]$, which solving the quadratic $g(\alpha) = 0$ shows to be
given by~\eqref{eq:alpha-star}:
\begin{equation*}
  \alpha^*(n_r,n_b)\coloneqq
  \begin{cases}
    \dfrac{\sqrt{n_r^2+(n_b-1)(n_r+n_b)}-n_r}{n_b-1}, & n_b\ge2,\\[8pt]
    \dfrac{n_r+1}{2n_r}, & n_b=1,
  \end{cases}
\end{equation*}
And $g(\alpha)>0$ if and only if $\alpha>\alpha^*$. Finally, if $\alpha\le\alpha^*$, then $c_2 \le c_1$, and by Step~2 with $t=1$, $c_t$ is non-increasing for all $t$.

\medskip\noindent\textbf{(b) Single peak.} Let $t^*\coloneqq\min\{t\ge1:\ c_{t+1}\le c_t\}$, with $t^*=\infty$ if no such $t$ exists. By definition, $c_{t+1}>c_t$ for all $t<t^*$, so $c_t$ is strictly increasing for $t\le t^*$. If $t^*<\infty$, Step~2 with $t=t^*$ shows that $c_t$ is non-increasing for $t\ge t^*$.

It remains to show that $t^*<\infty$ when $\alpha<1$, and to bound $N_{t^*}$. Since $K_t\ge 2N_t$ and $\alpha^2\ge0$,~\eqref{eq:theta_t definition} gives
\begin{equation*}
    \theta_t\le\frac{1}{2(1-\alpha)N_t}.
\end{equation*}
Since $N_t\to\infty$, there is a first index $t_1$ with $N_{t_1}\ge n_r/\bigl(2(1-\alpha)\bigr)$, and then $\theta_{t_1}\le 1/n_r=c_1$. We claim that $t^*\le t_1$. Assume, by contradiction, that $t^*>t_1$. Then $c_t$ is increasing up to $t_1$, so $c_{t_1}\ge c_1\ge\theta_{t_1}$. By~\eqref{eq:c_T diff identity}, $c_{t_1+1}-c_{t_1}=(1-\gamma_{t_1})(\theta_{t_1}-c_{t_1})\le0$, so $t_1$ belongs to the set defining $t^*$. Hence $t^*\le t_1$, which is a contradiction. Therefore, $t^*\le t_1$.

Since $N_t$ is increasing, $N_{t^*}\le N_{t_1}$. If $t_1=1$, then $N_{t_1}=n_r$. Otherwise, by the minimality of $t_1$, $N_{t_1-1}<n_r/\bigl(2(1-\alpha)\bigr)$, and hence $N_{t_1}=N_{t_1-1}+n_b<n_r/\bigl(2(1-\alpha)\bigr)+n_b$. In both cases,
\begin{equation*}
    N_{t^*}\le\max\Bigl\{n_r,\ \frac{n_r}{2(1-\alpha)}+n_b\Bigr\}.
\end{equation*}

\medskip\noindent\textbf{(c) Bounded growth.} Since $\gamma_t < 1$ (Lemma~\ref{lem:range}),
\begin{equation*}
    c_{t+1} = \gamma_tc_t+\beta_t\le c_t+\beta_t.
\end{equation*}
Summing over $t \ge 1$ and using Lemma~\ref{lem:beta-summability}, we get
\begin{equation*}
    c_t \le c_1 + \sum_{s \ge 1} \beta_s \le \frac{1}{n_r} + \frac{1}{n_r} = 2c_1.
\end{equation*}
\end{proof}}

\section{The value of identifying real samples}
\label{sec:detection_value}

{\color{black}
We quantify the gain from perfect detection of synthetic samples under the setting of Theorem~\ref{prop:rate}, i.e., $\alpha_t \equiv \alpha$ and $n_b(t) \equiv n_b$. The pool contains, in expectation, $N_{\mathrm{oracle}}\coloneqq(1-\alpha)N_t+\alpha n_r$ real samples, which is approximately the sample size available to an oracle that identifies and keeps only them. On the other hand, the sample mean of $N$ i.i.d.\ real samples has an error of $\Tr(\boldsymbol{\Sigma})/N$, so the pooled estimator has the effective sample size $N_{\mathrm{eff}}\coloneqq \Tr(\boldsymbol{\Sigma})/\mathbb{E}\bigl[\|\boldsymbol{\mu}^{(t)}-\boldsymbol{\mu}\|_2^2\bigr]=\Tr(\boldsymbol{\Sigma})/\Tr(\boldsymbol{C}_t)$ (by~\eqref{eq:MSE-trace}), i.e., the number of i.i.d.\ real samples that attain the same error. The gain from perfect detection is the ratio
\begin{equation}\label{eq:gain-def}
  G_\alpha(N_t)\coloneqq\frac{N_{\mathrm{oracle}}}{N_{\mathrm{eff}}}
  =\bigl((1-\alpha)N_t+\alpha n_r\bigr)\,\frac{\Tr(\boldsymbol{C}_t)}{\Tr(\boldsymbol{\Sigma})}.
\end{equation} 
As $N_t\to\infty$, Theorem~\ref{prop:rate} gives
\begin{equation}\label{eq:gain-asymp}
  G_\alpha(N_t)\asymp
  \begin{cases}
    1, & 0\le\alpha<\tfrac12,\\
    \ln N_t, & \alpha=\tfrac12,\\
    N_t^{2\alpha-1}, & \tfrac12<\alpha<1.
  \end{cases}
\end{equation}
In words, when generated samples are a minority of each new batch, perfect detection improves the effective sample size by at most a constant factor, no matter how much data is collected. When they are a majority, the improvement grows without bound as more data is collected. Moreover, the gain grows with $\alpha$ in the long run: for any $0\le\alpha<\alpha'<1$ with $\alpha'>\tfrac12$, we have $G_{\alpha'}(N_t)/G_\alpha(N_t)\to\infty$ as $N_t\to\infty$. Thus, asymptotically, the gain from detection is largest exactly where the pooled estimator converges most slowly.

At a fixed, finite $N_t$, however, the gain does not increase all the way to $\alpha=1$. At the two extremes,
\begin{equation}\label{eq:gain-alpha1}
  G_0(N_t)=1,\qquad
  1\le G_1(N_t)=n_r\,\frac{\Tr(\boldsymbol{C}_t)}{\Tr(\boldsymbol{\Sigma})}\le2,
\end{equation}
where the bounds follow from Proposition~\ref{prop: alpha = 1 perm bias} and Lemma~\ref{lem:bound}. At $\alpha=0$ there is nothing to detect, and at $\alpha=1$ the oracle retains only the $n_r$ initial real samples. For fixed $t$, $G_\alpha(N_t)$ is continuous in $\alpha$, and by~\eqref{eq:gain-asymp}, for any fixed $\alpha\in(\tfrac12,1)$ it exceeds $2$ once $N_t$ is large enough. Hence, for large but finite $N_t$, the gain is maximized at some $\alpha$ strictly between $0$ and $1$. In practice, detection is therefore worth most when generated data dominates, but real data is still being added in meaningful amounts. The potential for such detection to improve performance is supported empirically by prior work on identifying and curating model-generated data \citep{drayson2025machine}.}

\section{Additional details on empirical verification and discussion}
\label{sec:empirical_verification_details}

Under the same setup as Section~\ref{sec:empirical_verification}, we present a numerical verification of the theoretical recursion of $\Tr(\boldsymbol{C}_t) = \mathbb{E}[\|\boldsymbol{\mu}^{(t)}-\boldsymbol{\mu}\|_2^2]$ (Equation~\eqref{eq:MSE-trace} and Theorem~\ref{thm:CT}).

\begin{figure}[t]
  \centering
  \includegraphics[width=\textwidth]{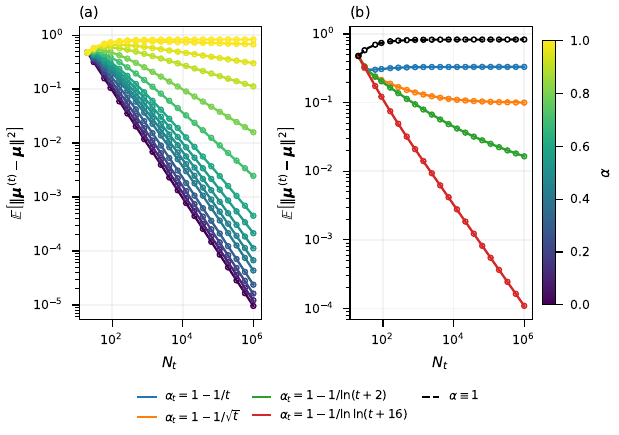}
  \caption{{\color{black} Exact $\mathbb{E}[\|\boldsymbol{\mu}^{(t)}-\boldsymbol{\mu}\|_2^2]$ against simulation. (a) Constant $\alpha$ (color bar) and (b) the four time-varying sequences of Figure~\ref{fig:alpha_convergence}, with $\alpha\equiv1$ (dashed). Lines are the recursion of Theorem~\ref{thm:CT}, and circles are Monte-Carlo averages.}}
  \label{fig:mu_theory_validation}
\end{figure}

{\color{black} \textbf{Exact mean-square error.}
The recursion of Theorem~\ref{thm:CT} gives the mean-square error of $\boldsymbol{\mu}^{(t)}$
exactly, without sampling; Figure~\ref{fig:rate_and_gain}(a) and (b) use these exact values.
Figure~\ref{fig:mu_theory_validation} compares them with the Monte-Carlo averages of
Section~\ref{sec:empirical_verification}: in all $19$ regimes and across the whole range of $N_t$, the two differ by at most $3.4$ standard errors, i.e., only by Monte-Carlo noise.}

Finally, we verify numerically that $\boldsymbol{\mu}^{(t)}$ is unbiased and that the bias of
$\boldsymbol{\Sigma}^{(t)}$ is exactly $-\boldsymbol{C}_t$ (Theorem~\ref{thm:estimators_bias}).
Both checks use $T=400$ iterations, so that $N_t$ ranges up to about $4000$.

\begin{figure}[t]
  \centering
  \includegraphics[width=\textwidth]{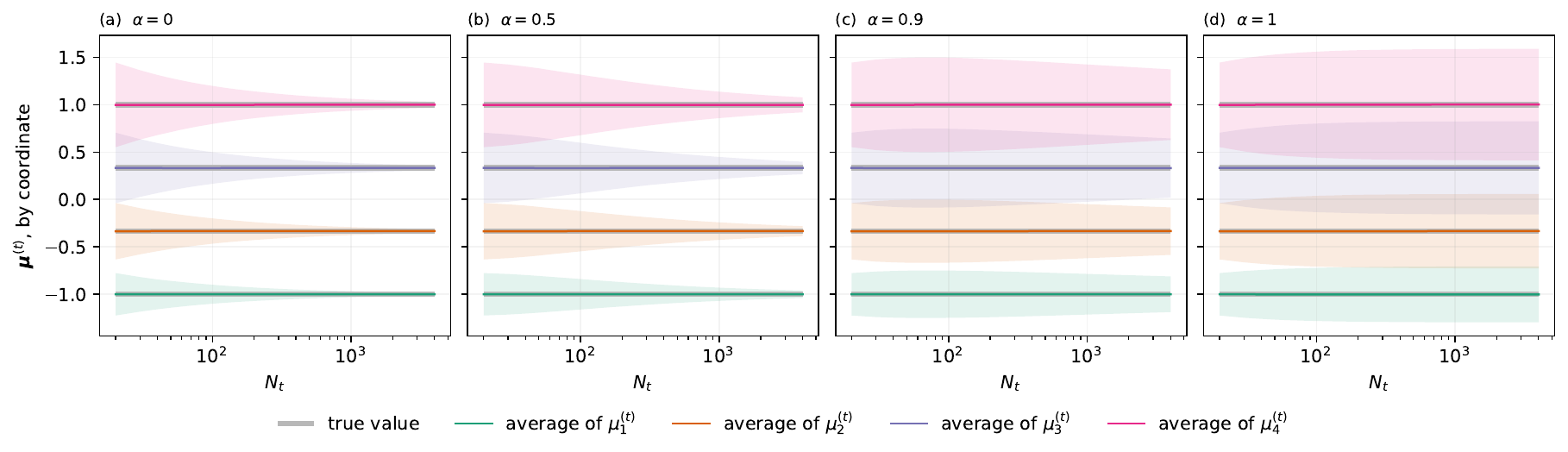}
  \caption{Unbiasedness of $\boldsymbol{\mu}^{(t)}$ for four constant values of $\alpha$. For each coordinate $i$, the colored line is the Monte-Carlo average of $\mu^{(t)}_i$, the wide gray line underneath is the true value $\mu_i$, and the shaded band is $\pm1$ standard deviation of a single run's $\mu^{(t)}_i$.}
  \label{fig:bias_mu}
\end{figure}

\begin{figure}[t]
  \centering
  \includegraphics[width=\textwidth]{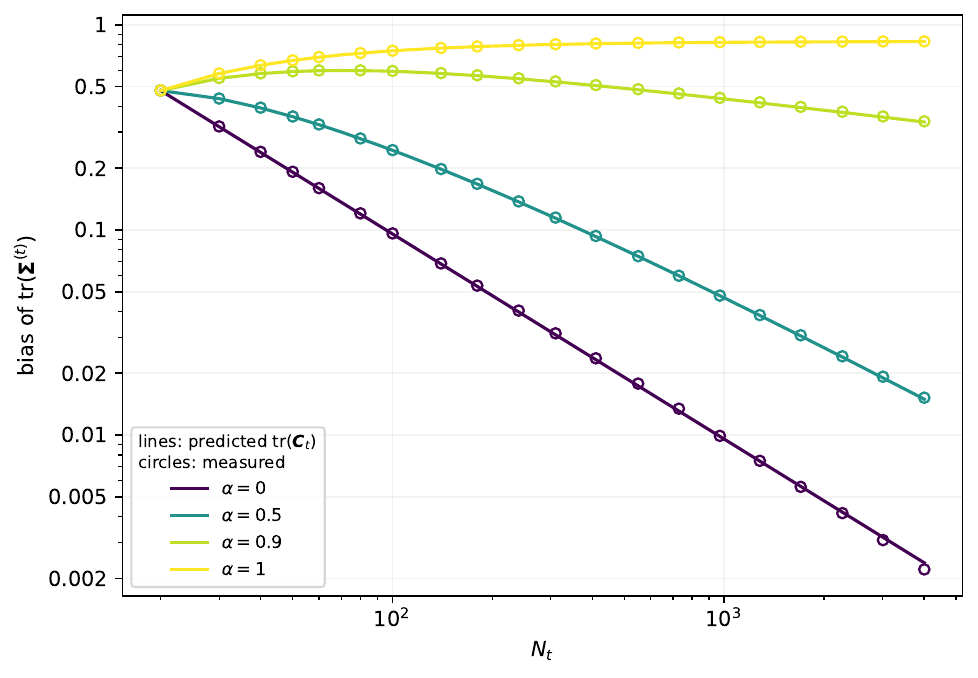}
  \caption{The bias of $\boldsymbol{\Sigma}^{(t)}$ for four constant values of $\alpha$. Lines are the predicted downward bias of the trace, $\Tr(\boldsymbol{C}_t)$, computed from the recursion of Theorem~\ref{thm:CT}; circles are the measured $\Tr(\boldsymbol{\Sigma})-\Tr(\widehat{\mathbb{E}}[\boldsymbol{\Sigma}^{(t)}])$, averaged over $2 \times 10^6$ Monte-Carlo runs.}
  \label{fig:bias_sigma}
\end{figure}

\textbf{Unbiasedness of the mean.}
Theorem~\ref{thm:estimators_bias}(a) states that $\mathbb{E}[\boldsymbol{\mu}^{(t)}]=\boldsymbol{\mu}$ for every $t \ge 1$. Figure~\ref{fig:bias_mu} illustrates this with $M=50{,}000$ independent runs of $T=400$ iterations. In all four regimes, the average of each
coordinate lies on its true value at every $N_t$. The shaded bands show that this is not because single runs are accurate: they scatter widely, and at $\alpha=0.9$ and $\alpha=1$ the band barely narrows, since the estimator converges slowly or not at all. At $\alpha=1$, in
particular, $\boldsymbol{\mu}^{(t)}$ is unbiased without being consistent, in line with Proposition~\ref{prop: alpha = 1 perm bias}. To quantify the agreement beyond what the figure can resolve, we compare the Monte-Carlo average with the true value in units of its standard error: over all $19$ regimes, all $400$ iterations and all four coordinates, the difference never exceeds 2.81 standard errors, as expected from Monte-Carlo noise alone.

\textbf{Bias of the covariance.}
Unlike the mean, $\boldsymbol{\Sigma}^{(t)}$ is biased: by
Theorem~\ref{thm:estimators_bias}(b), $\mathbb{E}[\boldsymbol{\Sigma}^{(t)}]-\boldsymbol{\Sigma} = -\boldsymbol{C}_t$, so the covariance is underestimated by exactly the covariance of the mean estimator. Figure~\ref{fig:bias_sigma} checks this for the trace. In all four regimes, the measured bias lies on the predicted $\Tr(\boldsymbol{C}_t)$ over the whole range: it decays like
$N_t^{-1}$ at $\alpha=0$, more slowly at $\alpha=0.5$ and $\alpha=0.9$, and at $\alpha=1$ it grows and then levels off at a positive value, so the covariance estimator remains biased
forever (Proposition~\ref{prop: alpha = 1 perm bias}). Across all $19$ regimes and all $400$ iterations, the measured bias differs from the prediction by at most 3.19 standard errors, as expected from Monte-Carlo noise alone.

\section{Experiments with Diffusion Models}
\label{app:empirical}

While our paper focuses on theoretical contributions, in this section we
empirically examine the extent to which the behavior observed in a practical
deep learning setting is consistent with our theoretical results.

\paragraph{Data and protocol.}
We consider CIFAR-10 \citep{krizhevsky2009learning} and a diffusion model \citep{sohl2015deep,song2021scorebased} based on the DDPM framework \citep{ho2020denoising}, using the U-Net implementation of
\citet{zoubohao_ddpm} (\url{https://github.com/zoubohao/DenoisingDiffusionProbabilityModel-ddpm-}), and train unconditional models. We split the
$50{,}000$ CIFAR-10 training images into a class-balanced validation set of
$5{,}000$ images and $45{,}000$ images available for training; the
$10{,}000$ test images are used only for evaluation. We set $n_r=n_b=5{,}000$
and $t_{\max}=9$, so that the pool at iteration $t$ contains
$N_t=n_r+(t-1)n_b$ images and the final pool, $N_9=45{,}000$, uses all
available training images. We consider four fixed mixing fractions,
$\alpha\in\{0,0.25,0.75,1\}$. At $t=1$ we train a DDPM on $n_r$
class-balanced real images; this model is shared by all four values of
$\alpha$. At each iteration $t\ge 2$, we accumulate the previous pool with
$\alpha n_b$ samples generated by the model of iteration $t-1$ and
$(1-\alpha)n_b$ previously unused, class-balanced real images, and %
the training 
treats real and generated
images identically: minibatches are drawn uniformly from the pool, with no
labels, reweighting or filtering. %

{\color{black} Unlike the theoretical setup, where each sample is drawn from the mixture independently, here the numbers of generated and real images in each batch are fixed. This keeps the consumption of real images exact and class-balanced, and differs negligibly from independent draws, whose count of generated images fluctuates by less than $1\%$ of the batch (one standard deviation) for $n_b=5{,}000$.}

\paragraph{Training.}
All models share one recipe: $1{,}000$ diffusion steps with a linear noise
schedule from $10^{-4}$ to $0.02$, the $\epsilon$-prediction objective, Adam
with learning rate $2\times10^{-4}$ after a $5{,}000$-step linear warm-up,
batch size $128$, gradient clipping at $1$, dropout $0.1$, and an exponential
moving average (EMA) of the weights with decay $0.9999$. All sampling uses the EMA weights and DDIM \citep{song2021ddim} with $200$ steps. We fix the \emph{number of optimization steps} rather than the number of epochs, so that pool size is not confounded with the amount of computation.
{\color{black} In each iteration $t$ the training is from scratch on the whole pool, except for $\alpha=0.75$ and $t\ge 4$, where we use warm-start from the previous iteration's model with learning rate of lower order.} 

\paragraph{Checkpoint selection.}
Since diffusion models overfit on datasets of this size, each model should be trained neither too little nor too long, so that pools of different sizes are compared fairly. We save a checkpoint every $50{,}000$ steps and select the
one minimizing the FID between $10{,}000$ generated samples and the validation
set; we report the FID of the same samples against the test set, which is
never used for selection. The default budget is $200{,}000$ steps; iterations
in which the validation FID was still decreasing at the end of the budget were
retrained with a larger budget (up to $800{,}000$ steps), which was needed only
for $\alpha\in\{0,0.25\}$. Validation and test FIDs selected the same
checkpoint in every iteration, and over all checkpoints they differed by a
nearly constant offset of $2.44\pm0.07$.

\paragraph{FID values.}
We compute FID with clean-fid \citep{parmar2022aliased} from $10{,}000$ DDIM samples against the held-out test set. Published DDPM results on CIFAR-10 (e.g., $3.17$ in \citet{ho2020denoising}) use $50{,}000$ samples, $1{,}000$-step sampling and training-set statistics, which yields substantially lower values: for our best model ($\alpha=0$, $t=9$, trained on $45{,}000$ images), $50{,}000$ samples against the training set give $5.1$, compared with $9.7$
on the test set. Our FIDs are therefore comparable across the curves of Figure~\ref{fig:ddpm}, but not to published numbers.

\paragraph{Reporting.}
We report the test FID of the selected checkpoint as a function of $N_t$ for
each $\alpha$, for a single training seed. The horizontal line is the
$\alpha=0$ model at $t=9$, i.e., a DDPM trained on all $45{,}000$ training
images with the same protocol. From duplicate runs, differences below
$\approx 0.5$ FID between consecutive iterations are within run-to-run
variation. We expect the curves for $\alpha<1$ to approach this reference
value, whereas for $\alpha=1$, where no new real data are introduced, we
expect the FID to approach a finite limiting value.

{\color{black} All experiments were run on a server with eight NVIDIA RTX 6000 Ada GPUs,
training one model per GPU. Training costs about $3.9$ GPU-hours per $100{,}000$ steps including checkpoint evaluation, e.g., about $8$, $15$ and $31$ hours for budgets of $200{,}000$, $400{,}000$ and $800{,}000$ steps; the runs shown in Figure~\ref{fig:ddpm} total roughly $400$ GPU-hours.}

\begin{figure}[t]
  \centering
  \includegraphics[width=\textwidth]{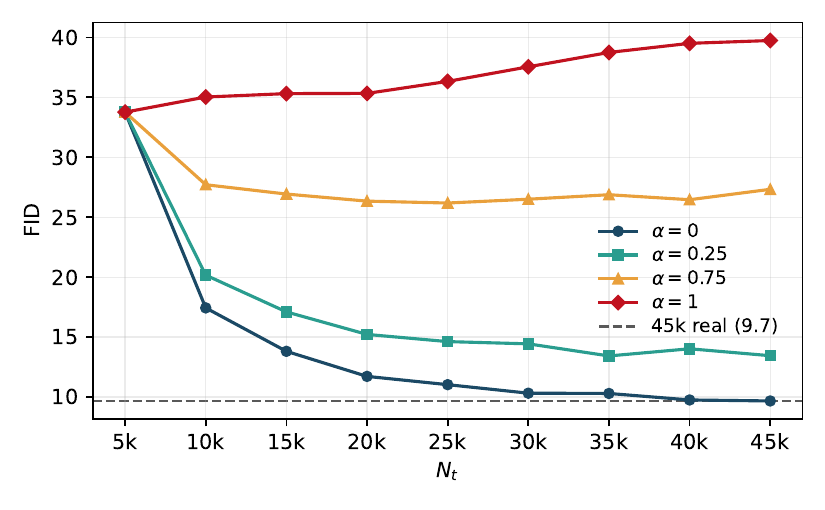}
  \caption{Test FID of DDPM on CIFAR-10 as a function of the pool size $N_t$, for mixing fractions $\alpha\in\{0,0.25,0.75,1\}$ (single seed). At each iteration, the checkpoint with the lowest validation FID is selected and its test FID is reported. %
  The dashed line is the $\alpha=0$ model at $t=9$, trained on all $45{,}000$ training images (FID $9.7$ on the test set, and $5.1$ with $50{,}000$ samples against the training set; see text). Differences below $\approx 0.5$ FID between consecutive iterations are within run-to-run variation.}
\label{fig:ddpm}
\end{figure}

\paragraph{Comparison with the theory.}
The central prediction of our theory holds: although real and generated images
are pooled and treated identically, with no detection, labeling or reweighting,
every $\alpha<1$ improves on the initial model, whereas $\alpha=1$ degrades.
At every iteration $t\ge2$, the FID for $\alpha\in\{0,0.25,0.75\}$ is below its
initial value of $33.8$, while for $\alpha=1$ it never falls below it. Thus, a
fixed fraction of fresh real data suffices to avoid collapse even when it
cannot be told apart from generated data. This comparison is meaningful beyond
the Gaussian setting because FID is the squared 2-Wasserstein distance between Gaussians fitted to the Inception features of generated and test images, i.e., the quantity of Theorem~\ref{thm:unified_sharp_criterion} evaluated in a feature space. The finer predictions carry over as well. First, the curves are ordered by $\alpha$ at every iteration $t\ge2$: a larger fraction of generated data always yields a worse model. Second, the improvement slows as $\alpha$
grows, in line with Theorem~\ref{prop:rate}: from $N_1=5{,}000$ to $N_3=15{,}000$, the FID decreases by $19.9$ for $\alpha=0$, $16.7$ for
$\alpha=0.25$ and only $6.8$ for $\alpha=0.75$, the one value above $\tfrac12$, whose predicted rate $N_t^{-1/2}$ is slower than the rate $N_t^{-1}$ of $\alpha<\tfrac12$. For $\alpha=0.25$, the curve follows that of $\alpha=0$ with a roughly constant offset of $3$--$4$ FID, consistent with synthetic data
inflating the constant but not the exponent. Third, at $\alpha=1$ the FID rises
from $33.8$ to $39.7$, with increments that shrink over the last iterations,
matching the bounded, non-decreasing error of
Propositions~\ref{prop: alpha = 1 perm bias} and~\ref{prop:transient}.

The value of identifying real samples (Section~\ref{sec:convergence_rate_sec})
is also visible. Discarding all generated images would leave, at $t=9$,
$35{,}000$ real images for $\alpha=0.25$ and $15{,}000$ for $\alpha=0.75$,
i.e., the $\alpha=0$ model at $N_t=35{,}000$ (FID $10.3$) and at $N_t=15{,}000$ (FID $13.8$). Compared with FIDs of $13.5$ and above $27$, respectively, perfect detection would gain about $3$ FID for $\alpha=0.25<\tfrac12$, but more than $13$ for $\alpha=0.75>\tfrac12$,
as the theory predicts.

\end{document}